\documentclass[journal]{IEEEtai}

\usepackage[colorlinks,urlcolor=blue,linkcolor=blue,citecolor=blue]{hyperref}

\usepackage{color,array}

\usepackage{graphicx}
\usepackage{cite}
\usepackage{tabularx}
\usepackage{graphicx}
\usepackage[caption=false,font=normalsize,labelfont=sf,textfont=sf]{subfig}
\usepackage{stfloats}
\usepackage{array}
\usepackage{multirow}
\usepackage[export]{adjustbox}
\usepackage{pgfplots}
\usepackage{pgfplotstable}
\pgfplotsset{compat=1.17}
\usepackage[absolute,overlay]{textpos}
\usepackage{amsmath}
\usepackage{amssymb}
\usepackage[utf8]{inputenc}
\usepackage[T1]{fontenc}
\usepackage{booktabs}

\usepackage{diagbox}
\newcommand{\revision}[1]{#1}

\usepackage{amsmath,amsfonts,amssymb,amsthm}
\theoremstyle{plain}

\newtheorem{assumption}{Assumption}
\begin{document}
\bstctlcite{BSTcontrol}

\title{Demystifying MuZero Planning: Interpreting the Learned Model} 

\author{
Hung Guei, \IEEEmembership{Member, IEEE},
Yan-Ru Ju,
Wei-Yu Chen,
and Ti-Rong Wu, \IEEEmembership{Member, IEEE}
\thanks{
This work was supported by the National Science and Technology Council (NSTC) of the Republic of China (Taiwan) under Grant 113-2221-E-001-009-MY3, Grant 113-2221-E-A49-127, Grant 113-2634-F-A49-004, and Grant 114-2221-E-A49-005. \textit{(Corresponding author: Ti-Rong Wu)}
}

\thanks{Hung Guei, Yan-Ru Ju, and Ti-Rong Wu are with the Institute of Information Science, Academia Sinica, Taipei 11529, Taiwan. (e-mail: hguei@iis.sinica.edu.tw; yanru@iis.sinica.edu.tw; tirongwu@iis.sinica.edu.tw)}
\thanks{Wei-Yu Chen was with the Institute of Information Science, Academia Sinica, Taipei 11529, Taiwan, and the Department of Electrical Engineering, National Taiwan University, Taipei, Taiwan. He is now with the College of Computing, Georgia Institute of Technology, Atlanta, GA 30332, USA. (e-mail: b09901084@ntu.edu.tw)}
}


\maketitle

\begin{abstract}
MuZero has achieved superhuman performance in various games by using a dynamics network to predict \revision{the} environment dynamics for planning, without relying on simulators.
However, the latent states learned by the dynamics network make its planning process opaque.
This paper aims to demystify MuZero's model by interpreting the learned latent states.
We incorporate observation reconstruction and state consistency into MuZero training and conduct an in-depth analysis to evaluate latent states across two board games: 9x9 Go and Gomoku, and three Atari games: Breakout, Ms. Pacman, and Pong.
Our findings reveal that while the dynamics network becomes less accurate over longer simulations, MuZero still performs effectively by using planning to correct errors.
Our experiments also show that the dynamics network learns better latent states in board games than in Atari games.
These insights contribute to a better understanding of MuZero and offer directions for future research to improve the performance, robustness, and interpretability of the MuZero algorithm.
\revision{The code and data are available at https://rlg.iis.sinica.edu.tw/papers/demystifying-muzero-planning.}
\end{abstract}

\begin{IEEEImpStatement}
Reinforcement learning is one of the most important fields in artificial intelligence, particularly in designing agents for decision-making problems.
MuZero, a zero-knowledge learning algorithm, masters various tasks by learning the environment through a dynamics network, ranging from games such as board games and Atari to non-game applications like physical control and optimization problems.
Despite its success, the relationship between the dynamics network and MuZero planning remains unclear.
Our work systematically demonstrates the inaccuracy of the dynamics network and shows the correlation between the learning quality and the complexity of environmental observations.
Most importantly, \revision{we} establish that MuZero can correct the inaccuracies caused by the dynamics network within its planning process.
These findings not only demystify the planning process of MuZero but also identify avenues for enhancing its performance, robustness, and interpretability, offering a pathway toward more transparent and efficient AI systems across domains requiring complex decision-making.
\end{IEEEImpStatement}


\section{Introduction}\label{sec:introduction}

\IEEEPARstart{R}{einforcement} learning has shown significant success across various domains, with particularly notable achievements in games \cite{silver_mastering_2016, vinyals_grandmaster_2019, openai_dota_2019}.
One of the most remarkable milestones is AlphaZero \cite{silver_general_2018}, an algorithm that leverages environment simulators for planning and masters board games like Chess and Go through self-play without using any human knowledge.
Building on this advancement, MuZero \cite{schrittwieser_mastering_2020} introduces a \textit{dynamics network} that learns the rules of environment dynamics, enabling planning without requiring environment simulators.
This dynamics network further extends MuZero to complex environments, including visual environments like Atari games \cite{schrittwieser_mastering_2020, danihelka_policy_2022}, physics-based simulations \cite{hubert_learning_2021}, stochastic environments \cite{antonoglou_planning_2021}, and even non-game applications \cite{mandhane_muzero_2022,wang_optimizing_2023}.

Despite the effectiveness of MuZero, the dynamics network increases the opacity of its planning process, as they are not directly interpretable.
Consequently, several studies have been conducted to investigate the planning process of MuZero.
\revision{However, these studies only highlight inaccuracies in MuZero's dynamics network and their influence on the overall performance.}
\revision{For example, Hamrick et al. \cite{hamrick_role_2020} demonstrated that simpler and shallower planning is often as performant as more complex planning, implying a common limitation that inaccurate dynamics network leads to suboptimal search results.}
Vries et al. \cite{vries_visualizing_2021} found that the observation embeddings and latent states learned by the dynamics network often diverge, potentially leading to unstable planning, and proposed adding \revision{two regularization objectives} during training to align the latent states.
Moerland et al. \cite{he_what_2023} \revision{showed} that MuZero struggles to accurately evaluate policies and values when dealing with \revision{a long-term look-ahead planning or} out-of-distribution states.

\revision{Compared to previous studies, this paper aims to demystify why MuZero still achieves superhuman performance with an inaccurate dynamics network.}
\revision{Specifically, we investigate the impact of the latent states learned by MuZero during planning process with two techniques, \textit{observation reconstruction} \cite{hamrick_role_2020,vries_visualizing_2021,ye_mastering_2021,scholz_improving_2021,oren_epistemic_2025} and \textit{state consistency} \cite{ye_mastering_2021,scholz_improving_2021,niu_lightzero_2023}, in two board games: 9x9 Go and Gomoku, and three Atari games: Breakout, Ms. Pacman, and Pong.}
The contributions of this paper can be summarized as follows.
First, we confirm that the dynamics network becomes increasingly inaccurate when simulating environments further into the future, consistent with previous findings.
Second, we demonstrate that the dynamics network learns better when the observations are visually simple, as evidenced by more accurate observation reconstruction in board games compared to Atari games.
Third, we establish that although the latent states gradually become inaccurate, MuZero can still play effectively by planning to correct the inaccuracies and maintain the playing performance.
However, excessively large search trees can still diminish the playing performance.

In summary, through empirical experiments across various games, our work offers a deeper understanding of MuZero and provides directions for future research to improve the performance, robustness, and interpretability of the MuZero algorithm when applied to other domains.

\section{MuZero and its improvements}\label{sec:muzero}

MuZero \cite{schrittwieser_mastering_2020}, a model-based reinforcement learning algorithm, builds upon AlphaZero \cite{silver_general_2018} by introducing a learned model for predicting environment dynamics.
Its architecture includes three key networks: \textit{representation}, \textit{dynamics}, and \textit{prediction}.
The representation network $h$ encodes an \textit{observation} $o_t$ into a latent state, called \textit{hidden state} $s_t$, at time step $t$, denoted by $h(o_t) = s_t$. 
The dynamics network $g$ provides state transitions, $g(s_t, a_t) = r_{t+1}^{(1)}, s_{t+1}^{(1)}$, where $r_{t+1}^{(1)}$ is the \textit{predicted reward}, and $s_{t+1}^{(1)}$ is the next hidden state.
Here, the superscript ${(k)}$ indicates that the hidden state $s_t^{(k)}$ has been unrolled by the dynamics network for $k$ steps.
For simplicity, the superscript is omitted when $k=0$.
The prediction network $f$ predicts a \textit{policy} $p_t$ and a \textit{value} $v_t$ of a given hidden state $s_t$, denoted by $f(s_t) = p_t, v_t$.

\begin{figure}[t]
    \centering
    \includegraphics[width=1\linewidth]{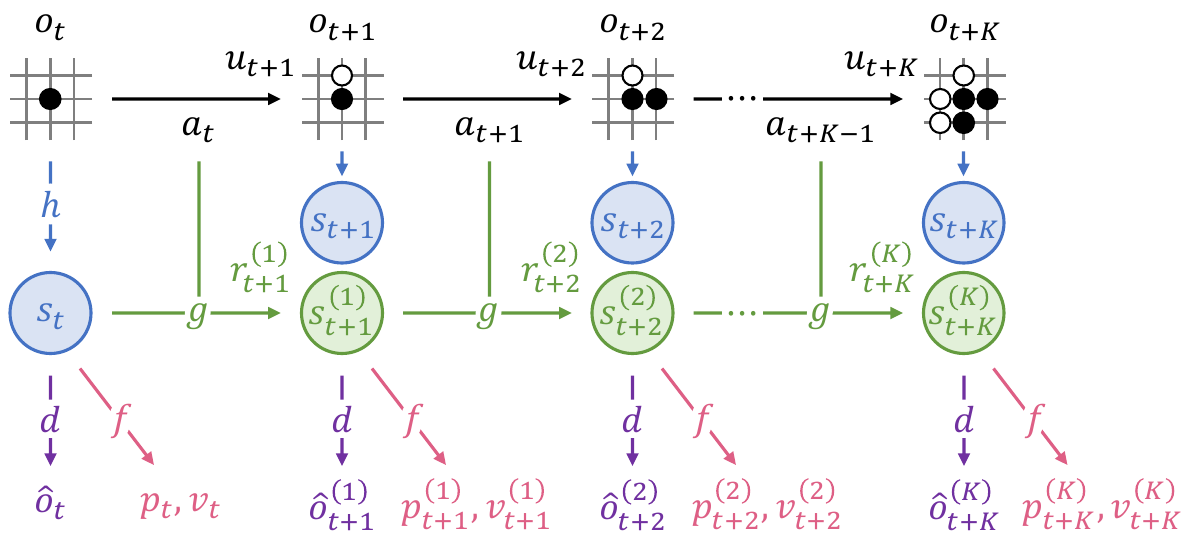}
    \caption{An illustration of the MuZero unrolling process.}
    \label{fig:muzero-decoder-unroll}
\end{figure}

The training process of MuZero is composed of self-play and optimization. 
Self-play uses Monte Carlo tree search (MCTS) \cite{kocsis_bandit_2006, coulom_efficient_2007, browne_survey_2012} by applying three networks to collect training game records. 
Namely, an MCTS search tree is constructed for each move, in which the representation network $h$ encodes the observation $o_t$ at the root node, the dynamics network $g$ expands the child nodes for planning ahead, and the prediction network $f$ estimates \revision{a} policy and \revision{a} value for each node to guide the search tree.
Once the MCTS is complete, an action $a_t$ is sampled from the \textit{search policy} $\pi_t$ and applied to the environment, resulting in an \textit{observed reward} $u_{t+1}$ and the next observation $o_{t+1}$.
This process continues until the game ends and an \textit{outcome} $z$ is obtained.
The optimization trains the networks using collected game records.
For each training record, MuZero obtains $s_t = h(o_t)$ and recurrently unrolls $s_t$ for $K$ steps to obtain all $s_{t+k}^{(k)}$, $r_{t+k}^{(k)}$, $p_{t+k}^{(k)}$, and $v_{t+k}^{(k)}$ for $1 \leq k \leq K$, as illustrated in Fig. \ref{fig:muzero-decoder-unroll}.
The networks are optimized jointly by policy loss $l^p$, value loss $l^v$, and reward loss $l^r$, as 
\begin{equation}\label{eq:loss_muzero}
\begin{split}
L_{\mu} = & \sum_{k=0}^{K} l^{p}(\pi_{t+k}, p_{t+k}^{(k)}) + \sum_{k=0}^{K} l^{v}(z_{t+k}, v_{t+k}^{(k)}) \\
& + \sum_{k=1}^{K} l^{r}(u_{t+k}, r_{t+k}^{(k)}) + c||\theta||^{2}.
\end{split}
\end{equation}
where $z_{t+k}$ is either the outcome $z$ for board games or the n-step return for Atari games, and the last term is an L2 regularization.
When newer networks are prepared, they will be used by self-play.

Several improvements have been proposed since the advent of MuZero, particularly two regularizations aimed at improving the state transitions of the dynamics network: \textit{observation reconstruction} and \textit{state consistency}.
For the observation reconstruction \revision{\cite{hamrick_role_2020}}, a \textit{decoder} network $d$, denoted as $d(s_t) = \hat o_t$, is adopted to obtain $\hat{o}_{t+k}^{(k)}$ for each $s_{t+k}^{(k)}$ as shown in Fig. \ref{fig:muzero-decoder-unroll}, then the network is optimized using the difference between $o_{t+k}$ and $\hat{o}_{t+k}^{(k)}$.
This allows MuZero to predict future observations without qualitatively changing the planning behavior \cite{hamrick_role_2020}.
For the state consistency \revision{\cite{ye_mastering_2021}}, SimSiam \cite{chen_exploring_2021} is adopted to align \revision{the} hidden states $s_{t+k}$ and $s_{t+k}^{(k)}$, thereby increasing the accuracy of the dynamics network.
The decoder loss $L_{d}$ and state consistency loss $L_{c}$ are defined as
\begin{equation}\label{eq:loss_decoder_consistency}
L_{d}=\sum_{k=0}^{K}{l^{d}(o_{t+k},\hat{o}_{t+k}^{(k)})} \mbox{ and } L_{c}=\sum_{k=1}^{K}{l^{c}(s_{t+k},s_{t+k}^{(k)})}.
\end{equation}
During the optimization, a joint loss $L = L_\mu + \lambda_d L_d + \lambda_c L_c$ is applied, in which $\lambda_d$ and $\lambda_c$ \revision{control} the scale of $L_d$ and $L_c$, respectively.

\revision{
These regularizations have been widely applied due to their generality.
For example, Hamrick et al. \cite{hamrick_role_2020} investigated the effect of planning with $L_{d}$ and showed that the performance is relatively unchanged.
Vries et al. \cite{vries_visualizing_2021} visualized the latent states in MuZero with $L_{d}$ and $L_{c}$ and proposed the concept of using them to help interpret the look-ahead search.
Ye et al. \cite{ye_mastering_2021} focused on learning with limited data and established the use of SimSiam for calculating $L_{c}$.
Scholz et al. \cite{scholz_improving_2021} explored various combinations of $L_{d}$ and $L_{c}$ to enhance performance.
Furthermore, several works \cite{niu_lightzero_2023,oren_epistemic_2025} adopted them as a known improvement.
In this paper, we provide a comprehensive analysis of how MuZero interprets and exploits latent states for planning by utilizing these regularizations.
}

\section{Interpreting the hidden states}\label{sec:interpreting-hidden-states}

This section evaluates the quality of hidden states learned by MuZero.
We first train five MuZero models, each on a different game, with a decoder network and state consistency: two board games, 9x9 Go and Gomoku (with Outer-Open rule), and three Atari games, Breakout, Ms. Pacman, and Pong.
We then use the decoder network to interpret the hidden states.

\begin{table}[t]
    \caption{Hyperparameters for training MuZero with a decoder.}
    \centering
    \begin{tabular}{lcc}
        \toprule
        Hyperparameter & Board Games & Atari Games\\
        \midrule
        Iteration & \multicolumn{2}{c}{300}\\
        Optimizer & \multicolumn{2}{c}{SGD}\\
        Optimizer: learning rate & \multicolumn{2}{c}{0.1}\\
        Optimizer: momentum & \multicolumn{2}{c}{0.9}\\
        Optimizer: weight decay & \multicolumn{2}{c}{0.0001}\\
        Training steps & \multicolumn{2}{c}{60k}\\
        Unroll steps & \multicolumn{2}{c}{5}\\
        Batch size & \multicolumn{2}{c}{512}\\
        Replay buffer size & 40k games & 1M frames\\
        Max frames per episode & - & 108k\\
        Discount factor & - & 0.997\\
        Priority exponent ($\alpha$) & - & 1\\
        Priority correction ($\beta$) & - & 0.4\\
        Bootstrap step (n-step return) & - & 5\\
        \# Blocks & 3 & 2\\
        \# Simulations & 16 & 18 \\
        Decoder loss coefficient ($\lambda_d$) & 1 & 25 \\
        Consistency loss coefficient ($\lambda_c$) & 0 & 1 \\
        \bottomrule
    \end{tabular}
    \label{tab:hyperparameters}
\end{table}

\subsection{Training MuZero Models with a Decoder}\label{sec:interpreting-hidden-states:training}
To train the MuZero model with a decoder network and state consistency, we use the MiniZero framework \cite{wu_minizero_2024}, an \revision{open-source} framework supporting AlphaZero and MuZero algorithms, for implementation.
Specifically, the hyperparameters used in training are listed in Table \ref{tab:hyperparameters}.
The experiments are conducted on a machine with two Intel Xeon Silver 4216 CPUs and four NVIDIA RTX A5000 GPUs.

Generally, we follow the MuZero network architecture \cite{schrittwieser_mastering_2020} that uses a residual network \cite{he_deep_2016} as the backbone, and follow the approach in \cite{ye_mastering_2021} to use the reverse order of operations of the representation network to design the decoder network.
For board games, the observations have a size of $b \times b \times c$, where $b$ is the board size and $c$ is the number of feature channels; the decoded observations have the same size as the observations.
Here, following \cite{wu_minizero_2024}, we use $b=9, c=18$ for Go, and $b=15, c=4$ for Gomoku, where the channels are composed of binary data (0 or 1) representing specific features: for Go, the 1st to 16th channels mark the own and opponent pieces of the last eight turns, and the 17th and 18th indicate whether it is for Black or White to play, respectively; for Gomoku, the 1st and 2nd channels are the own and opponent piece of the current turn, and the 3rd and 4th channels are also used for indicating the current player.
For Atari games, the observations have a size of $96 \times 96 \times 32$, namely the $96 \times 96$ image (resized from the original $160 \times 210$ game screen frame) and \revision{the} $32$ channels containing the last eight frames, each with three RGB planes and one action plane.
The decoded observations contain only the RGB planes of the current time step, namely only $96 \times 96 \times 3$ channels.
In addition, we increase the decoder loss coefficient $\lambda_d$ to $25$ to encourage the decoder to learn\footnote{For a higher $\lambda_d$, we clip the gradients of the decoder network using the PyTorch \cite{paszke_pytorch_2019} API \texttt{torch.nn.utils.clip\_grad\_value\_} with a clip setting of 0.001 during training to avoid potential loss spikes.} the visually complex Atari observations and apply state consistency (with the SimSiam network architecture \cite{chen_exploring_2021}) to improve hidden state alignment.

\begin{table}[t]
    \caption{The comparison of the playing performance between MuZero with and without a decoder in five games.
    }
    \centering
    \begin{tabular}{lrr}
        \toprule
        Game & w/ Decoder & w/o Decoder\\
        \midrule
        Go & \textbf{1088.74} & 1000.00 \\
        Gomoku & \textbf{1048.96} & 1000.00 \\
        \midrule
        Breakout & 358.90 & \textbf{383.17} \\
        Ms. Pacman & \textbf{4528.70} & 3732.80 \\
        Pong & 19.65 & \textbf{20.07} \\
        \bottomrule
    \end{tabular}
    \label{tab:compare-vanilla-vs-decoder}
\end{table}

To ensure the performance of each model, we evaluate the playing performance to verify whether training with a decoder network impacts MuZero's planning ability.
For the comparison, we additionally train a baseline model for each game, namely a MuZero model without a decoder\revision{,} using the same hyperparameters except setting $\lambda_d=0$.
For board games, we use the Elo ratings \cite{silver_general_2018} to evaluate the model by 200 games against the baseline model (search with 400 simulations per move), in which the baseline is anchored to 1000 Elo.
For Atari games, we calculate the average returns of the latest 100 finished self-play games. 
The evaluation results are summarized in Table \ref{tab:compare-vanilla-vs-decoder}, where MuZero with a decoder performs slightly better in board games, while its performance varies from game to game in Atari games.
Overall, adding a decoder network does not significantly affect overall playing performance, consistent with the findings of Hamrick et al. \cite{hamrick_role_2020}.

\subsection{Decoding hidden states into observations}\label{sec:interpreting-hidden-states:decoding}

\begin{figure}[t]
    \centering
    \small
    {\setlength{\tabcolsep}{0.1em}
    \begin{tabular}[]{cccccc}
        \revision{\footnotesize$o_t$} &
        \includegraphics[height=0.090\textwidth, valign=c]{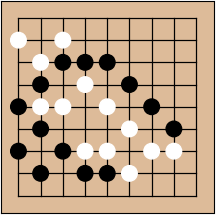} &
        \includegraphics[height=0.090\textwidth, valign=c]{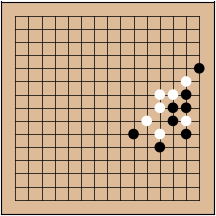} &
        \includegraphics[height=0.090\textwidth, valign=c]{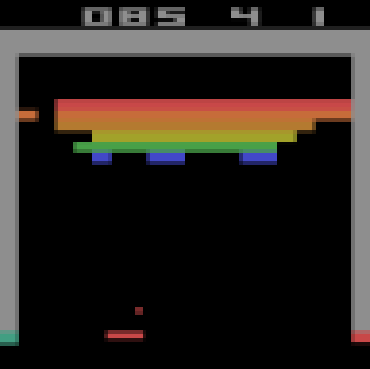} &
        \includegraphics[height=0.090\textwidth, valign=c]{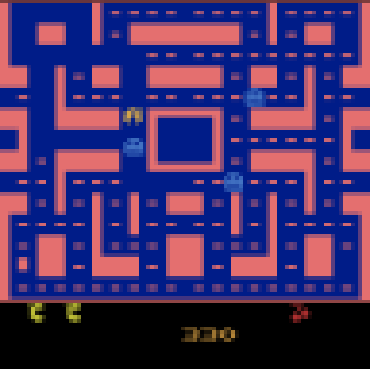} &
        \includegraphics[height=0.090\textwidth, valign=c]{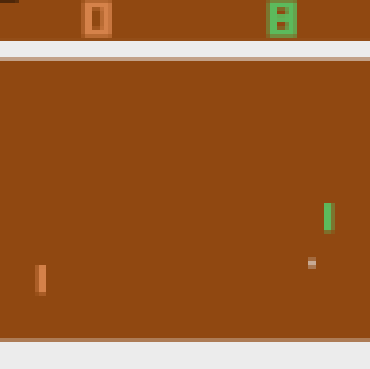} \vspace{0.2em} \\
        \revision{\footnotesize$\hat o_t$} &
        \includegraphics[height=0.090\textwidth, valign=c]{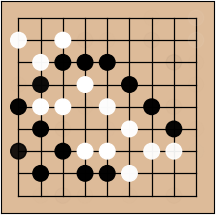} &
        \includegraphics[height=0.090\textwidth, valign=c]{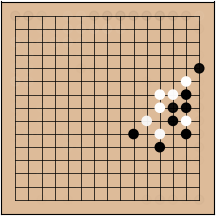} &
        \includegraphics[height=0.090\textwidth, valign=c]{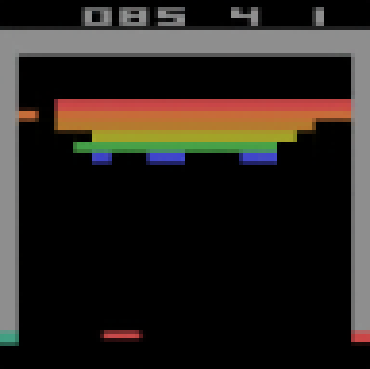} &
        \includegraphics[height=0.090\textwidth, valign=c]{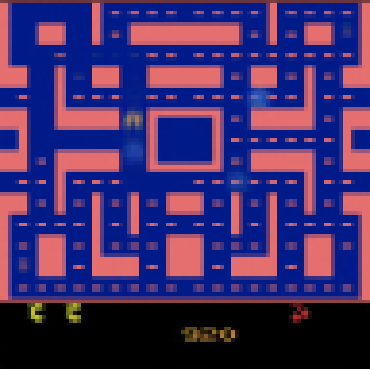} &
        \includegraphics[height=0.090\textwidth, valign=c]{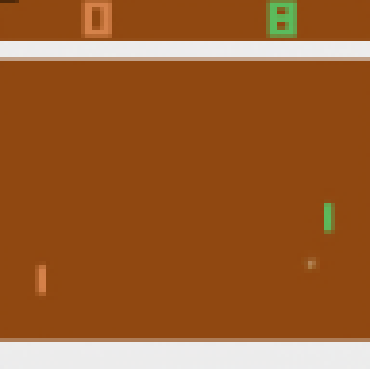} \vspace{0.3em} \\
        & \multicolumn{5}{c}{\footnotesize(a) \revision{Recent-training states}} \vspace{0.3em} \\

        \revision{\footnotesize$o_t$} &
        \includegraphics[height=0.090\textwidth, valign=c]{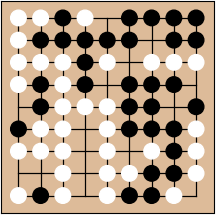} &
        \includegraphics[height=0.090\textwidth, valign=c]{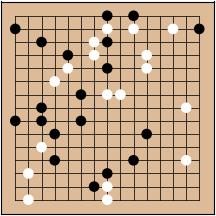} &
        \includegraphics[height=0.090\textwidth, valign=c]{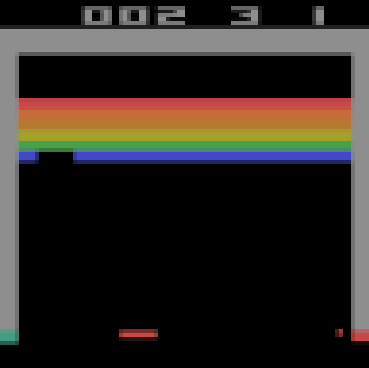} &
        \includegraphics[height=0.090\textwidth, valign=c]{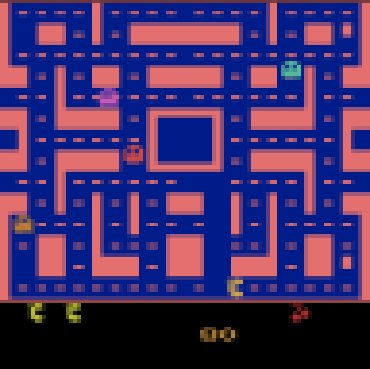} &
        \includegraphics[height=0.090\textwidth, valign=c]{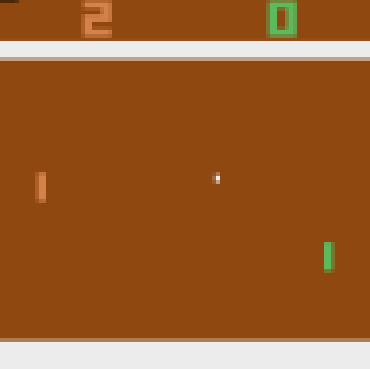} \vspace{0.2em} \\
        \revision{\footnotesize$\hat o_t$} &
        \includegraphics[height=0.090\textwidth, valign=c]{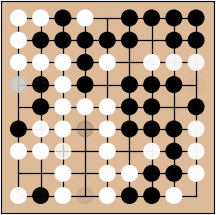} &
        \includegraphics[height=0.090\textwidth, valign=c]{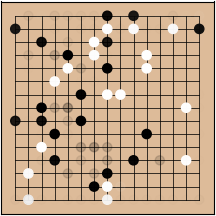} &
        \includegraphics[height=0.090\textwidth, valign=c]{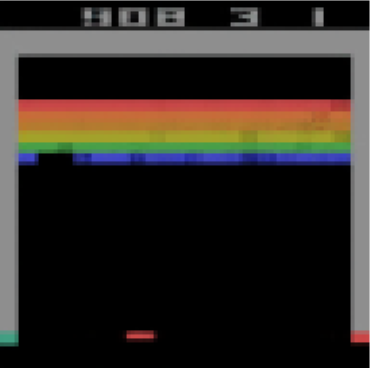} &
        \includegraphics[height=0.090\textwidth, valign=c]{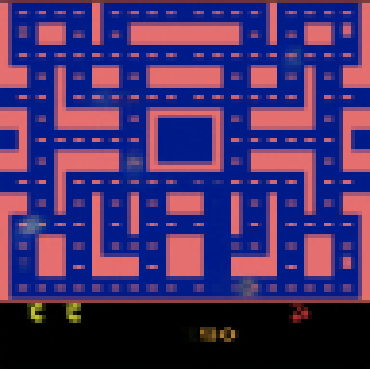} &
        \includegraphics[height=0.090\textwidth, valign=c]{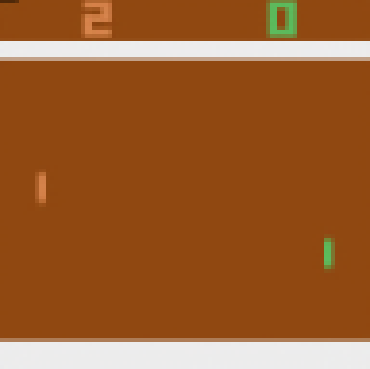} \vspace{0.3em} \\
        & \multicolumn{5}{c}{\footnotesize(b) \revision{Early-training states}} \\
    \end{tabular}}
    \caption{The comparison between true and decoded observations for \revision{recent-training states} and \revision{early-training} states across five games. The games, from left to right, are Go, Gomoku, Breakout, Ms. Pacman, and Pong.}
    \label{fig:decoder-performance-all-games}
\end{figure}

We first investigate whether the hidden states generated by the representation network can be accurately decoded into observations.
Specifically, we compare the decoded observations $\hat o_t$ with the true observations $o_t$ to evaluate performance in both \revision{\textit{recent-training states}} and \revision{\textit{early-training states}} for each game, as shown in Fig. \ref{fig:decoder-performance-all-games}.
\revision{As we use the latest model for analysis, recent-training states are from the final self-play iterations, whereas early-training states are from the initial iterations and are considered out-of-distribution states.}
For \revision{recent-training} states, the stones in board games are accurate, and the object outlines in Atari games generally match, with minor errors like a missing ball in Breakout or blurry ghosts in Ms. Pacman.
On the other hand, the decoded observations perform worse in \revision{early-training} states.
For example, several blurry stones are incorrectly shown in board games, and many common objects are blurry or even missing in Atari games, such as the bricks in Breakout, ghosts in Ms. Pacman, and the ball in Pong.
We conclude that the decoder network performs better in board games, possibly due to the challenge of learning complex Atari images, and also performs better in familiar \revision{recent-training} states.

\subsection{Decoding unrolled hidden states into observations}\label{sec:interpreting-hidden-states:unrolled}

\begin{figure}[t]
    \centering
    \small
    {\setlength{\tabcolsep}{0.1em}
    \begin{tabular}[]{cccccc}
        & \revision{\footnotesize$k=0$} & \revision{\footnotesize$k=1$} & \revision{\footnotesize$k=5$} & \revision{\footnotesize$k=10$} & \revision{\footnotesize$k=20$} \vspace{0em} \\
        \revision{\footnotesize$o_{t+k}$} &
        \includegraphics[height=0.087\textwidth, valign=c]{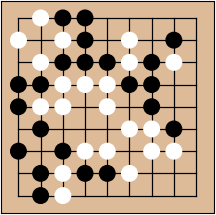} &
        \includegraphics[height=0.087\textwidth, valign=c]{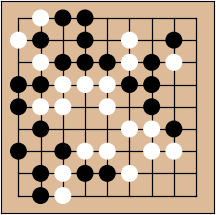} &
        \includegraphics[height=0.087\textwidth, valign=c]{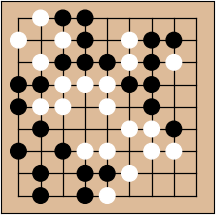} &
        \includegraphics[height=0.087\textwidth, valign=c]{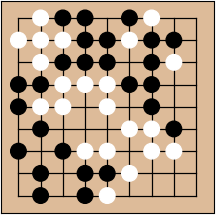} &
        \includegraphics[height=0.087\textwidth, valign=c]{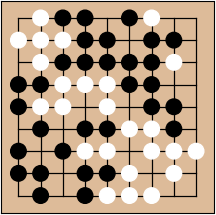} \vspace{0.2em} \\
        \revision{\footnotesize$\hat o_{t+k}$} & 
        \includegraphics[height=0.087\textwidth, valign=c]{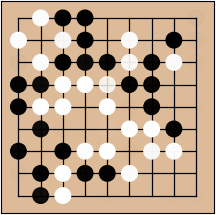} &
        \includegraphics[height=0.087\textwidth, valign=c]{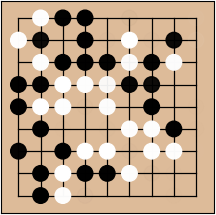} &
        \includegraphics[height=0.087\textwidth, valign=c]{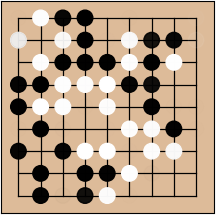} &
        \includegraphics[height=0.087\textwidth, valign=c]{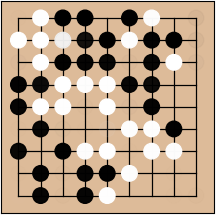} &
        \includegraphics[height=0.087\textwidth, valign=c]{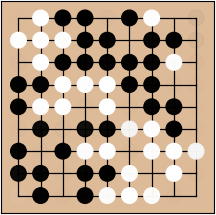} \vspace{0.2em} \\
        \revision{\footnotesize$\hat o_{t+k}^{(k)}$} &
        \includegraphics[height=0.087\textwidth, valign=c]{figures/unroll-go/representation_46.png} &
        \includegraphics[height=0.087\textwidth, valign=c]{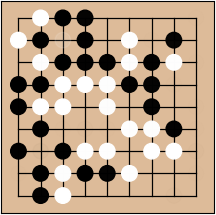} &
        \includegraphics[height=0.087\textwidth, valign=c]{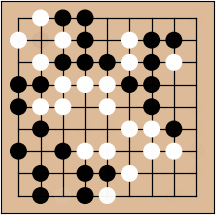} &
        \includegraphics[height=0.087\textwidth, valign=c]{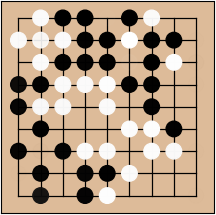} &
        \includegraphics[height=0.087\textwidth, valign=c]{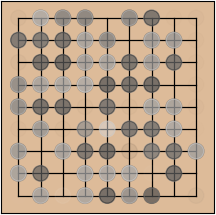} \vspace{0.3em} \\
        & \multicolumn{5}{c}{\footnotesize(a) Go} \vspace{0.3em} \\

        \revision{\footnotesize$o_{t+k}$} &
        \includegraphics[height=0.087\textwidth, valign=c]{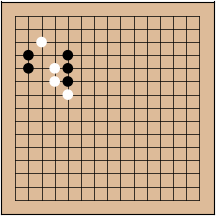} &
        \includegraphics[height=0.087\textwidth, valign=c]{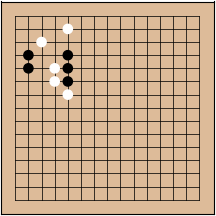} &
        \includegraphics[height=0.087\textwidth, valign=c]{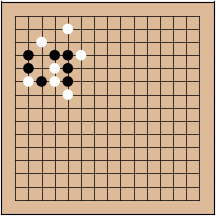} &
        \includegraphics[height=0.087\textwidth, valign=c]{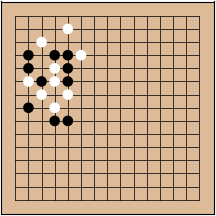} &
        \includegraphics[height=0.087\textwidth, valign=c]{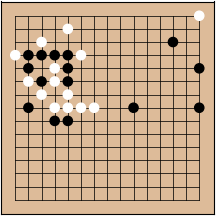} \vspace{0.2em} \\
        \revision{\footnotesize$\hat o_{t+k}$} & 
        \includegraphics[height=0.087\textwidth, valign=c]{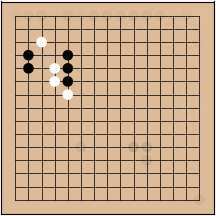} &
        \includegraphics[height=0.087\textwidth, valign=c]{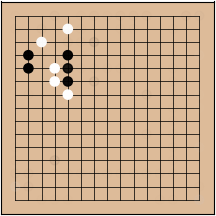} &
        \includegraphics[height=0.087\textwidth, valign=c]{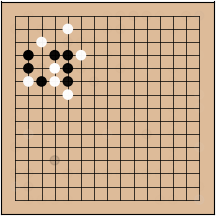} &
        \includegraphics[height=0.087\textwidth, valign=c]{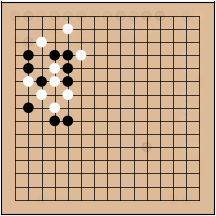} &
        \includegraphics[height=0.087\textwidth, valign=c]{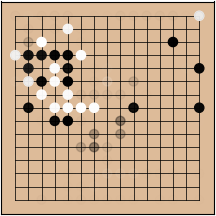} \vspace{0.2em} \\
        \revision{\footnotesize$\hat o_{t+k}^{(k)}$} &
        \includegraphics[height=0.087\textwidth, valign=c]{figures/unroll-gomoku/representation_9.png} &
        \includegraphics[height=0.087\textwidth, valign=c]{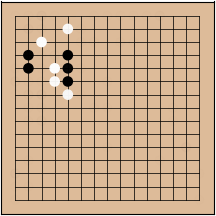} &
        \includegraphics[height=0.087\textwidth, valign=c]{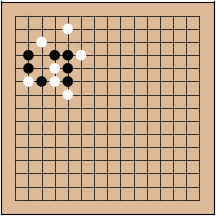} &
        \includegraphics[height=0.087\textwidth, valign=c]{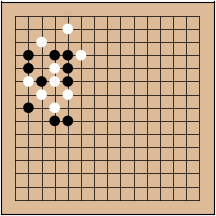} &
        \includegraphics[height=0.087\textwidth, valign=c]{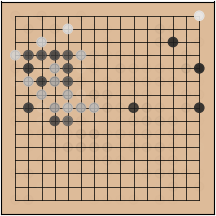} \vspace{0.3em} \\
        & \multicolumn{5}{c}{\footnotesize(b) Gomoku} \vspace{0em} \\
    \end{tabular}}
    \caption{The comparison between true and decoded observations at different unrolling steps in Go and Gomoku.}
    \label{fig:unroll-board-games}
\end{figure}

\begin{figure}[t]
    \centering
    \small
    {\setlength{\tabcolsep}{0.1em}
    \begin{tabular}[]{cccccc}
        & \revision{\footnotesize$k=0$} & \revision{\footnotesize$k=1$} & \revision{\footnotesize$k=5$} & \revision{\footnotesize$k=10$} & \revision{\footnotesize$k=20$} \vspace{0em} \\

        \revision{\footnotesize$o_{t+k}$} &
        \includegraphics[height=0.087\textwidth, valign=c]{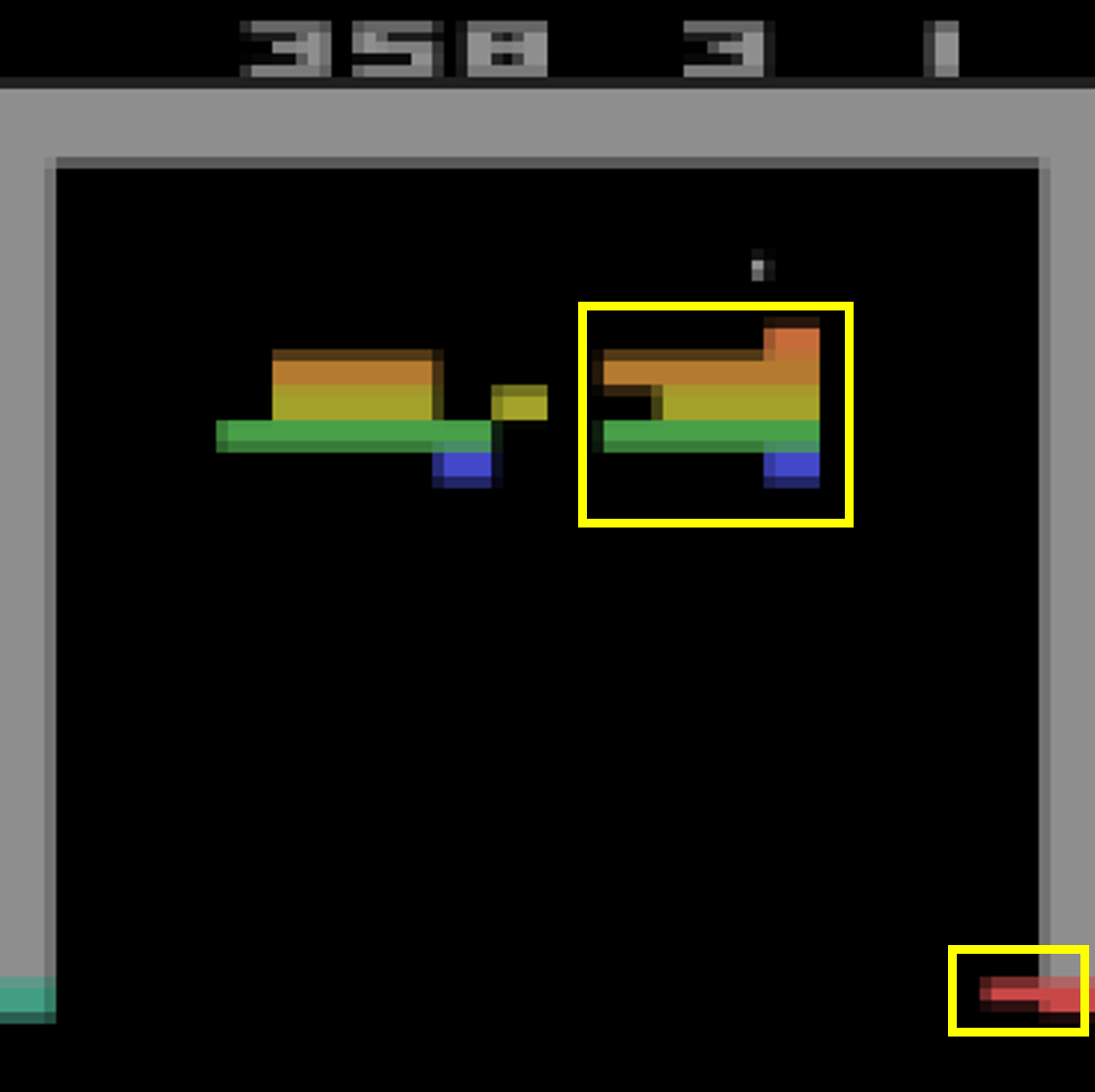} &
        \includegraphics[height=0.087\textwidth, valign=c]{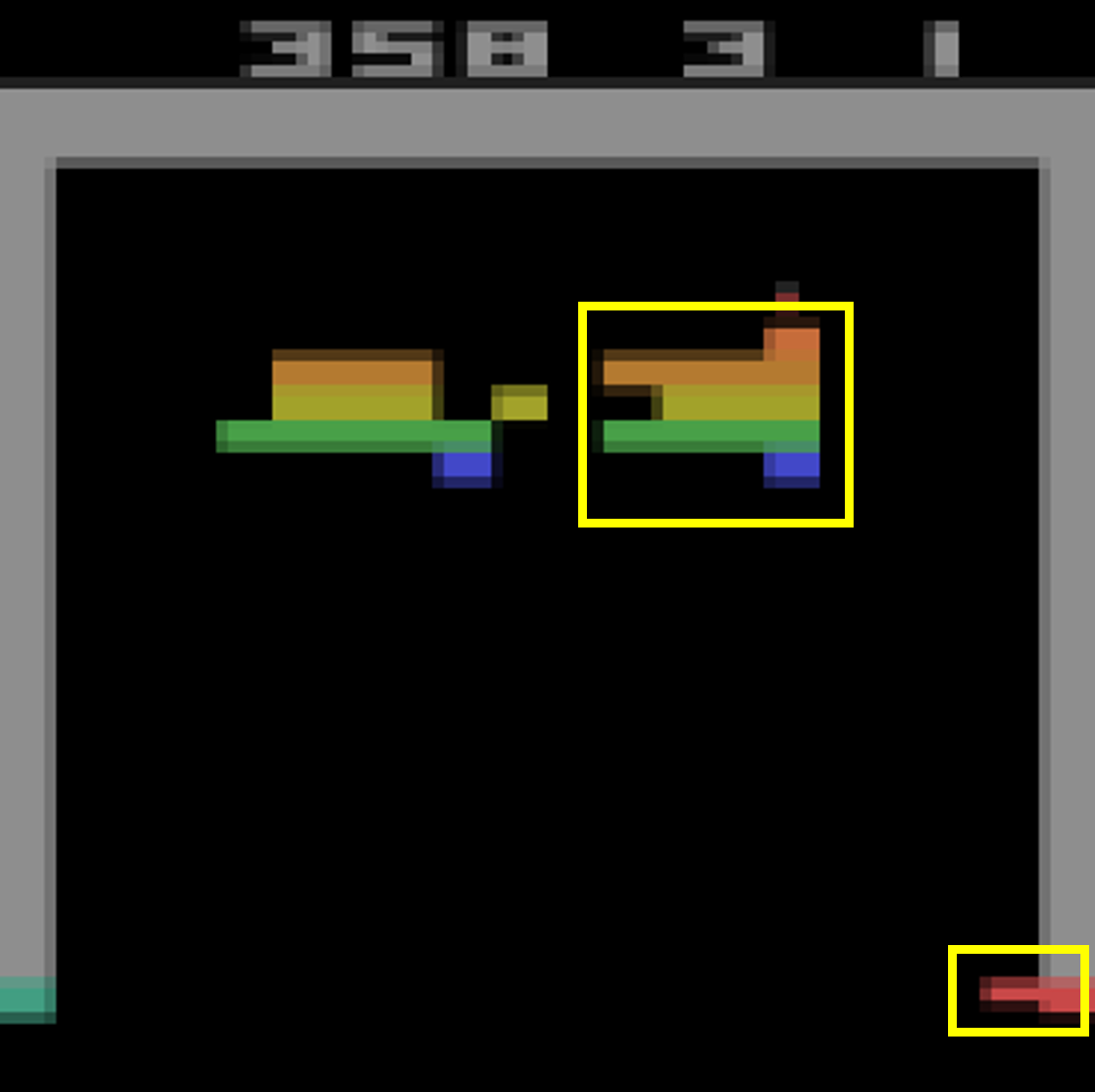} &
        \includegraphics[height=0.087\textwidth, valign=c]{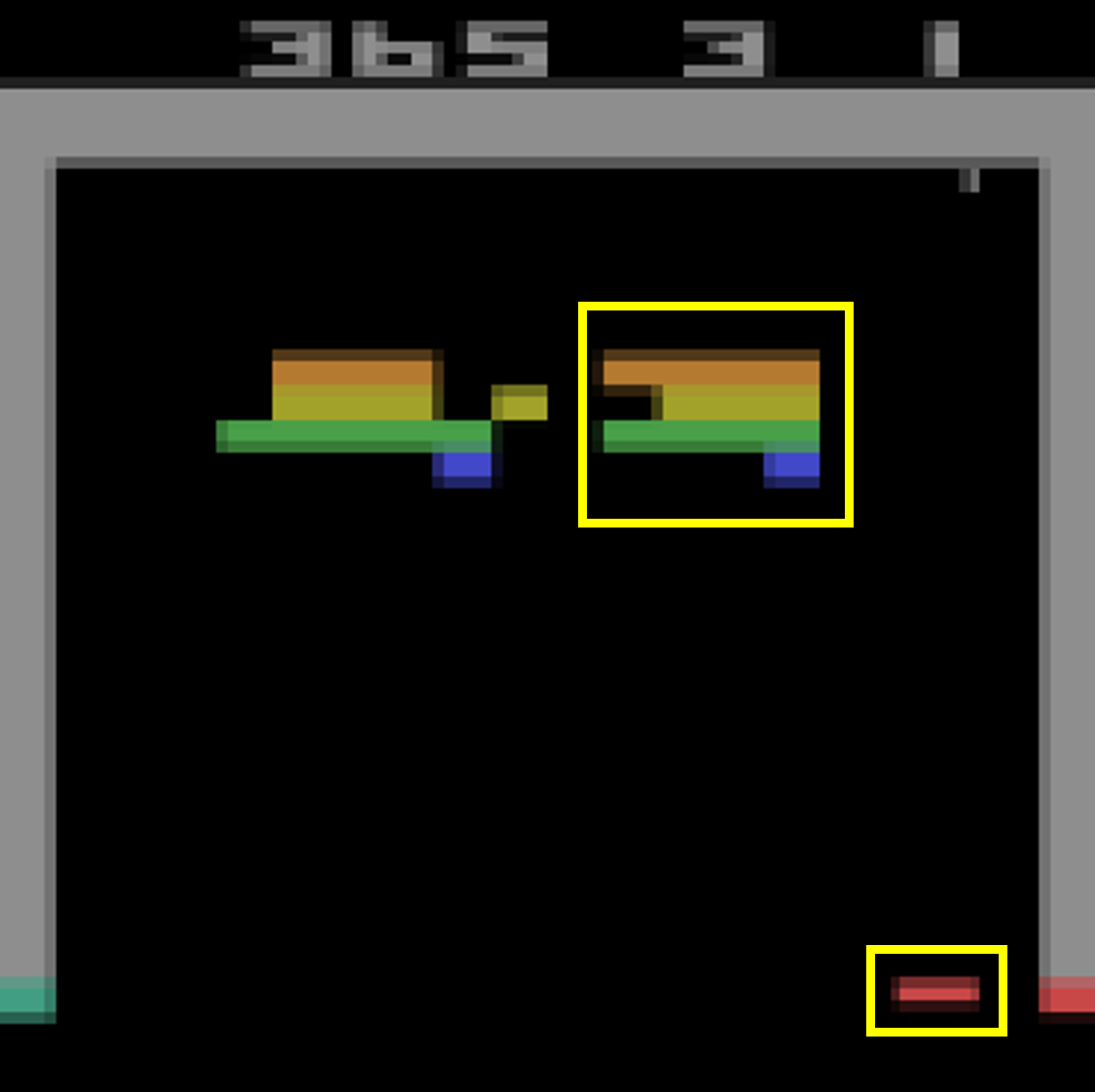} &
        \includegraphics[height=0.087\textwidth, valign=c]{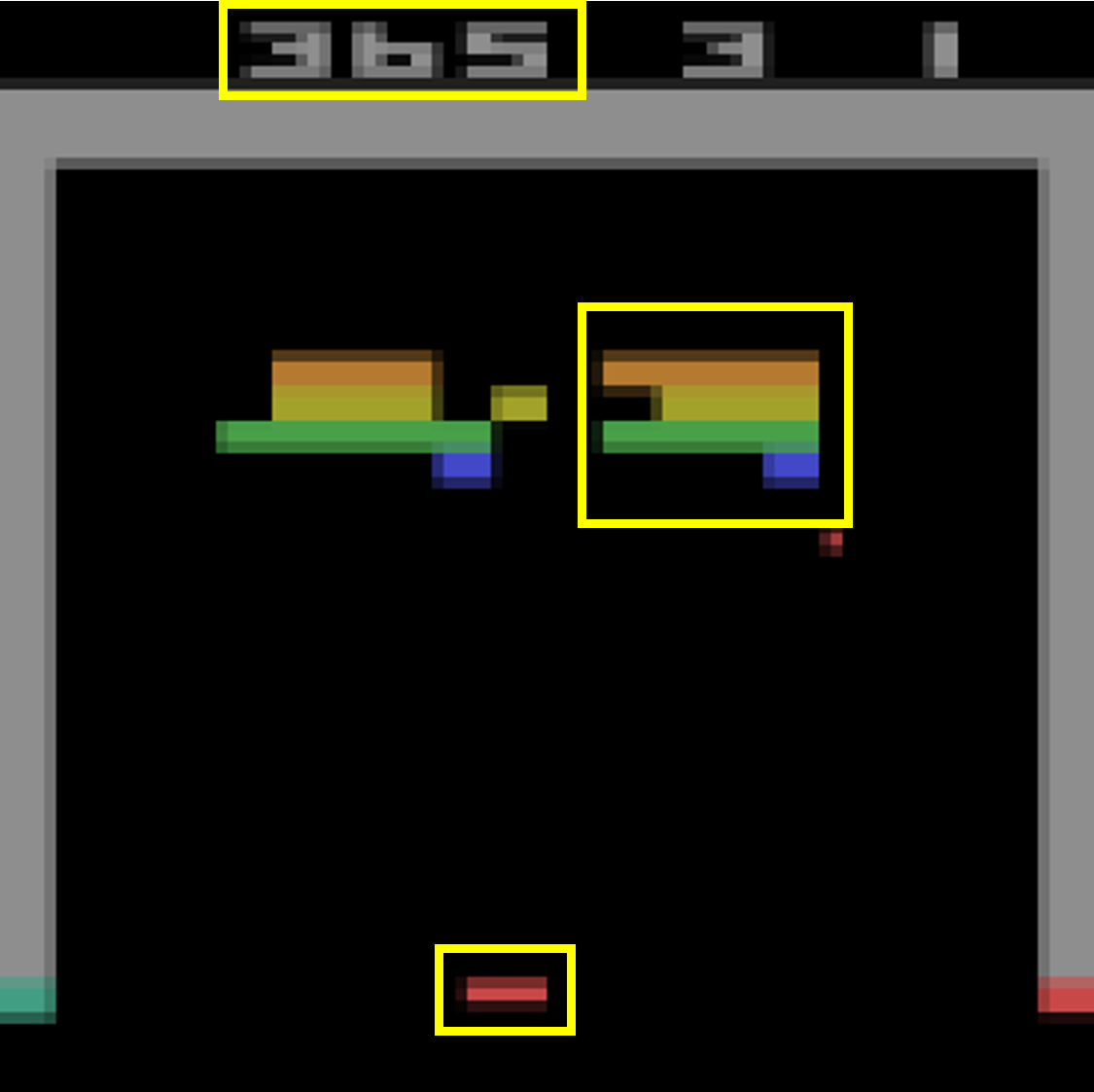} &
        \includegraphics[height=0.087\textwidth, valign=c]{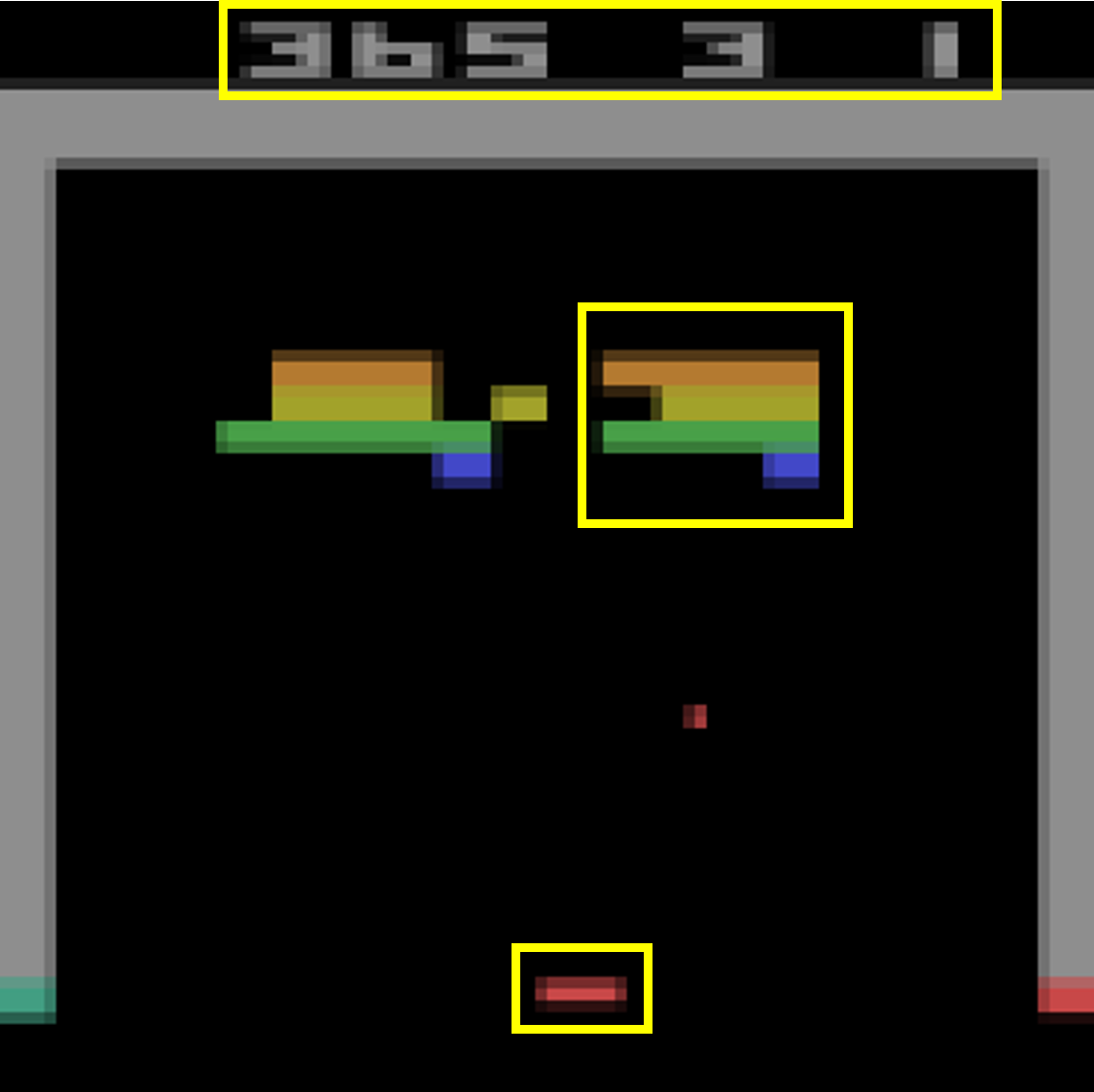} \vspace{0.2em} \\
        \revision{\footnotesize$\hat o_{t+k}$} & 
        \includegraphics[height=0.087\textwidth, valign=c]{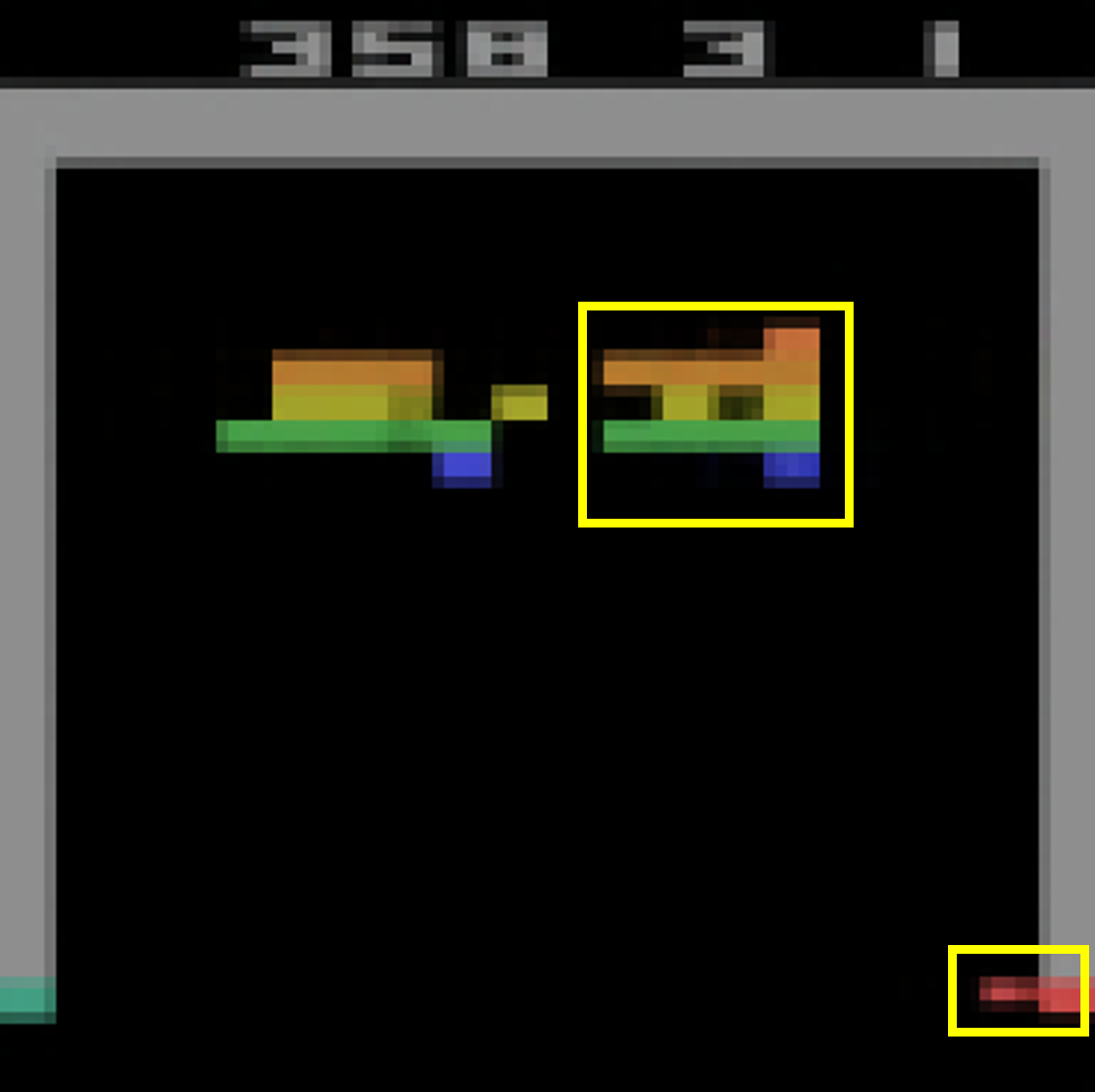} &
        \includegraphics[height=0.087\textwidth, valign=c]{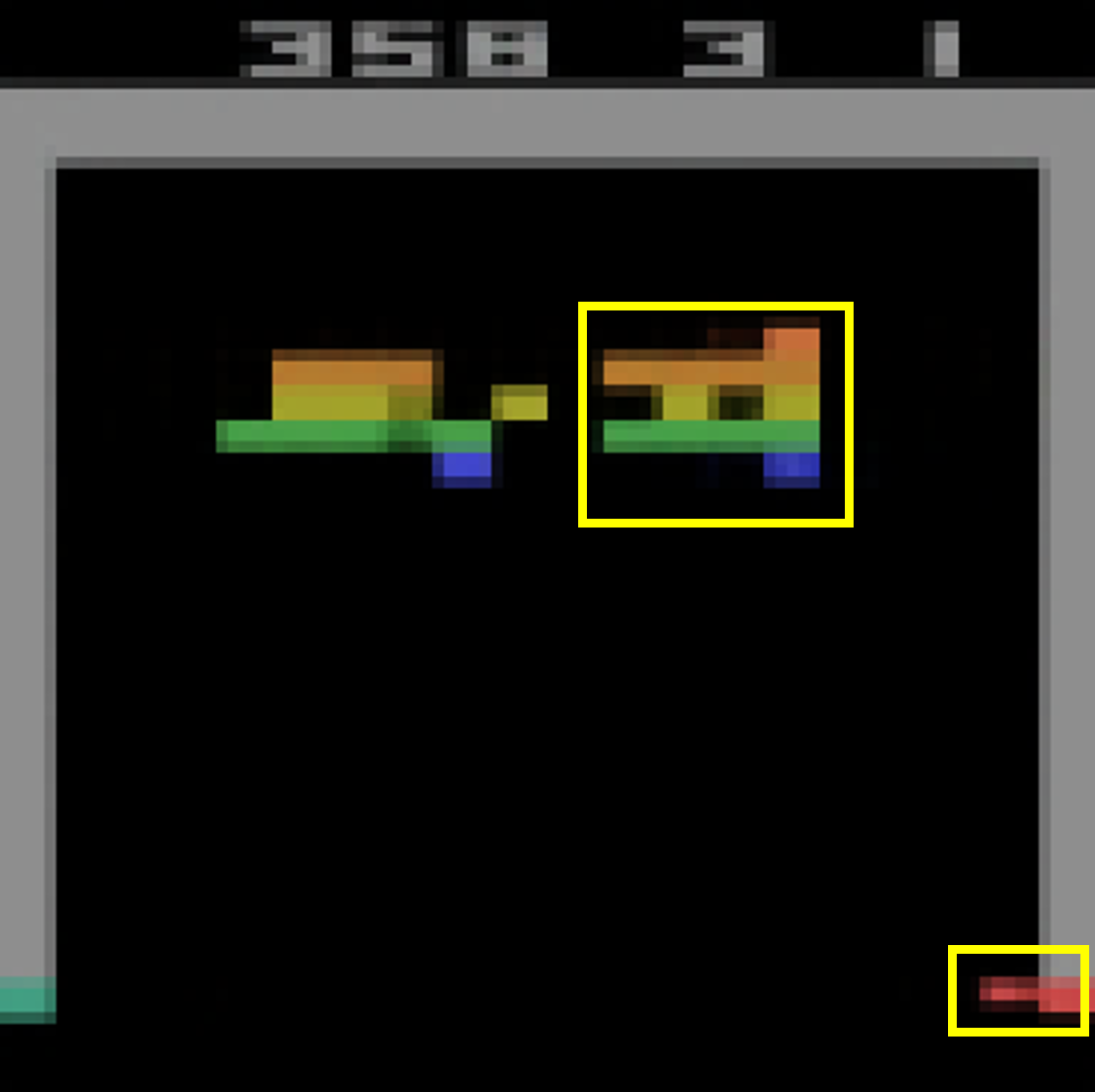} &
        \includegraphics[height=0.087\textwidth, valign=c]{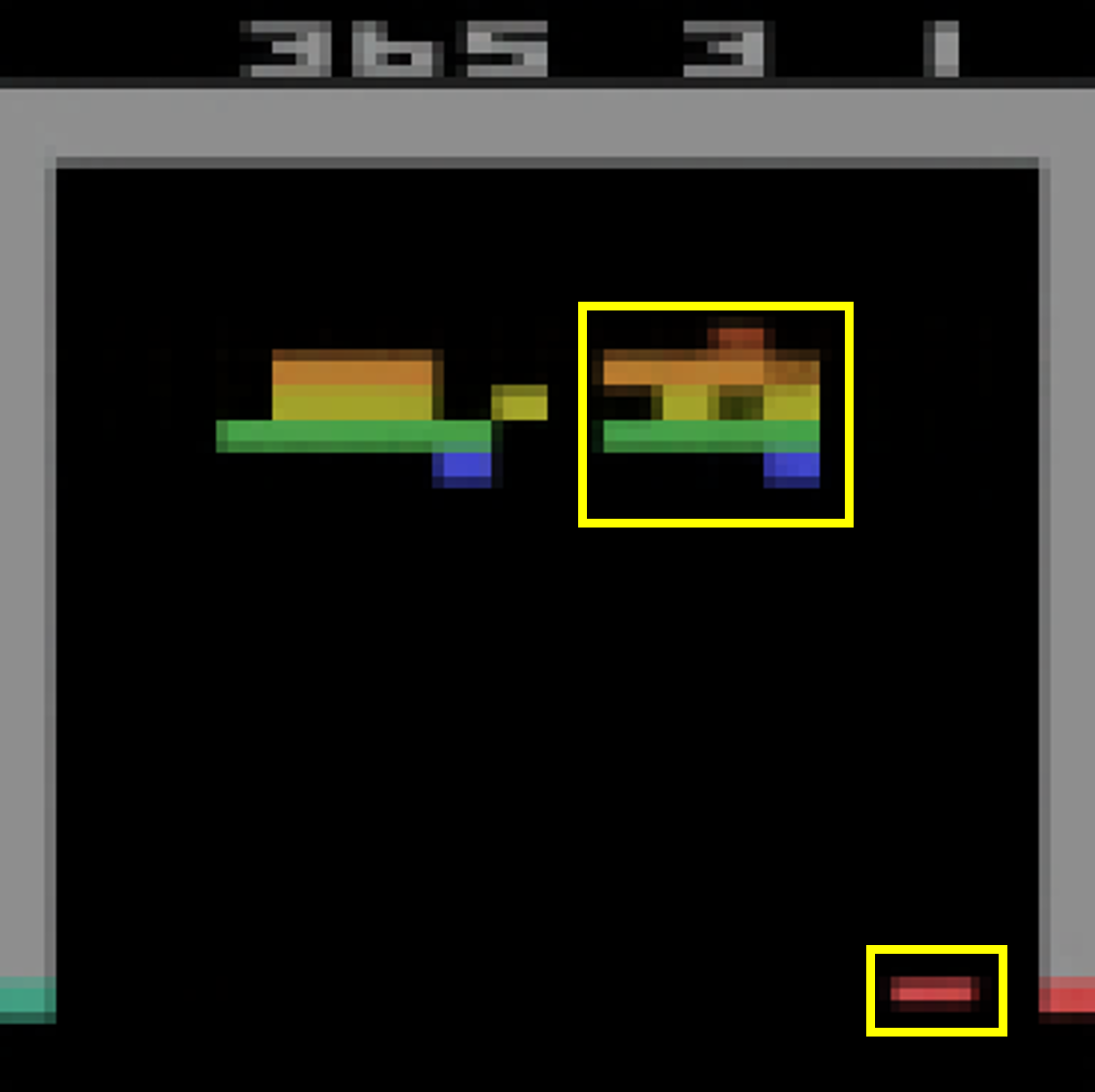} &
        \includegraphics[height=0.087\textwidth, valign=c]{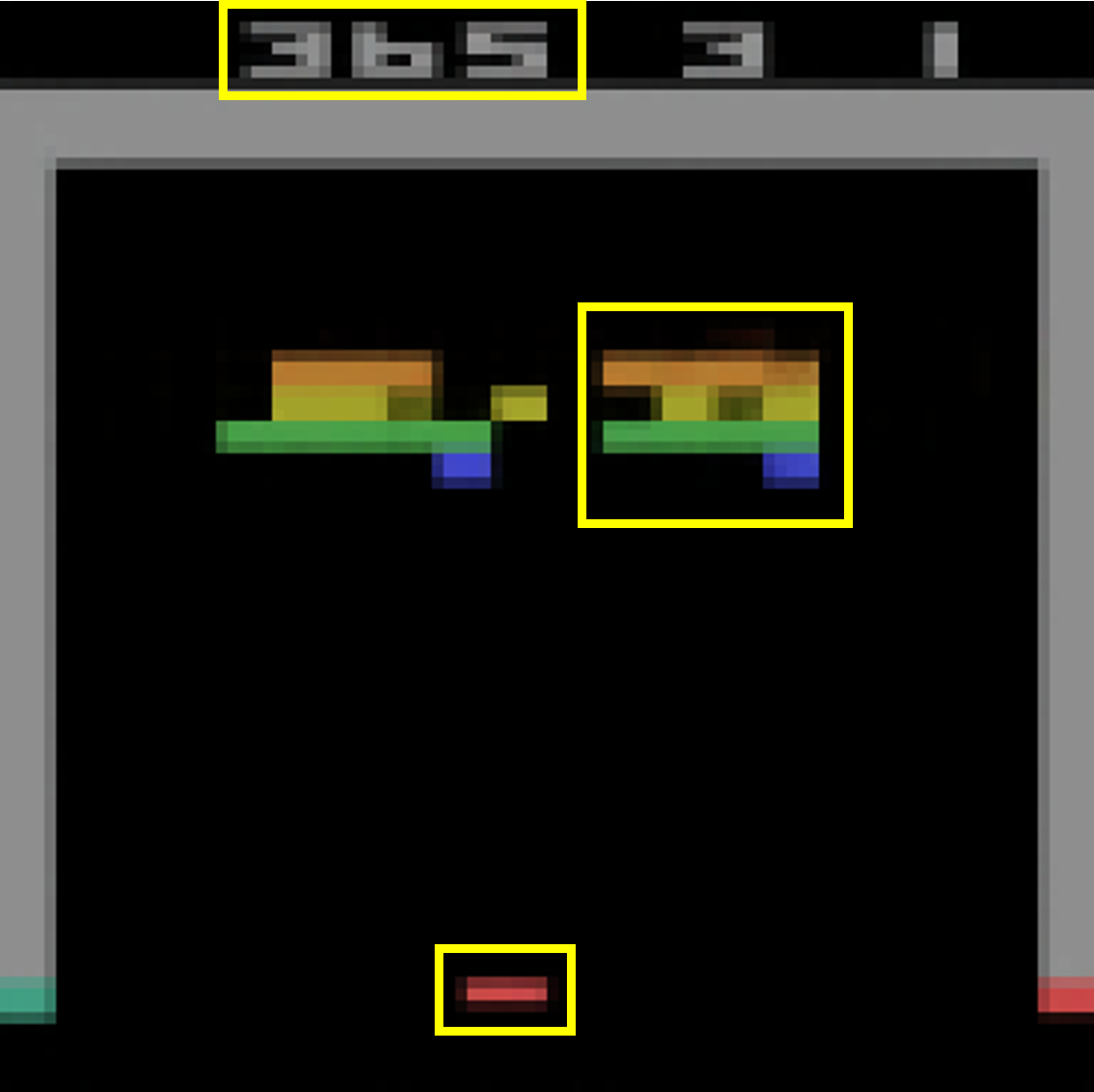} &
        \includegraphics[height=0.087\textwidth, valign=c]{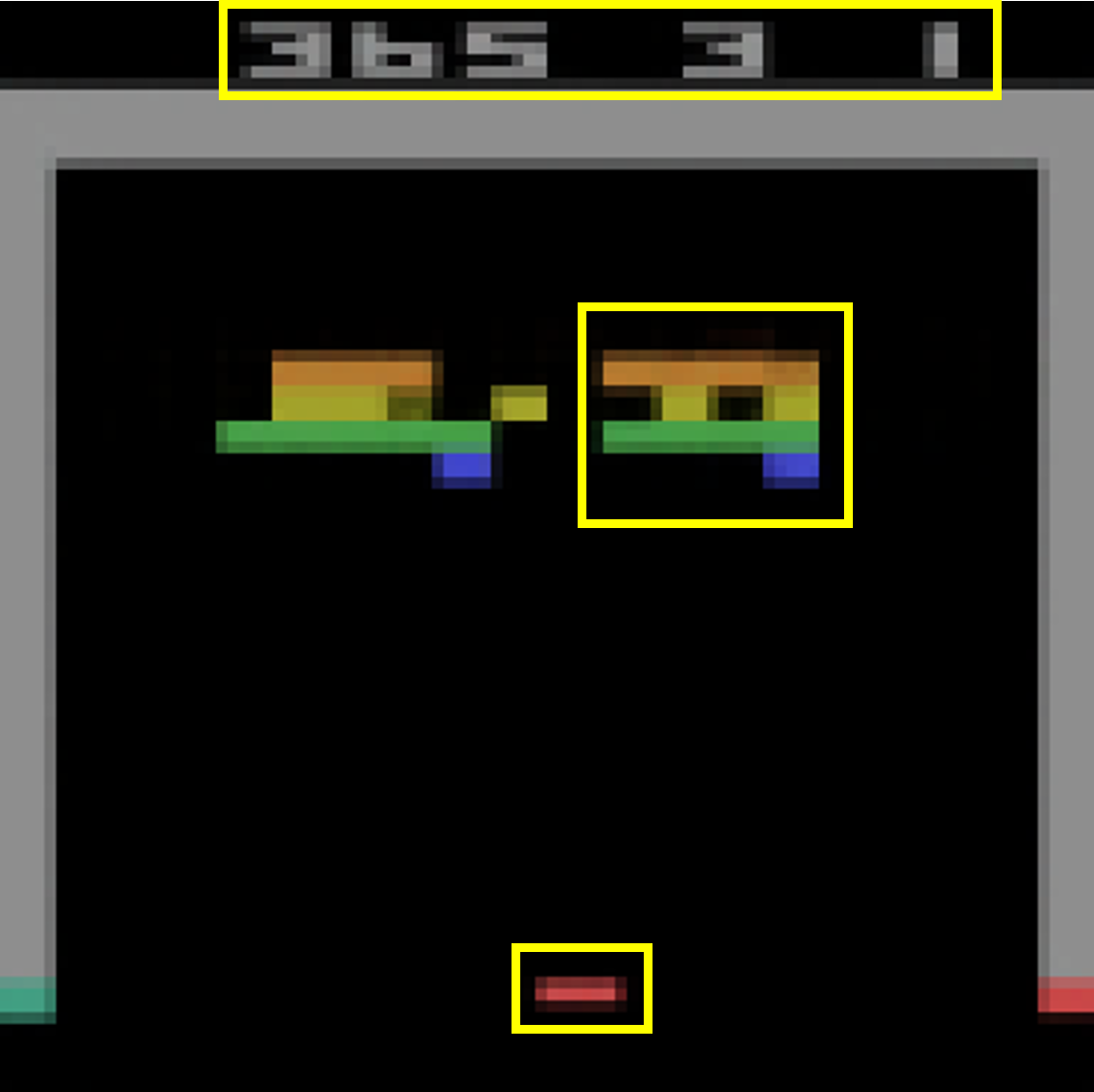} \vspace{0.2em} \\
        \revision{\footnotesize$\hat o_{t+k}^{(k)}$} &
        \includegraphics[height=0.087\textwidth, valign=c]{figures/unroll-breakout/representation_1481.png} &
        \includegraphics[height=0.087\textwidth, valign=c]{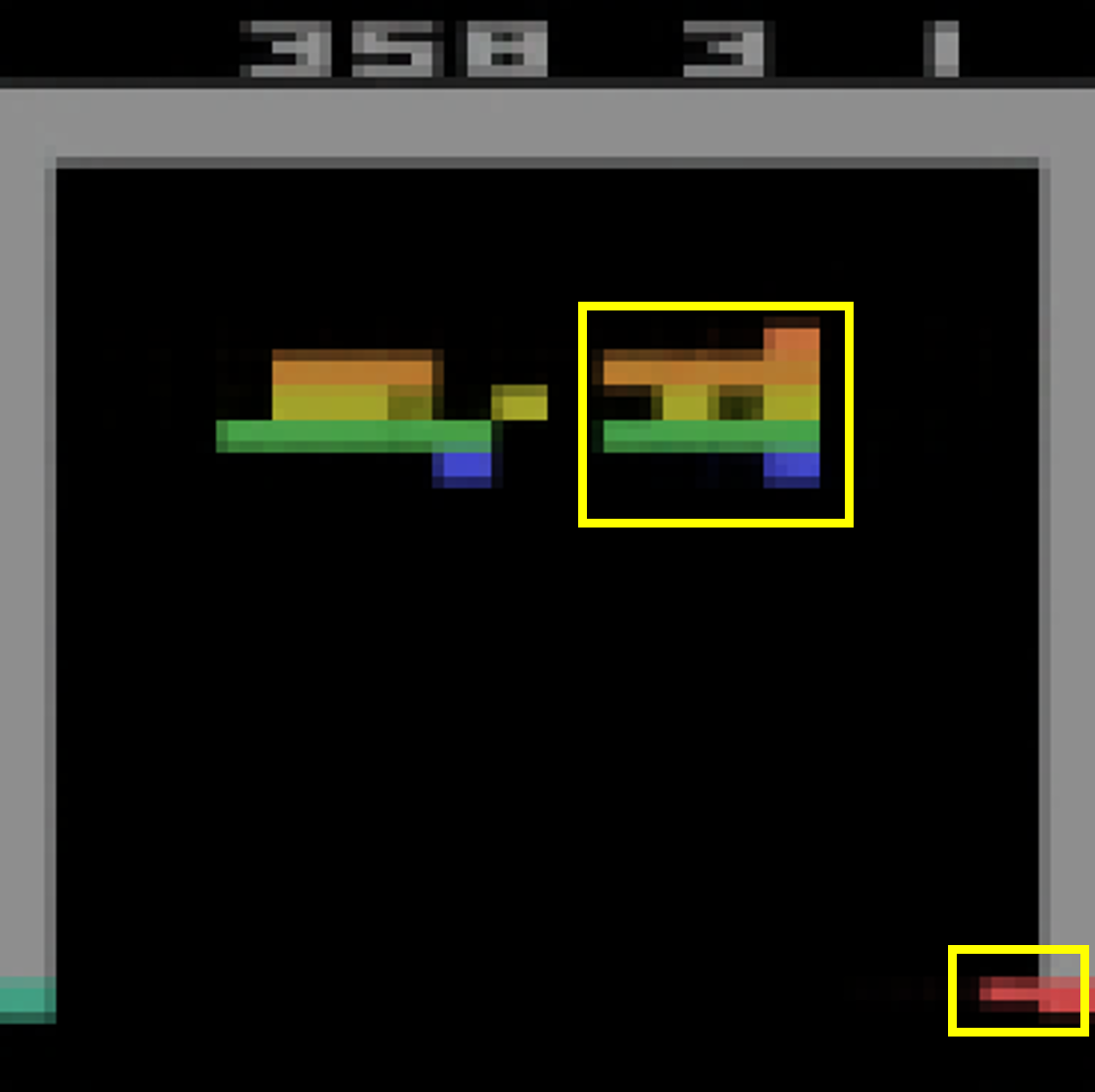} &
        \includegraphics[height=0.087\textwidth, valign=c]{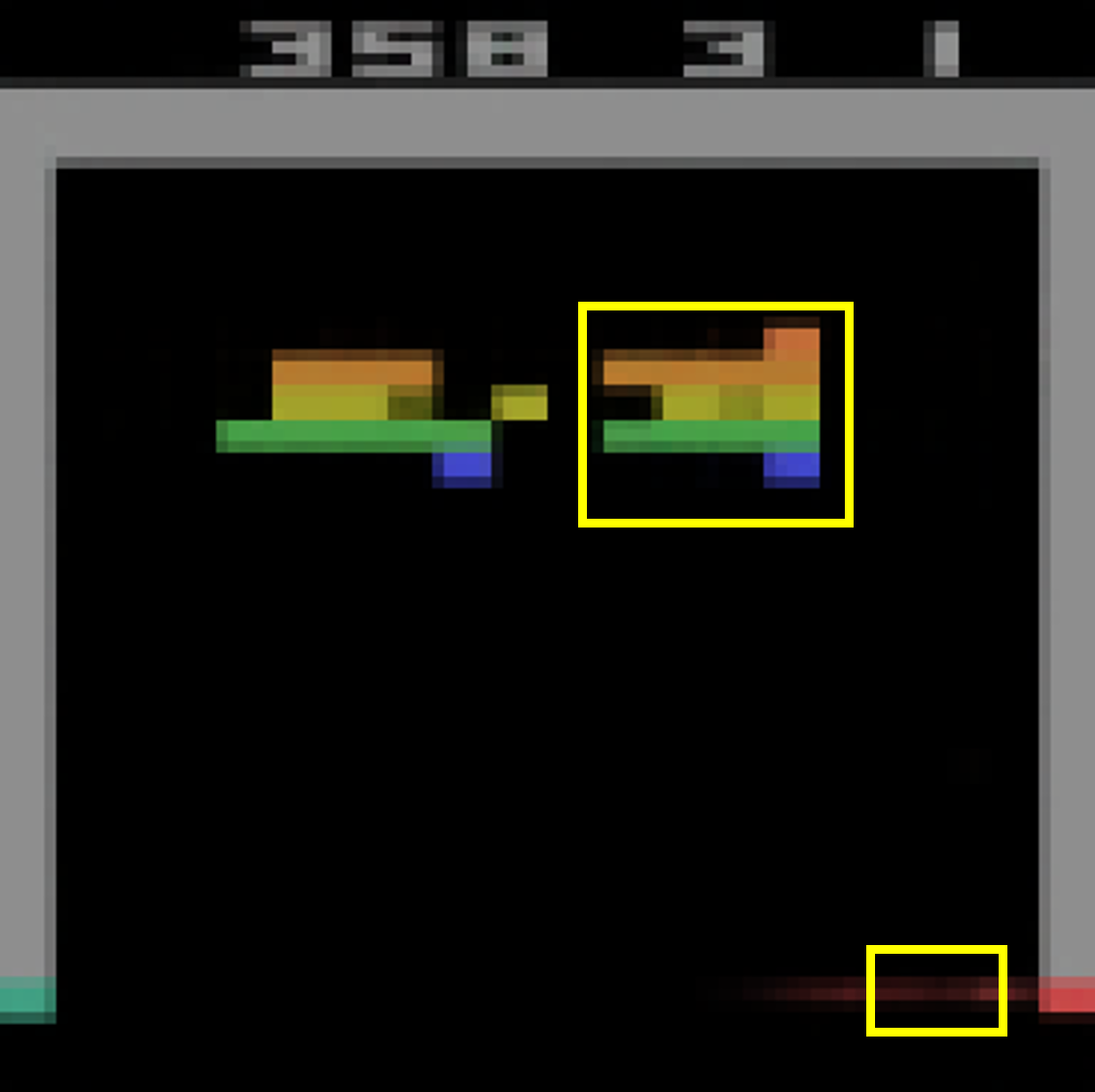} &
        \includegraphics[height=0.087\textwidth, valign=c]{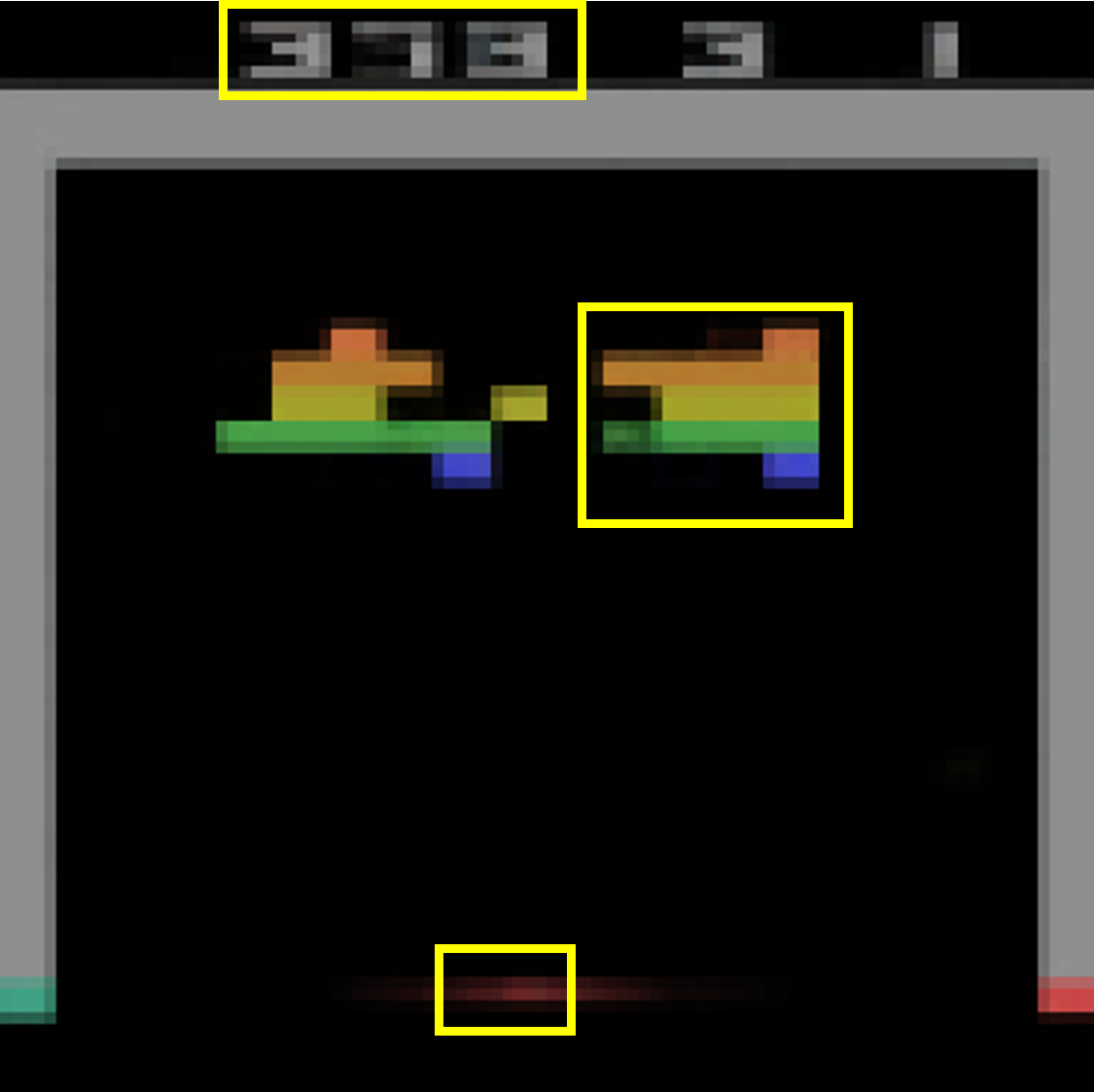} &
        \includegraphics[height=0.087\textwidth, valign=c]{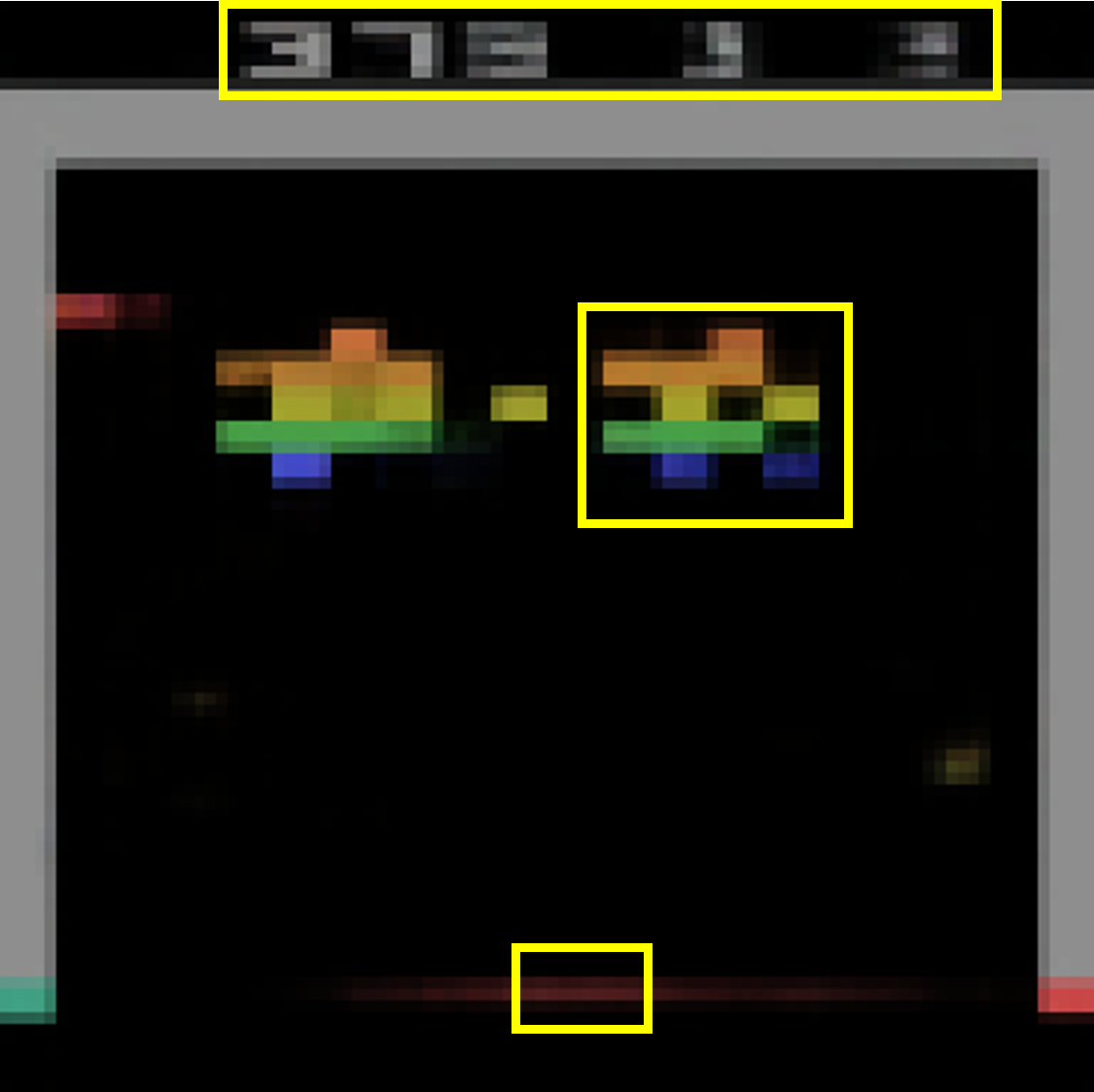} \vspace{0.3em} \\
        & \multicolumn{5}{c}{\footnotesize(a) Breakout} \vspace{0.3em} \\
        
        \revision{\footnotesize$o_{t+k}$} &
        \includegraphics[height=0.087\textwidth, valign=c]{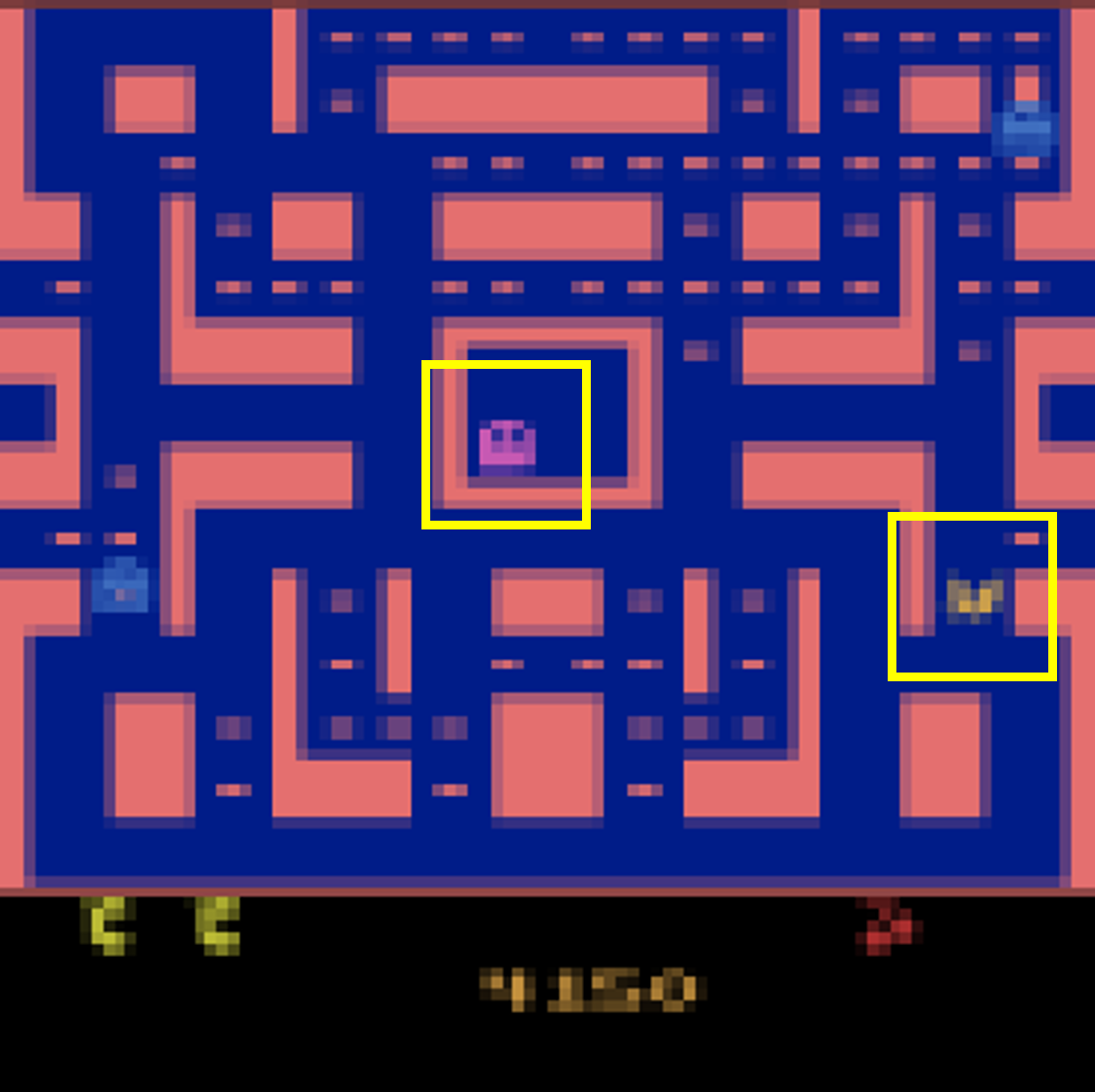} &
        \includegraphics[height=0.087\textwidth, valign=c]{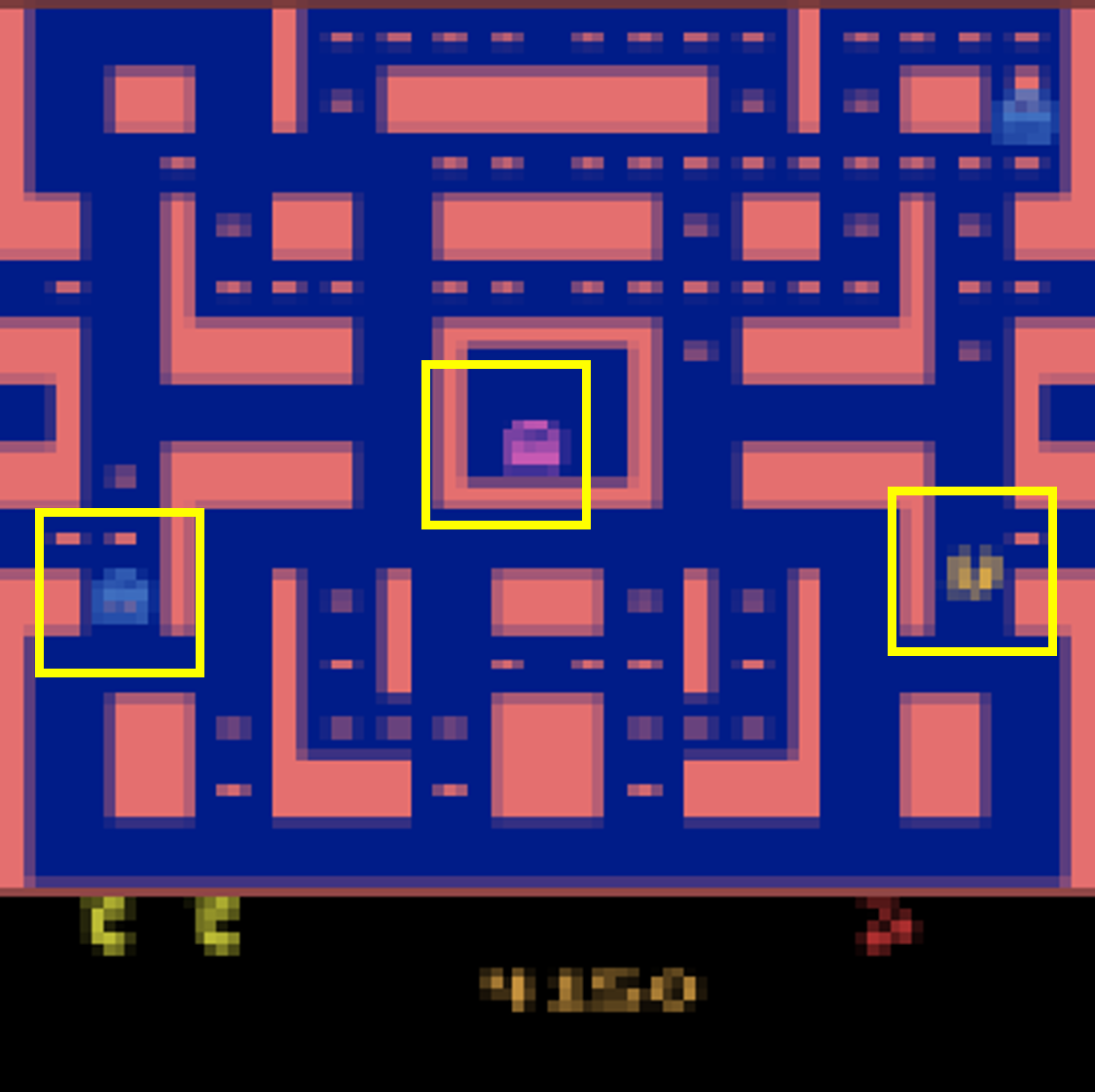} &
        \includegraphics[height=0.087\textwidth, valign=c]{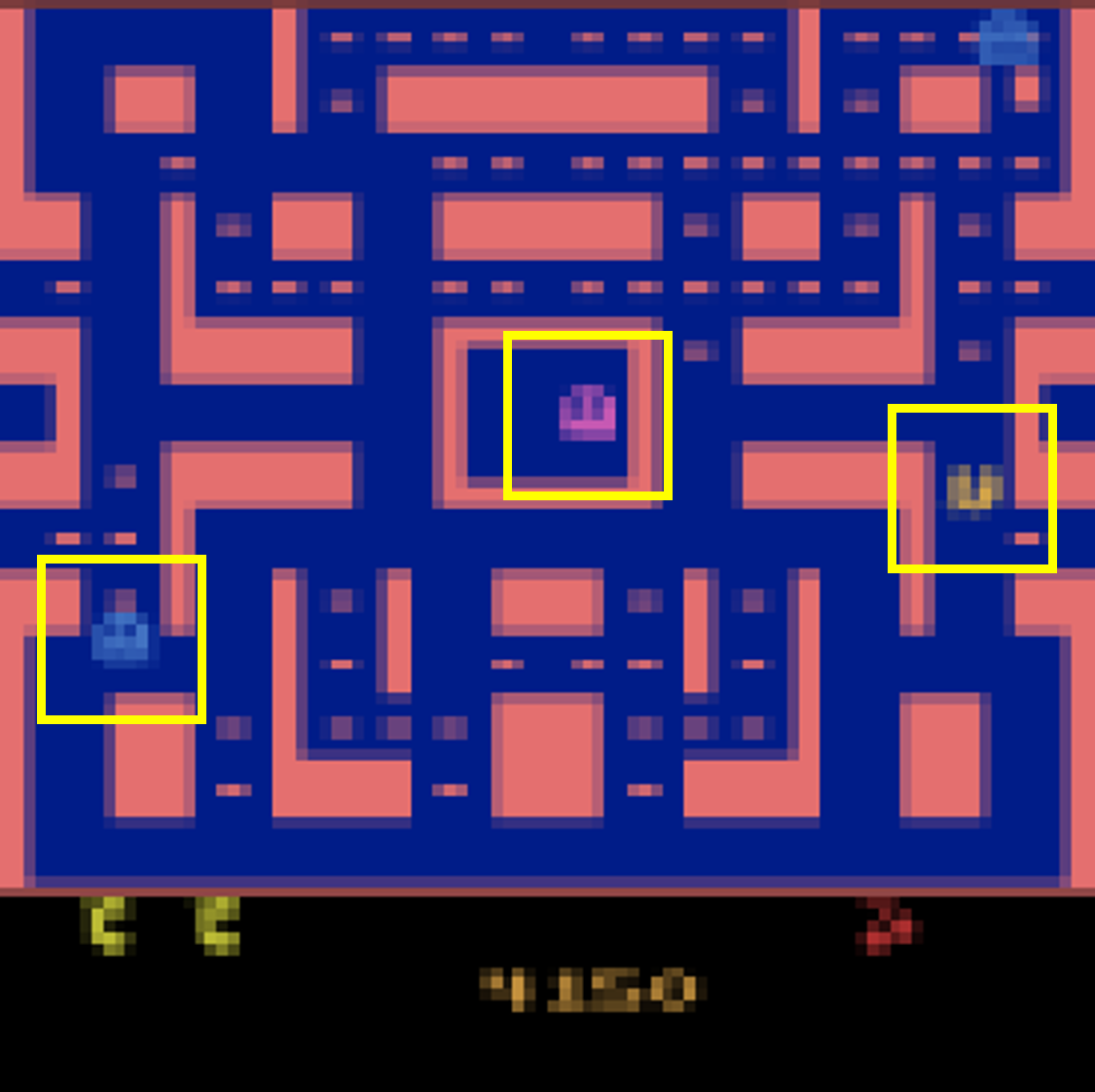} &
        \includegraphics[height=0.087\textwidth, valign=c]{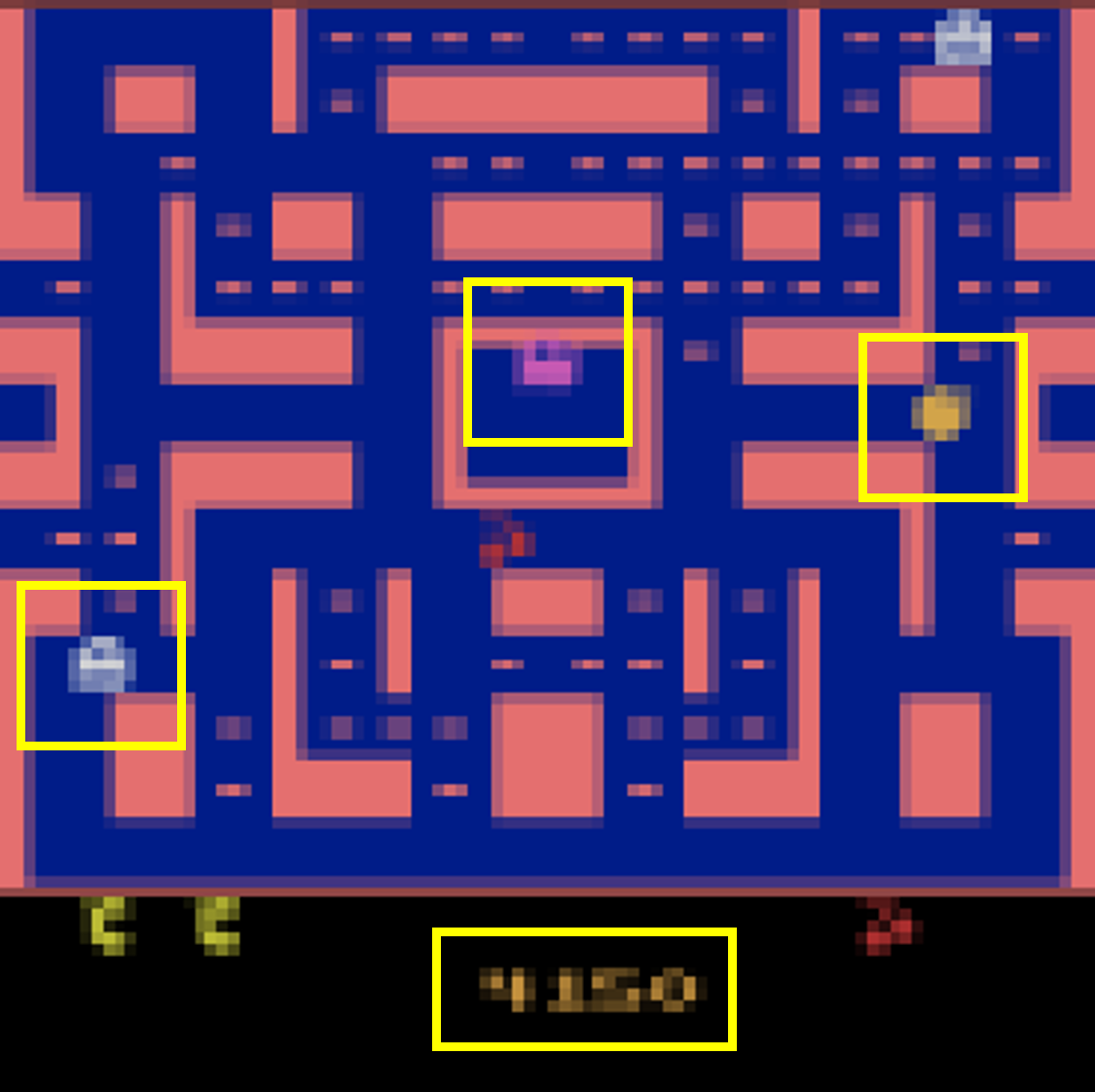} &
        \includegraphics[height=0.087\textwidth, valign=c]{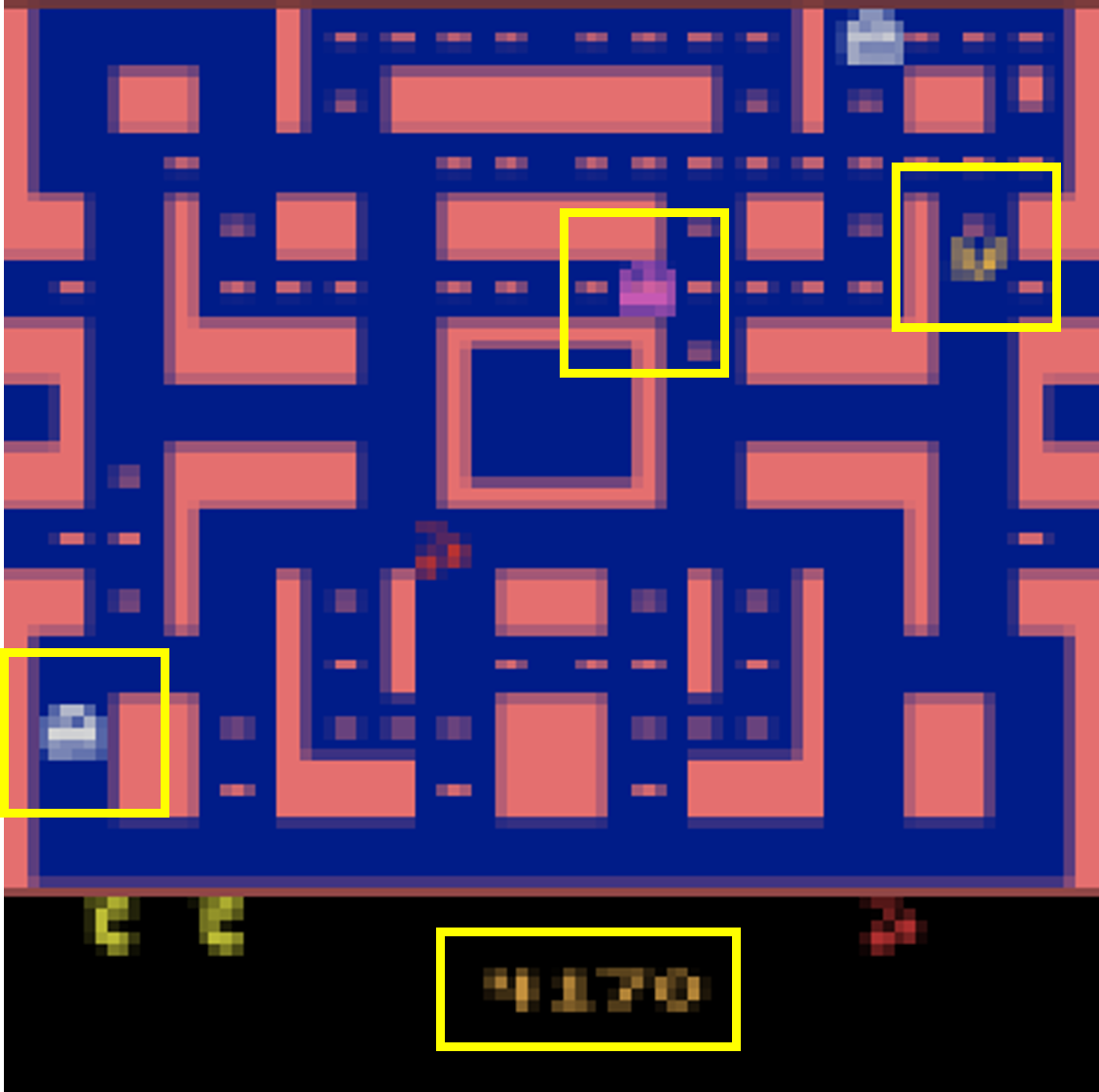} \vspace{0.2em} \\
        \revision{\footnotesize$\hat o_{t+k}$} & 
        \includegraphics[height=0.087\textwidth, valign=c]{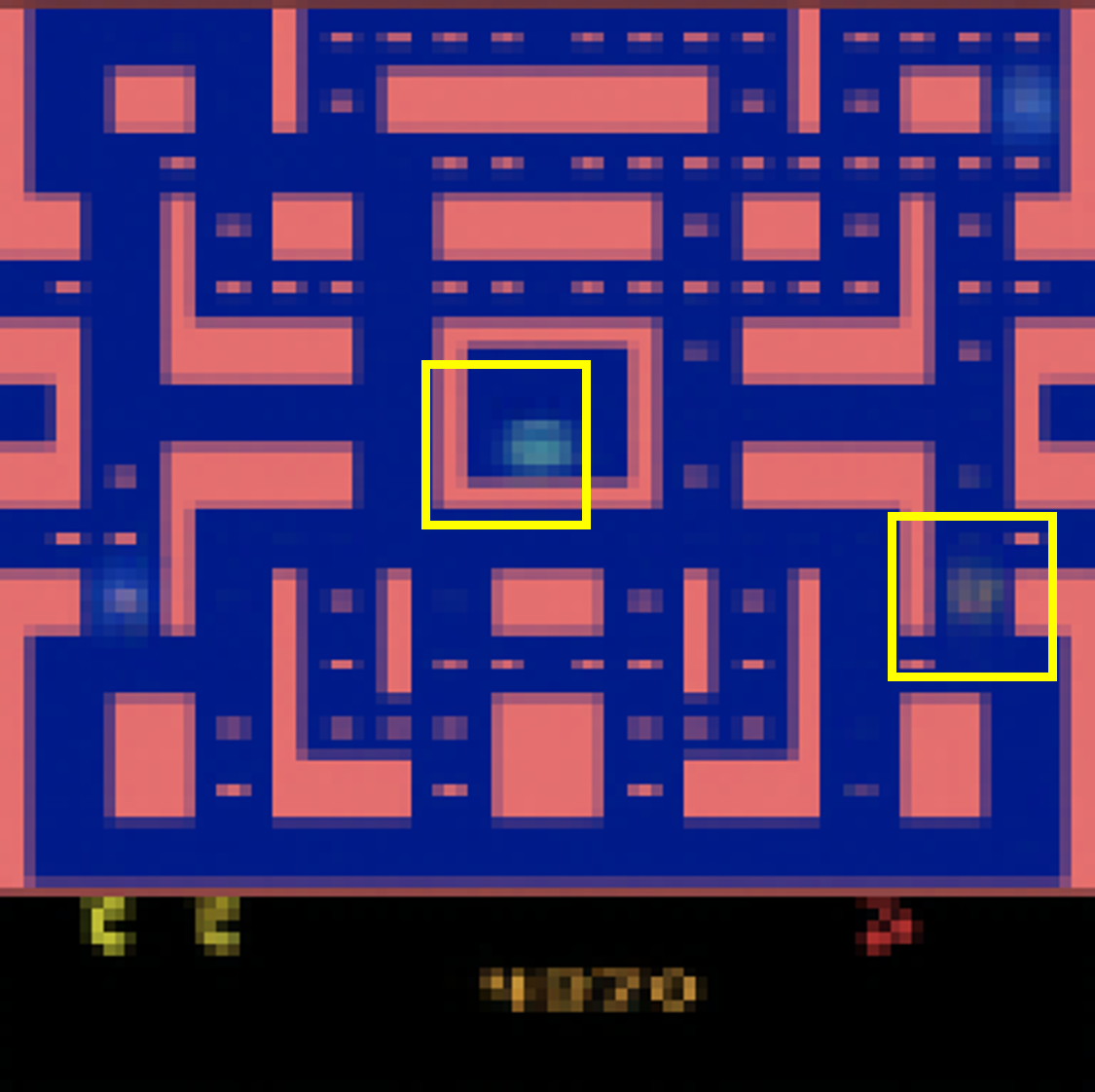} &
        \includegraphics[height=0.087\textwidth, valign=c]{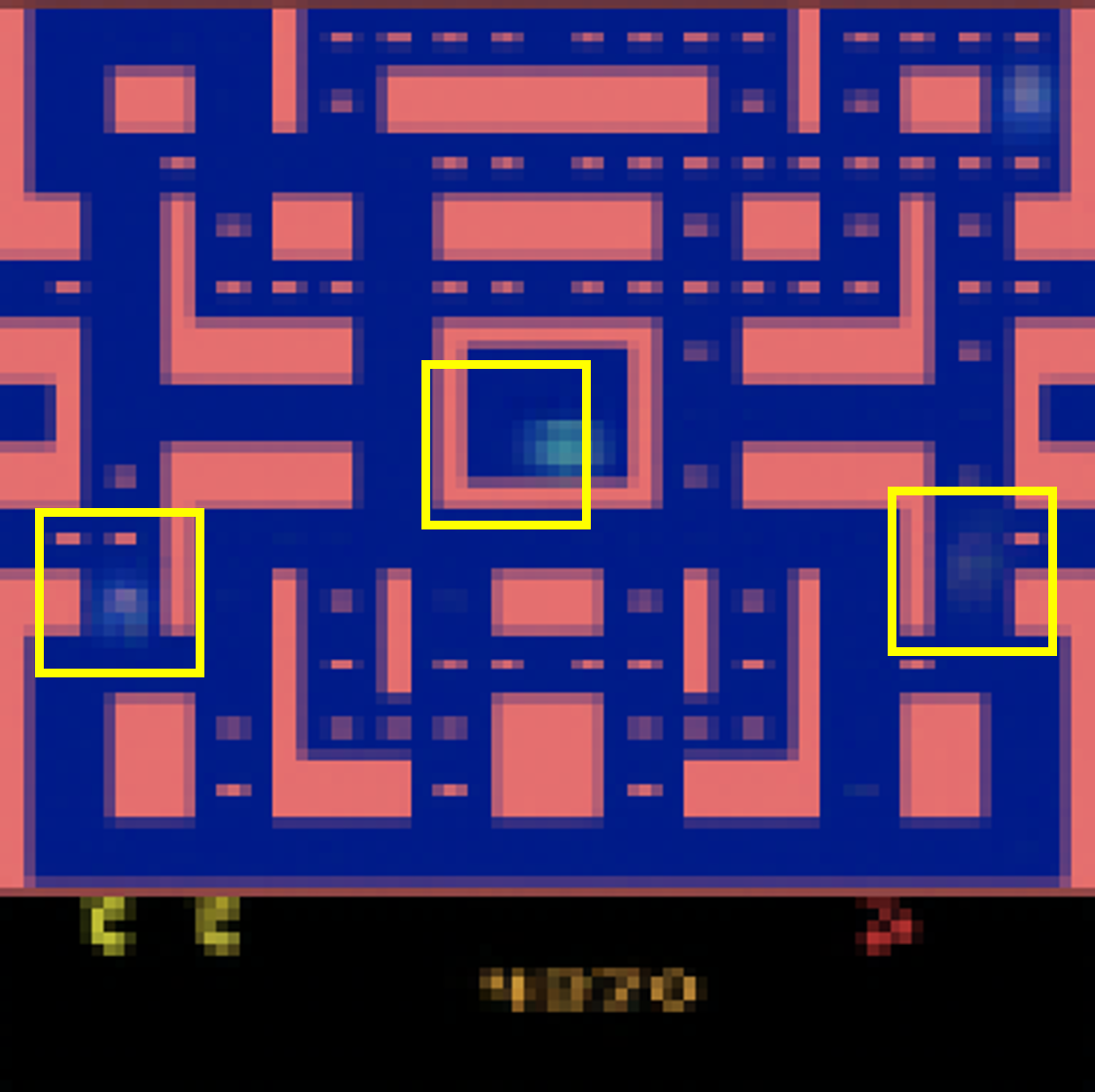} &
        \includegraphics[height=0.087\textwidth, valign=c]{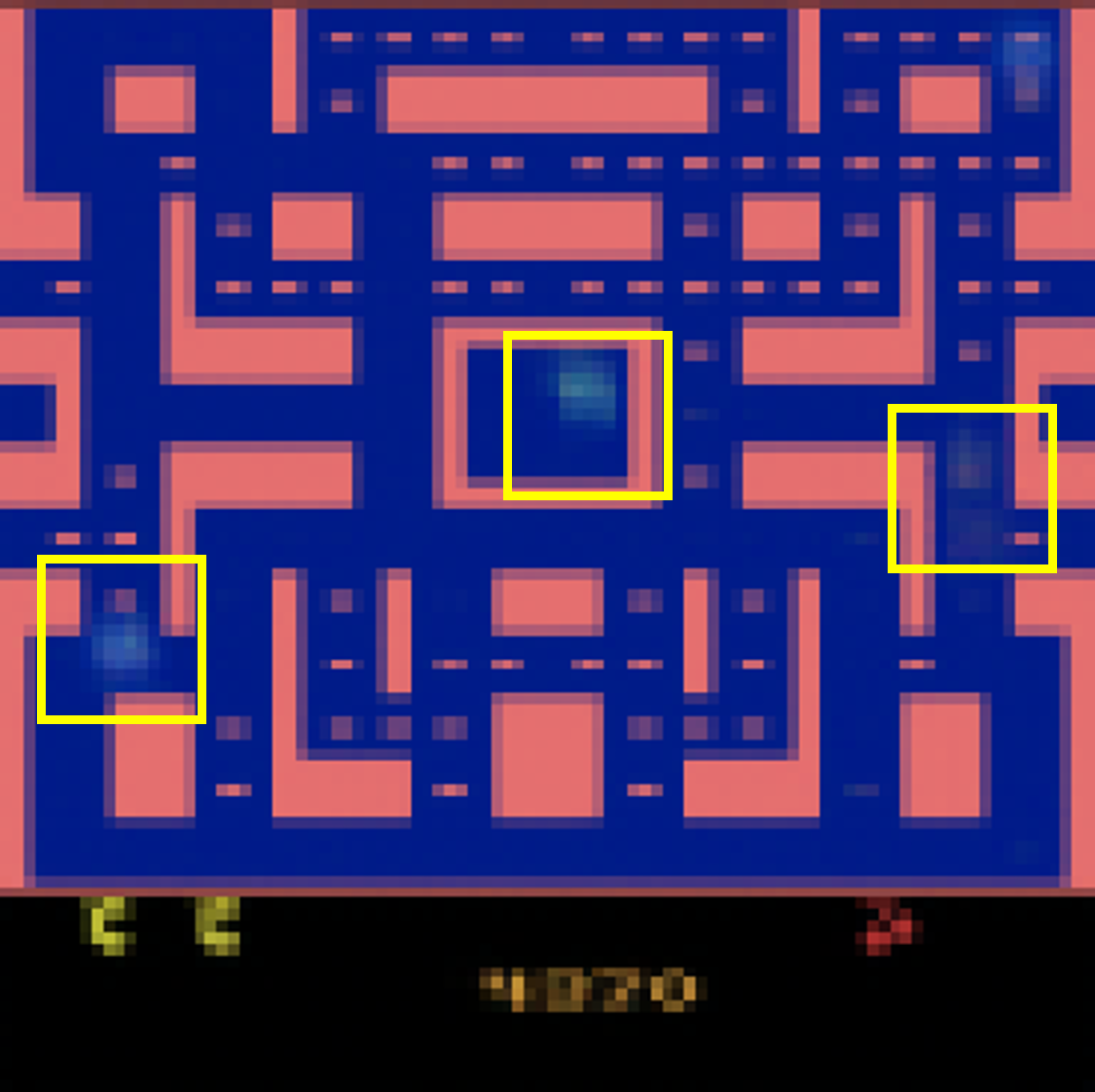} &
        \includegraphics[height=0.087\textwidth, valign=c]{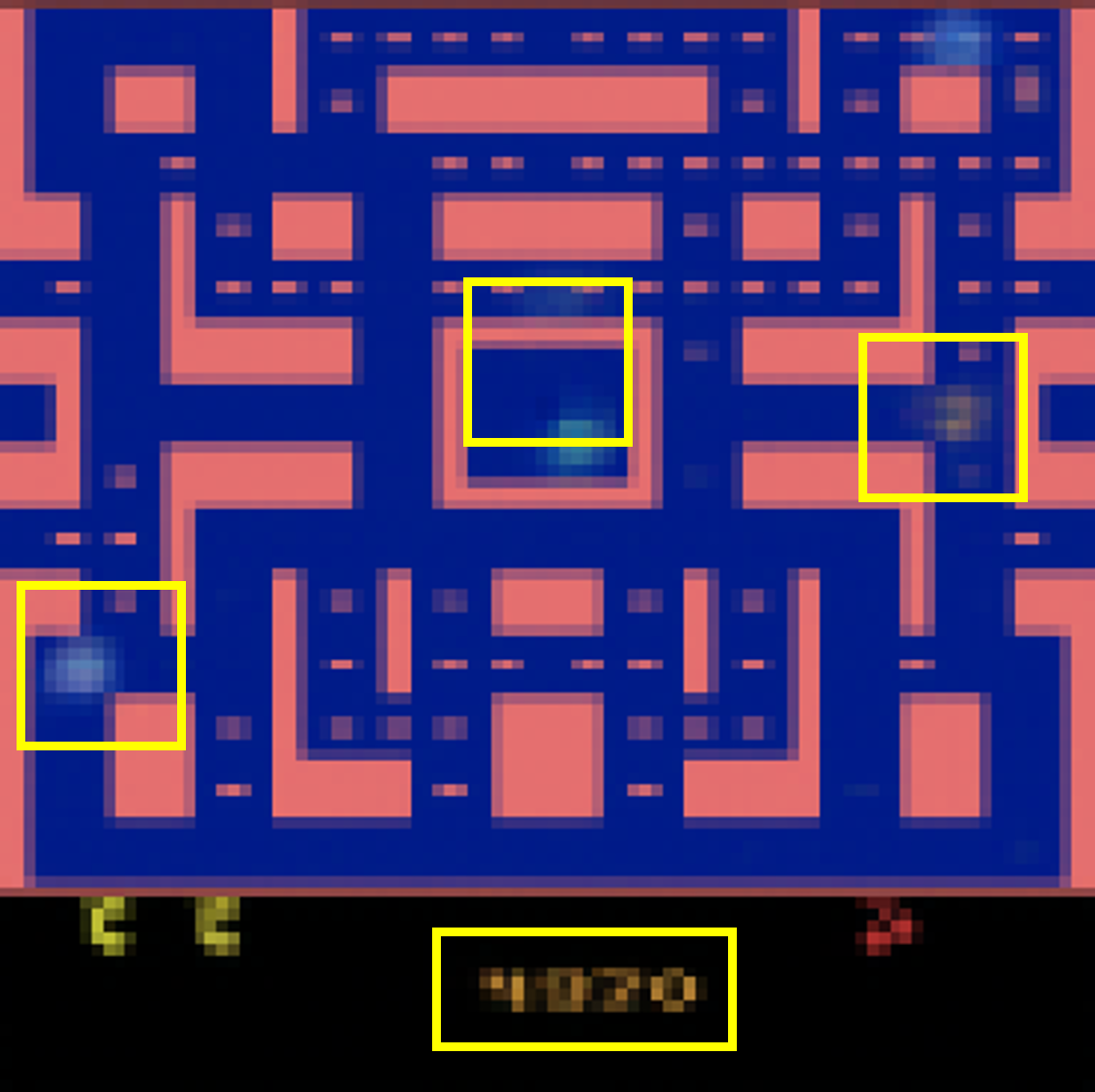} &
        \includegraphics[height=0.087\textwidth, valign=c]{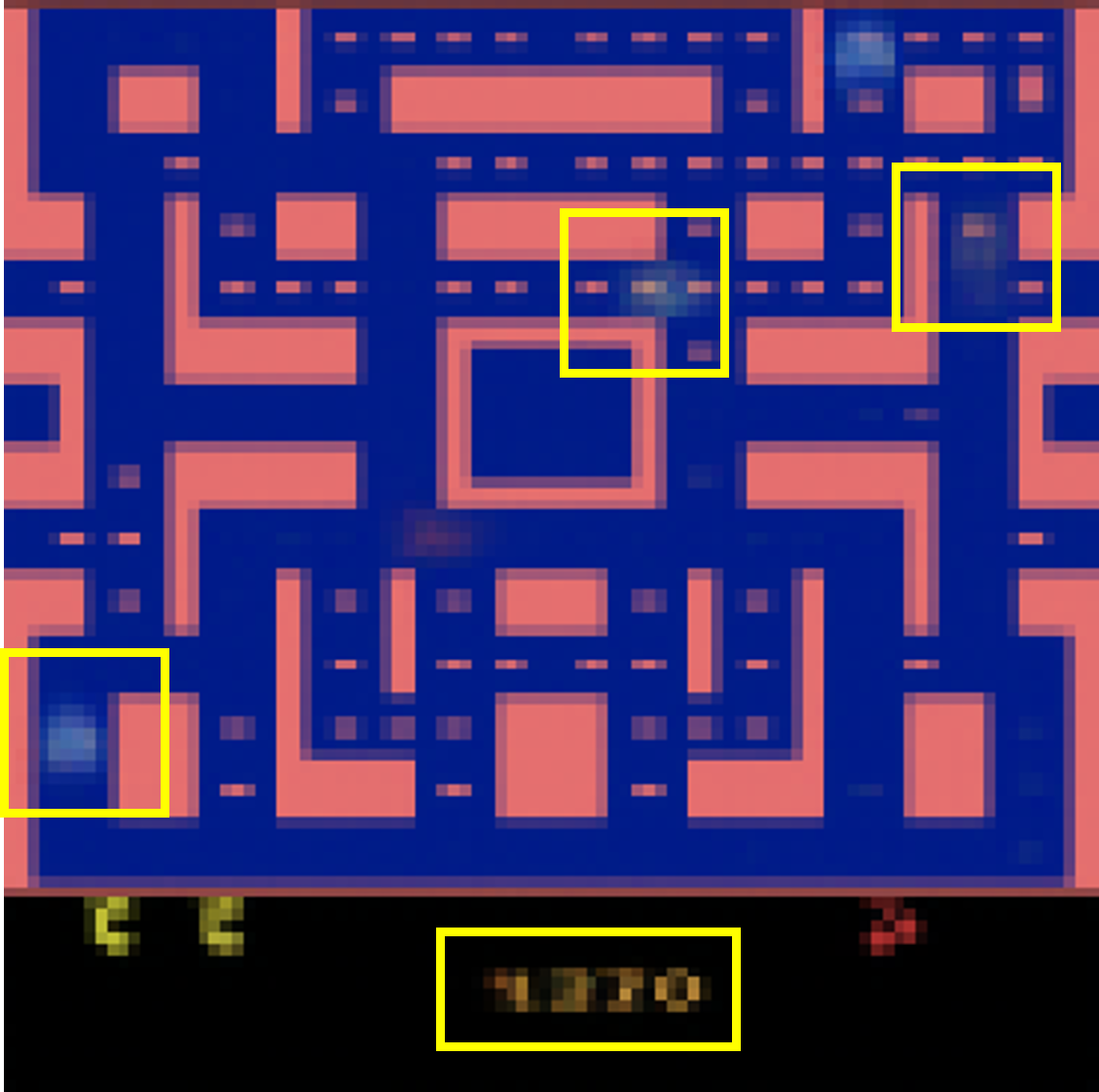} \vspace{0.2em} \\
        \revision{\footnotesize$\hat o_{t+k}^{(k)}$} &
        \includegraphics[height=0.087\textwidth, valign=c]{figures/unroll-pacman/representation_487.png} &
        \includegraphics[height=0.087\textwidth, valign=c]{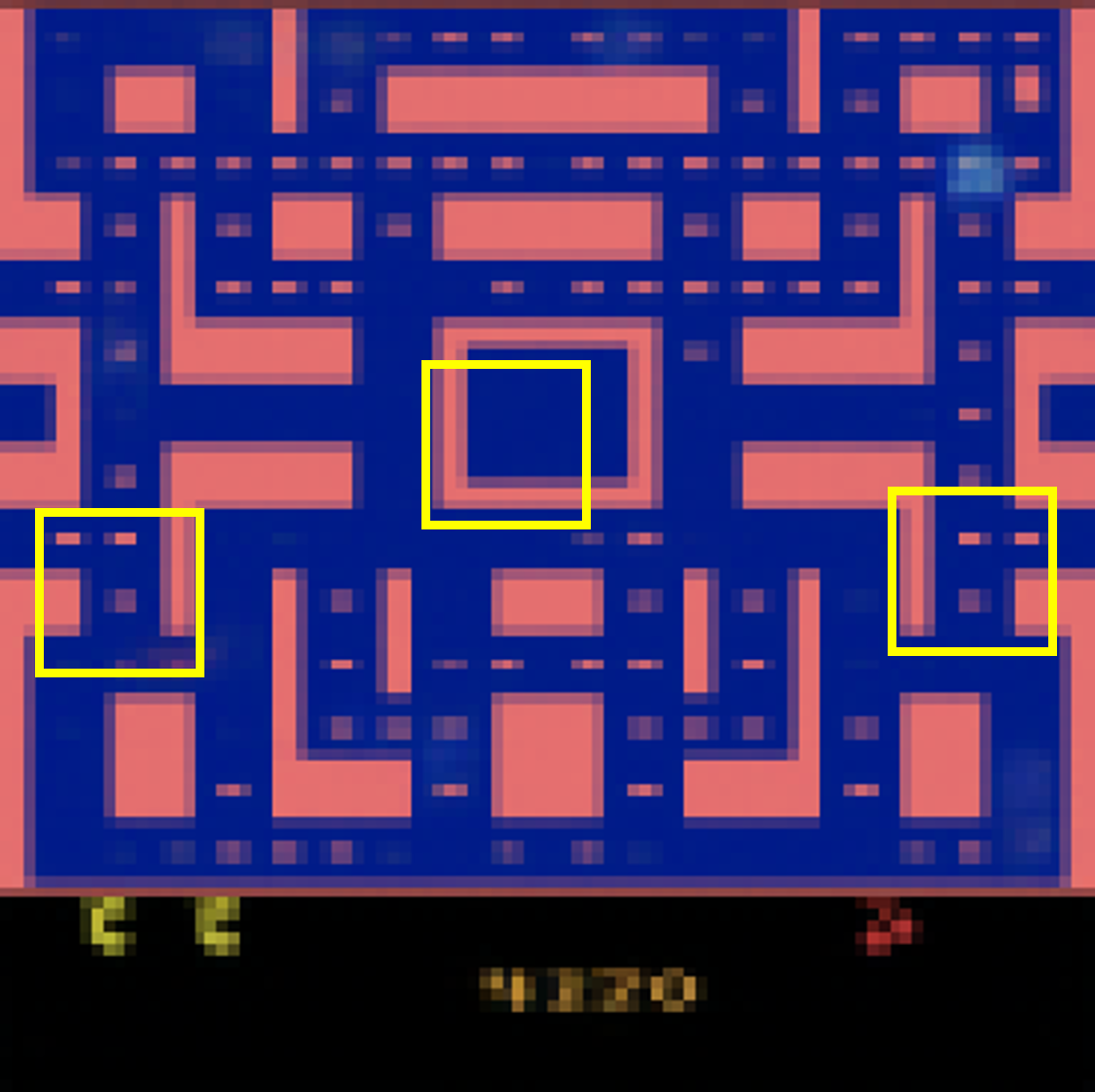} &
        \includegraphics[height=0.087\textwidth, valign=c]{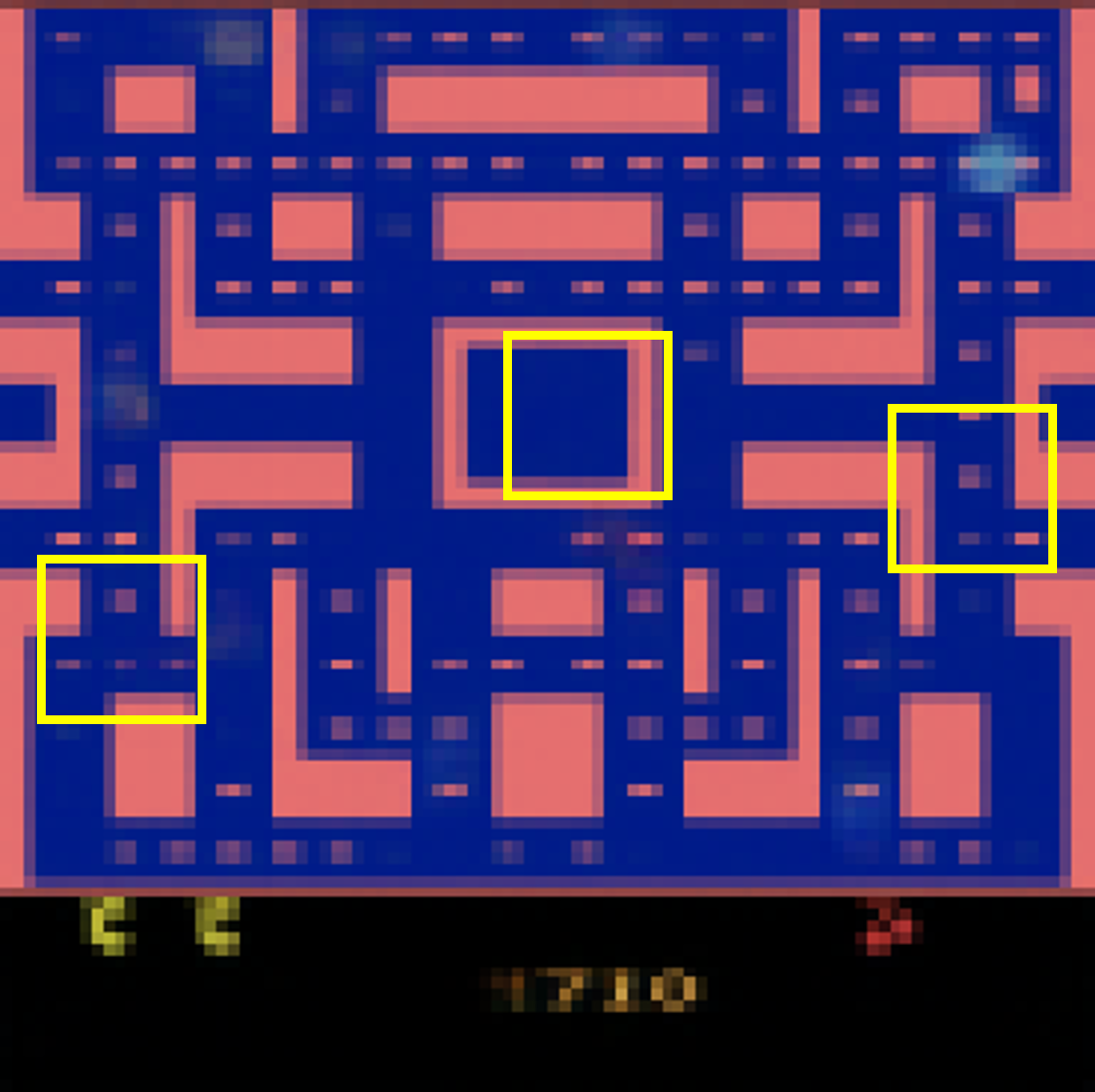} &
        \includegraphics[height=0.087\textwidth, valign=c]{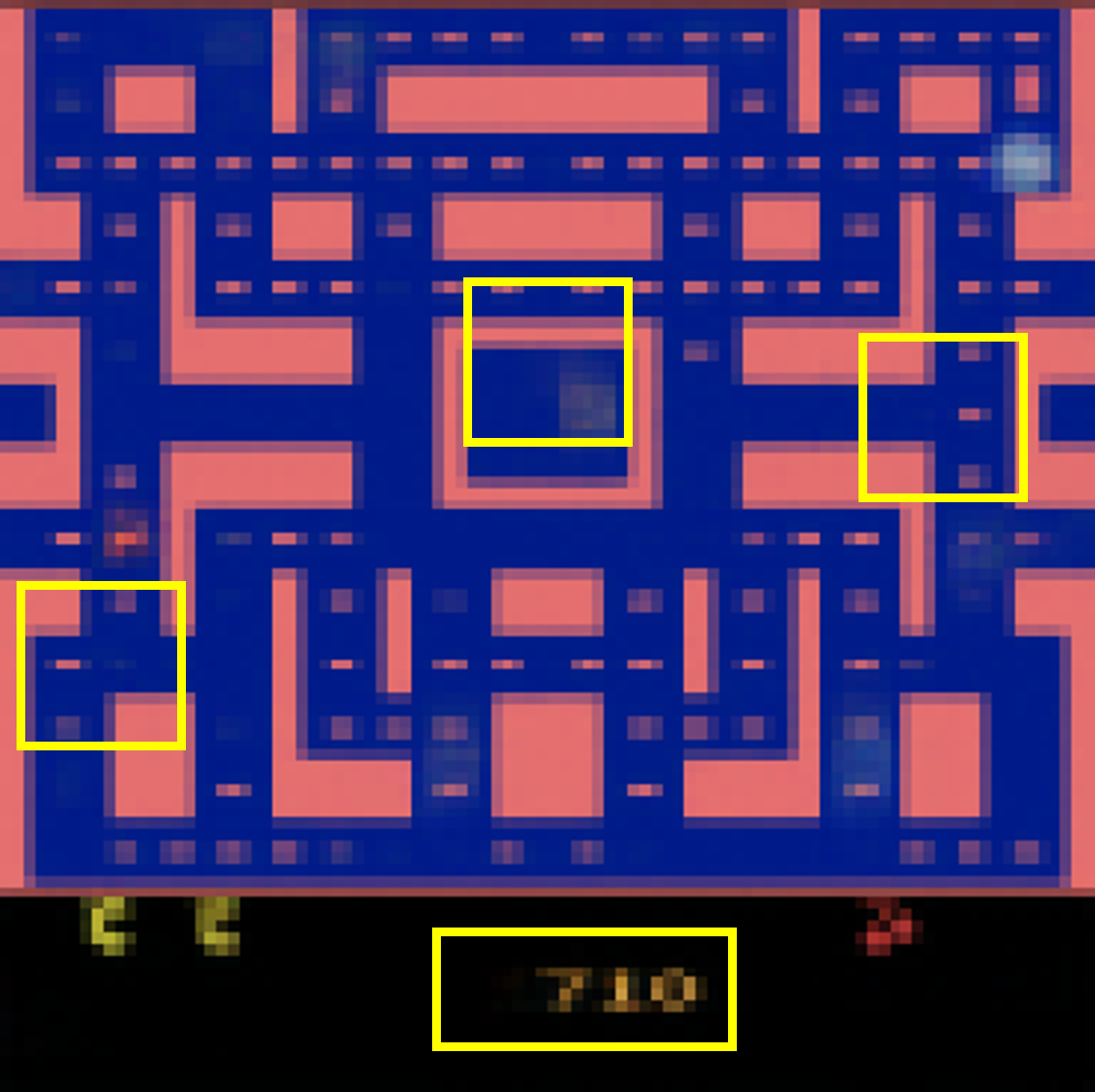} &
        \includegraphics[height=0.087\textwidth, valign=c]{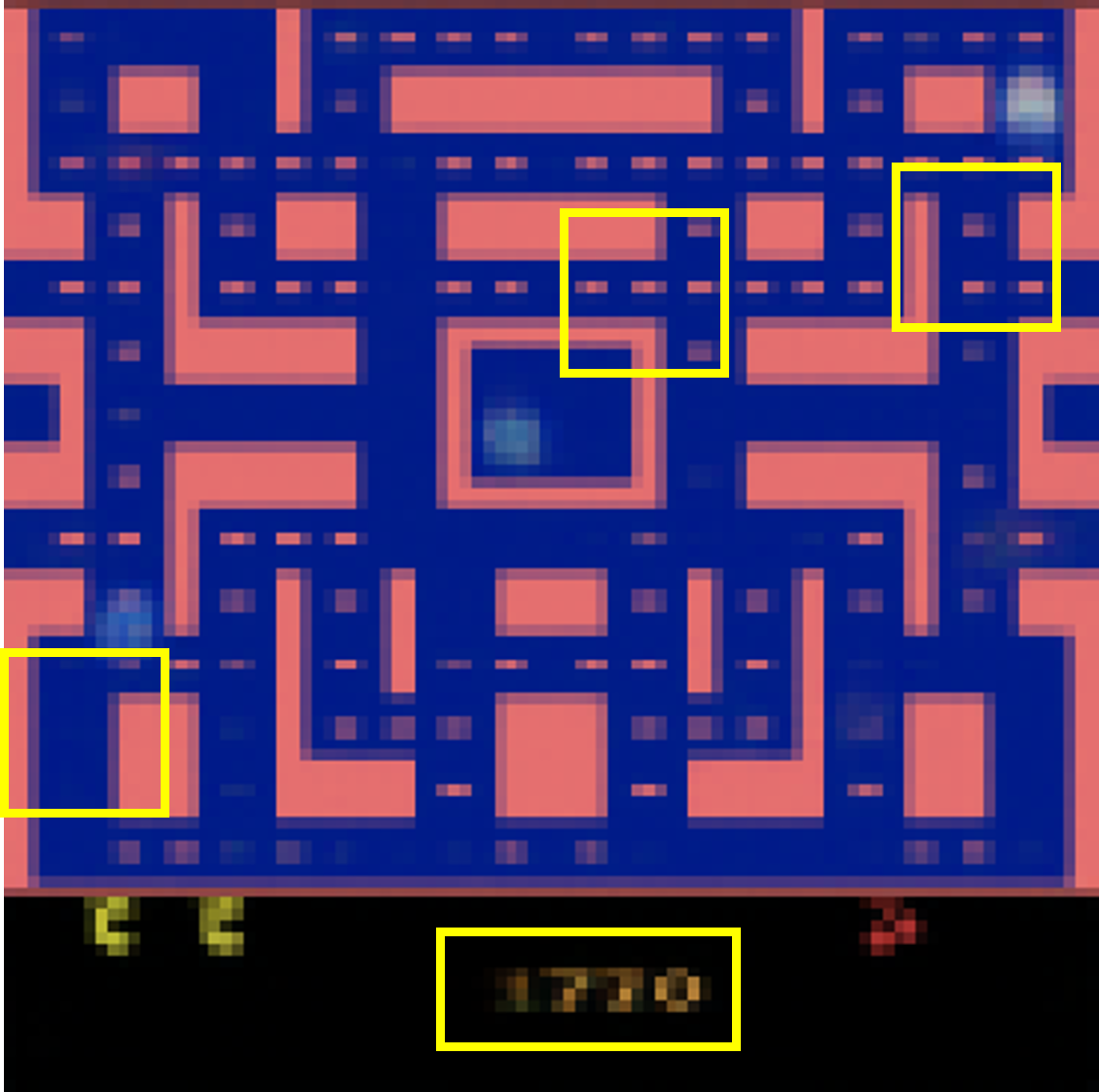} \vspace{0.3em} \\
        & \multicolumn{5}{c}{\footnotesize(b) Ms. Pacman} \vspace{0.3em} \\
        
        \revision{\footnotesize$o_{t+k}$} &
        \includegraphics[height=0.087\textwidth, valign=c]{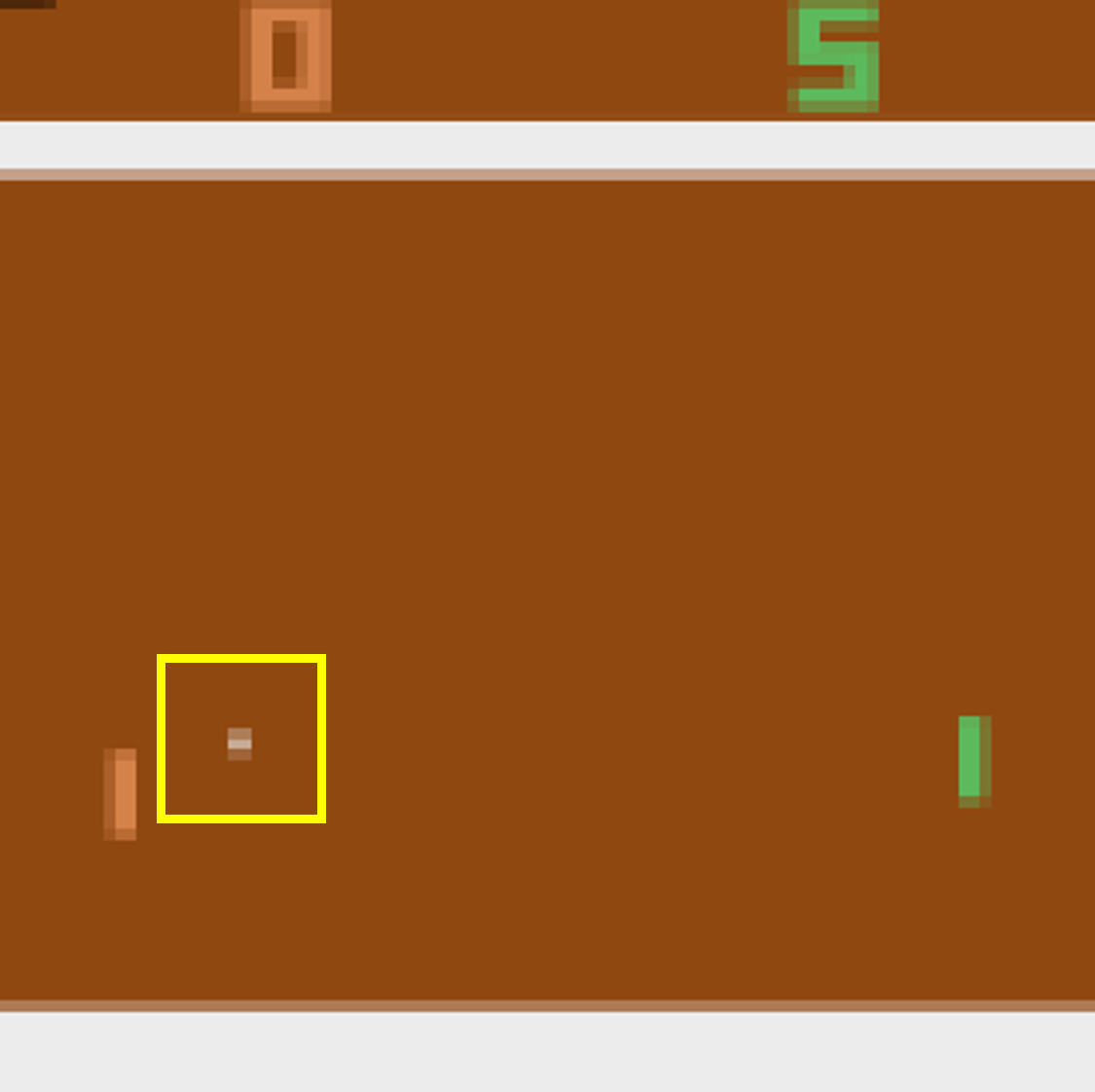} &
        \includegraphics[height=0.087\textwidth, valign=c]{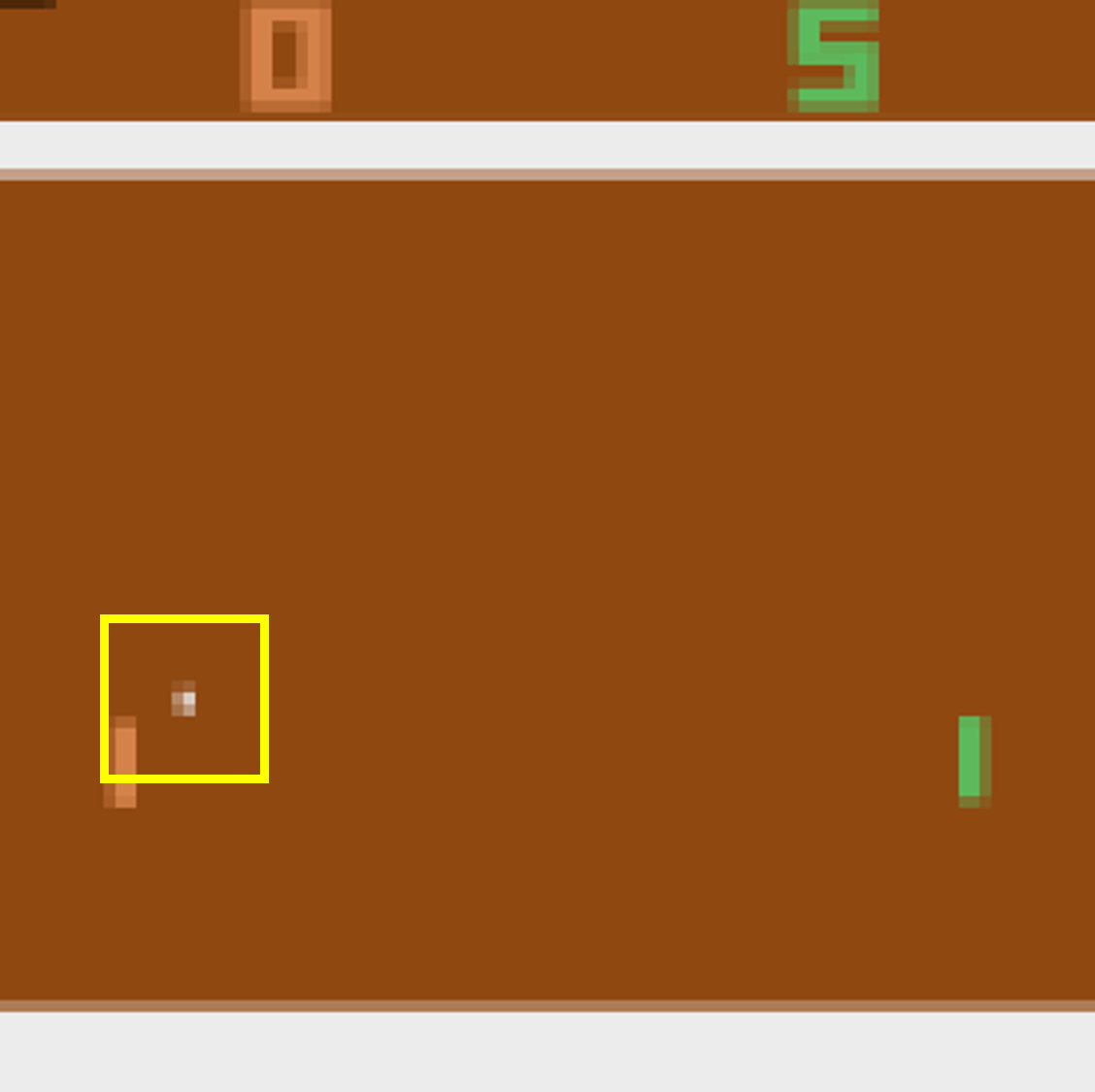} &
        \includegraphics[height=0.087\textwidth, valign=c]{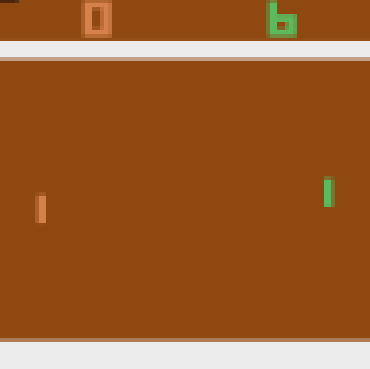} &
        \includegraphics[height=0.087\textwidth, valign=c]{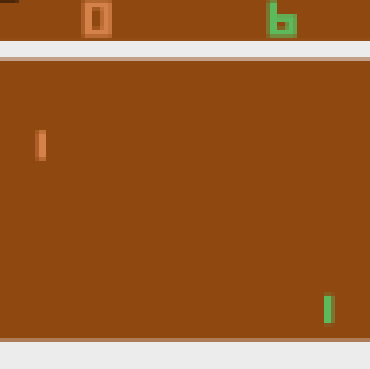} &
        \includegraphics[height=0.087\textwidth, valign=c]{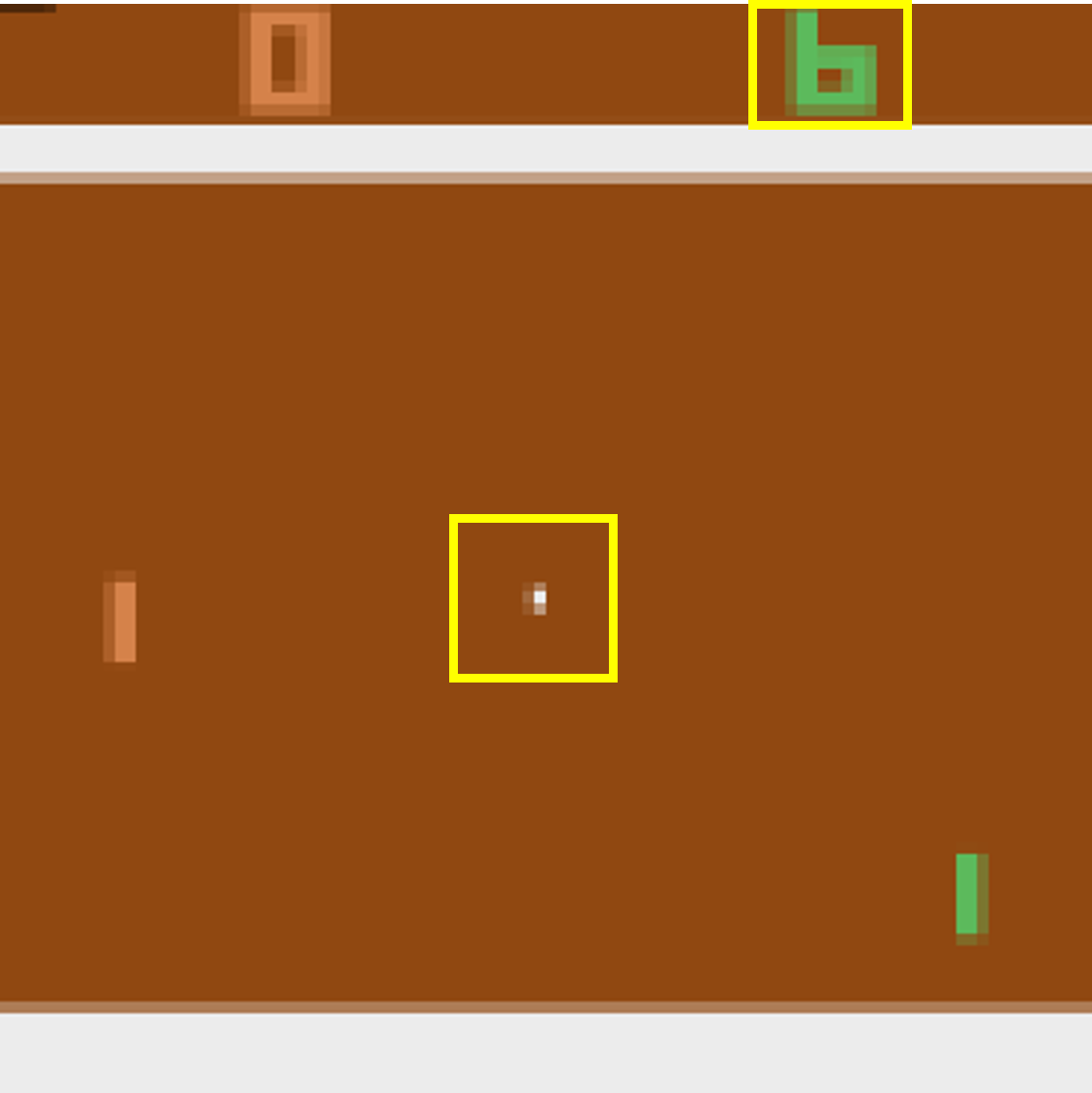} \vspace{0.2em} \\
        \revision{\footnotesize$\hat o_{t+k}$} & 
        \includegraphics[height=0.087\textwidth, valign=c]{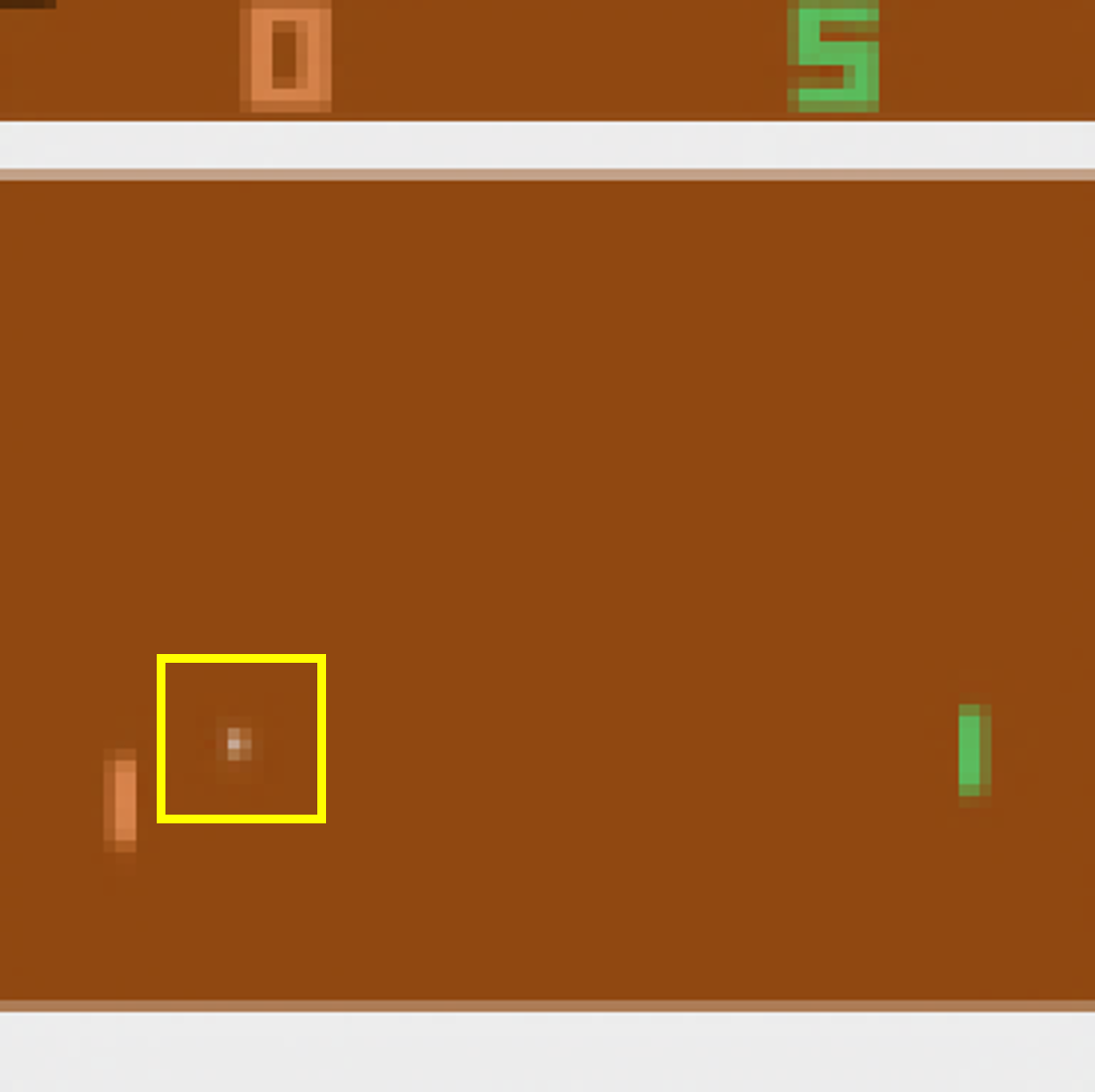} &
        \includegraphics[height=0.087\textwidth, valign=c]{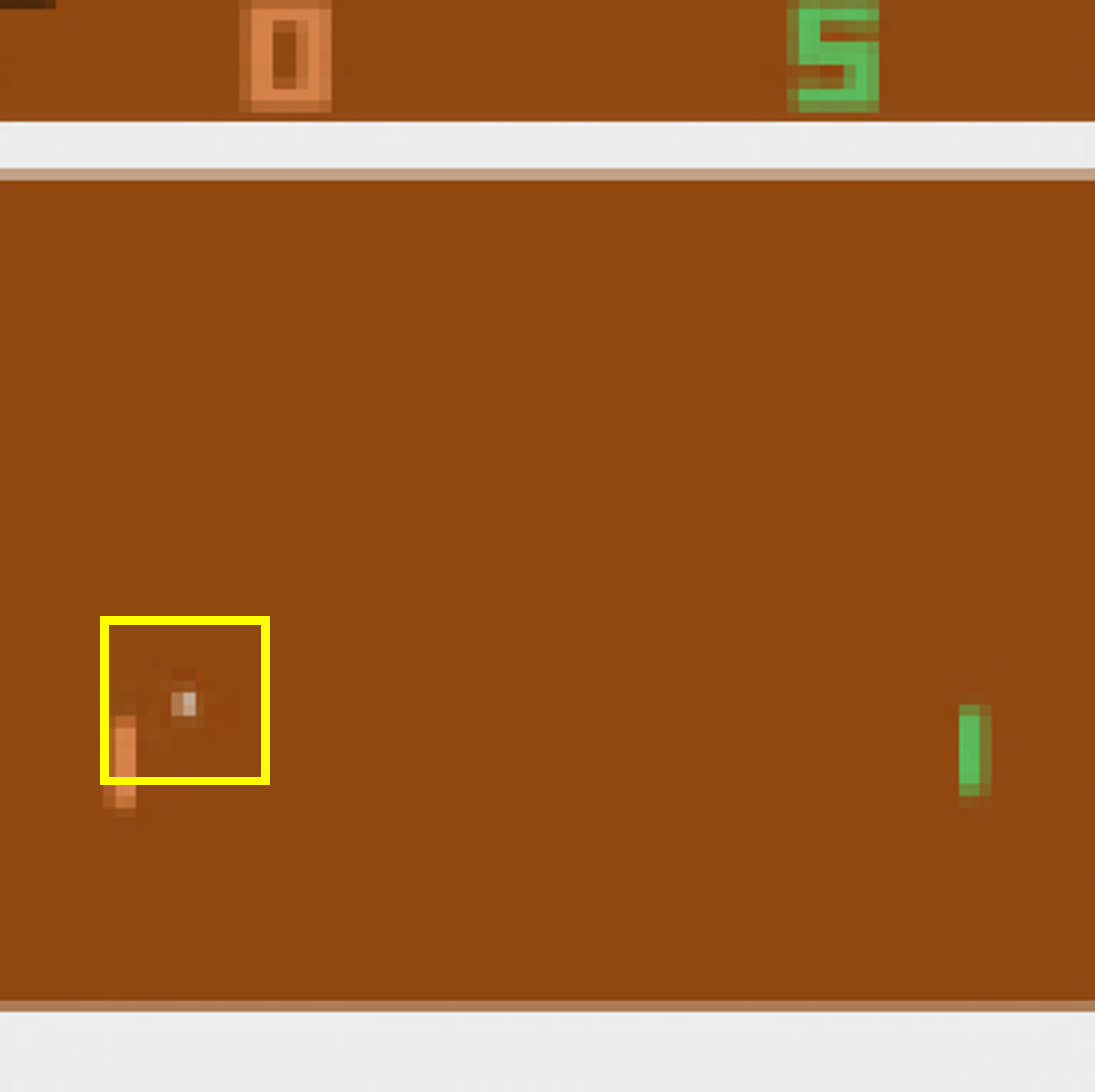} &
        \includegraphics[height=0.087\textwidth, valign=c]{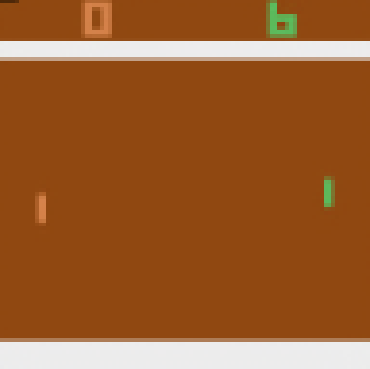} &
        \includegraphics[height=0.087\textwidth, valign=c]{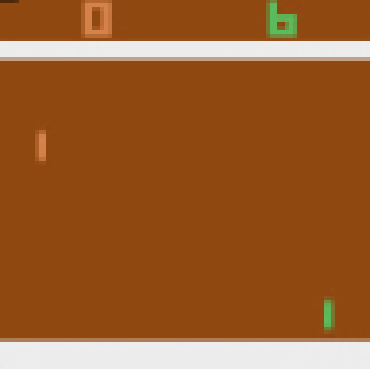} &
        \includegraphics[height=0.087\textwidth, valign=c]{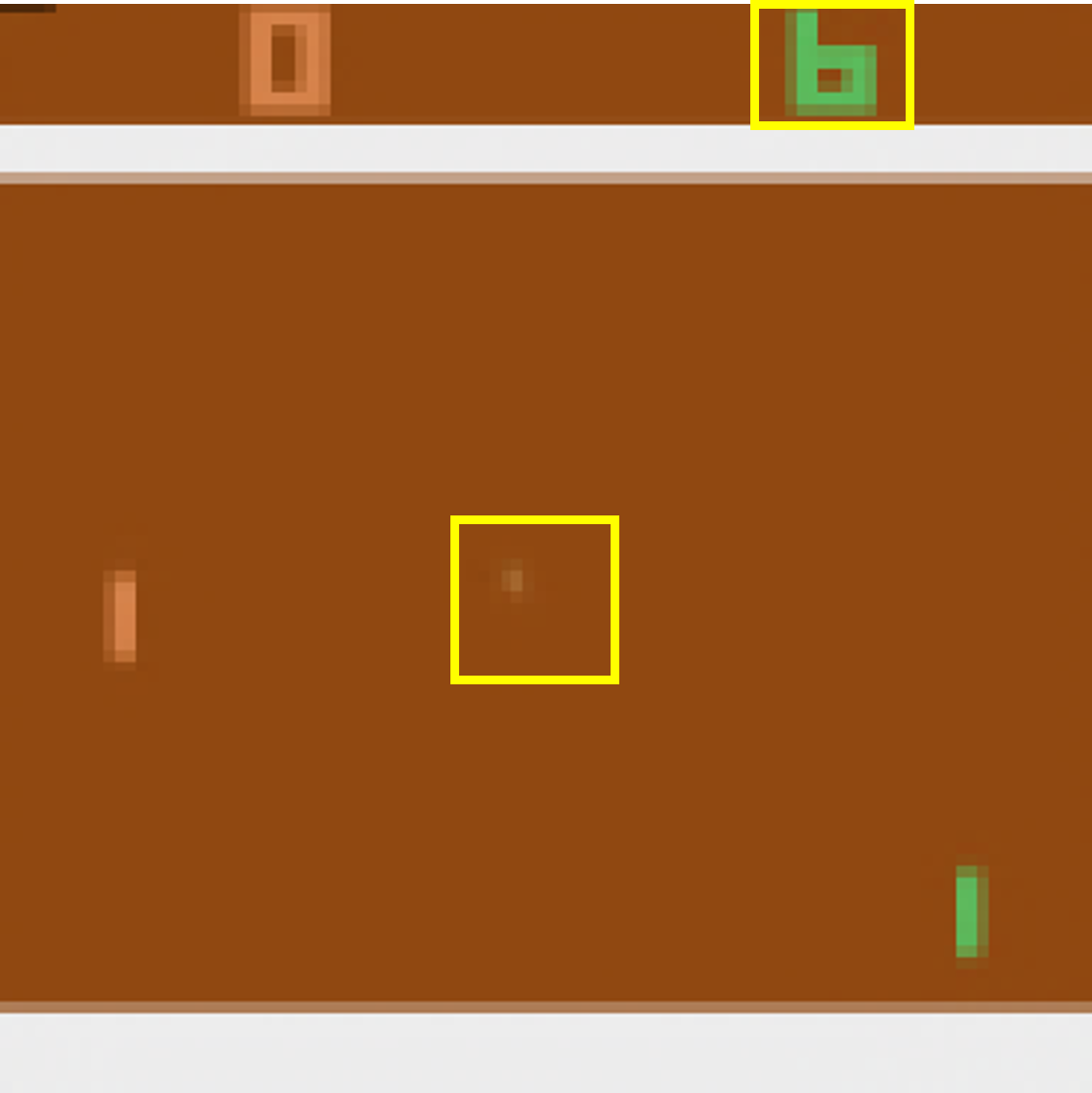} \vspace{0.2em} \\
        \revision{\footnotesize$\hat o_{t+k}^{(k)}$} &
        \includegraphics[height=0.087\textwidth, valign=c]{figures/unroll-pong/representation_464.png} &
        \includegraphics[height=0.087\textwidth, valign=c]{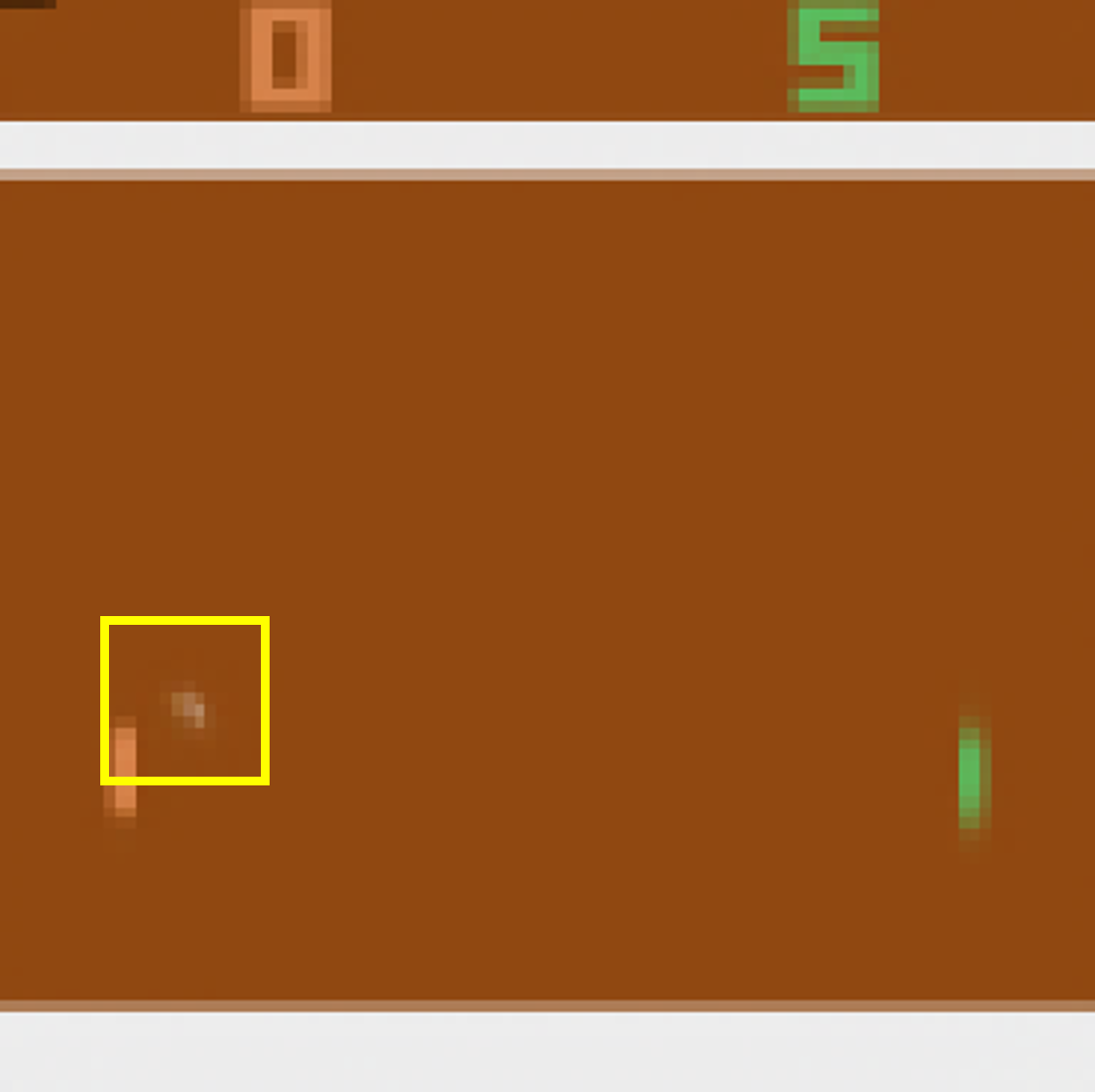} &
        \includegraphics[height=0.087\textwidth, valign=c]{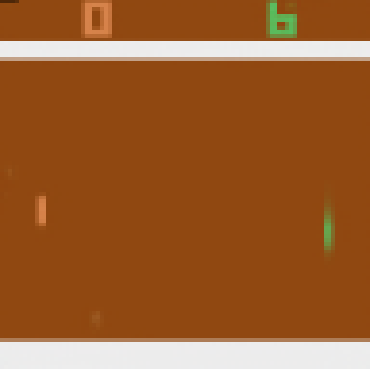} &
        \includegraphics[height=0.087\textwidth, valign=c]{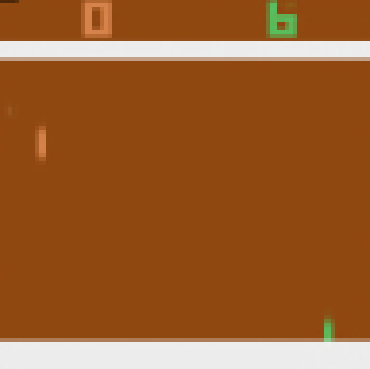} &
        \includegraphics[height=0.087\textwidth, valign=c]{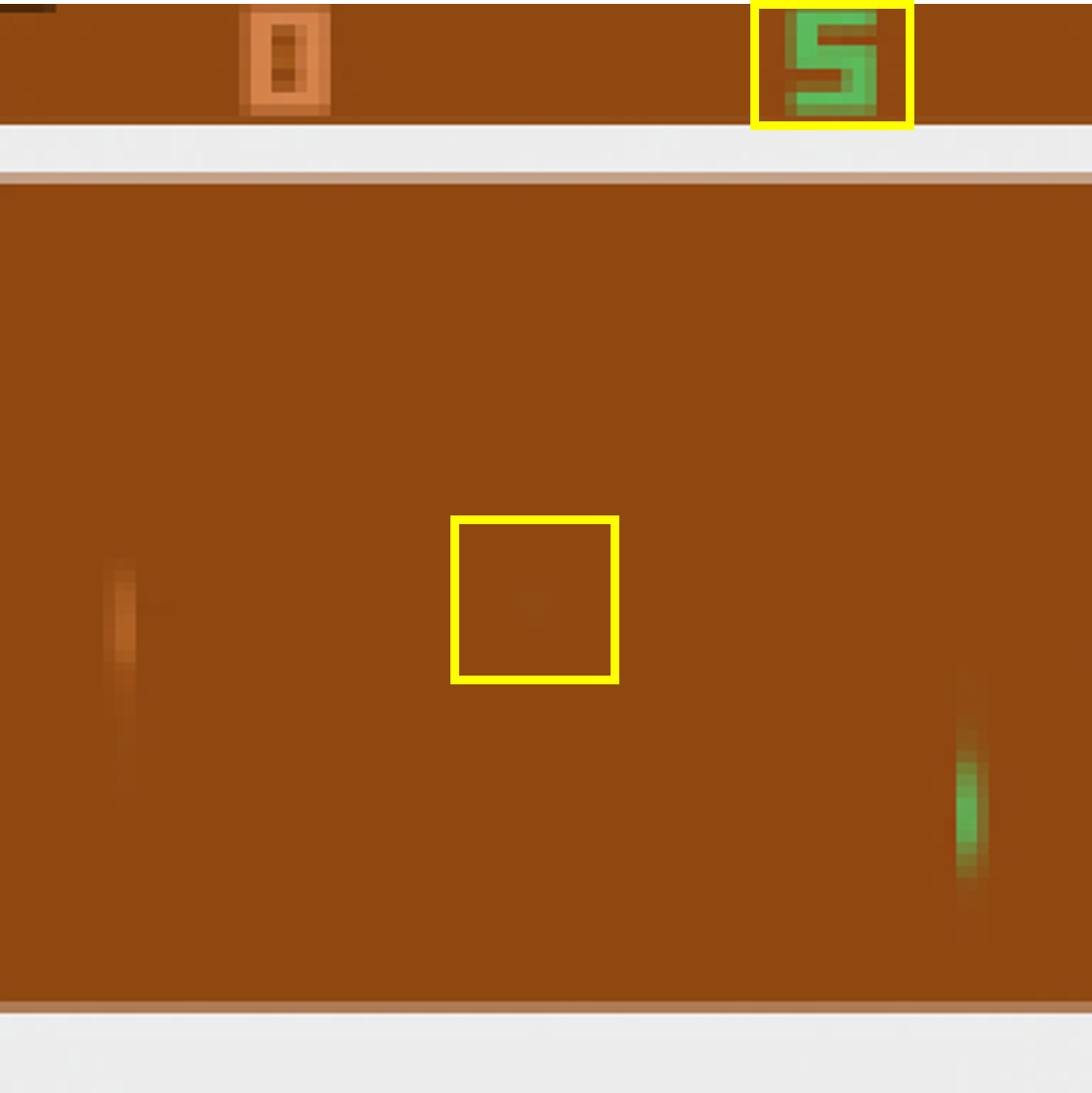} \vspace{0.3em} \\
        & \multicolumn{5}{c}{\footnotesize(c) Pong} \vspace{0em} \\
    \end{tabular}}
    \caption{The comparison between true and decoded observations at different unrolling steps in Breakout, Ms. Pacman, and Pong.
    \revision{The differences are highlighted in the yellow boxes.}}
    \label{fig:unroll-atari-games}
\end{figure}

Next, we evaluate the quality of the unrolled hidden states generated by the dynamics network, examining whether $\hat o_{t+k}^{(k)}$ still aligns with $o_{t+k}$ after unrolling $k$ steps.
Fig. \ref{fig:unroll-board-games} and Fig. \ref{fig:unroll-atari-games} present the results for five games under different $k$, including 0, 1, 5, 10, and 20.
For comparison, $\hat o_{t+k}$, the decoded observations for the hidden states generated by the representation network, are also included.
Note that when $k=0$, $\hat o_{t+k}$ is equivalent to $\hat o_{t+k}^{(k)}$ as no unrolling occurs.
For Go and Gomoku, $\hat o_{t+k}^{(k)}$ remain robust with a small number of unrolling steps\revision{,} but become blurry as $k$ increases like $\hat o_{t+20}^{(20)}$.
By examining $\hat o_{t+k}$, it can be observed that there are several incorrectly shown blurry pieces on the Gomoku board.
This implies that the representation network in Gomoku is less accurate than that in Go, which is likely due to a large board size.

Conversely, $\hat o_{t+k}^{(k)}$ in Atari games are less accurate, in which the larger $k$ is, the more mistakes are, especially for the small and moving objects.
In the three Atari games, the scores are prone to blur after unrolling.
Similarly, the paddle and the blocks in Breakout, the Pacman and ghosts in Ms. Pacman, and the ball in Pong can also appear blurry.
Nevertheless, the quality of $\hat o_{t+k}^{(k)}$ is significantly improved compared to those in \cite{ye_mastering_2021}, possibly because of a much higher decoder loss coefficient ($\lambda_d$) in this work.
Furthermore, $\hat o_{t+k}$ show differences in these games: Breakout cannot decode the ball, Ms. Pacman cannot clearly decode any moving objects, while Pong can blurrily decode the ball.
To summarize, the quality of the learned representation and dynamics networks are generally correlated with the visual complexity of the observations, and the hidden states generally become less accurate during unrolling through the dynamics network, similar to the findings in previous works \cite{vries_visualizing_2021, he_what_2023}.

\subsection{Analyzing the quality of unrolled hidden states}

\begin{figure*}[t]
    \centering
    \small
    {\setlength{\tabcolsep}{0.1em}
    \begin{tabular}{ccccc}
    \multicolumn{5}{c}{\includegraphics[width=0.95\textwidth]{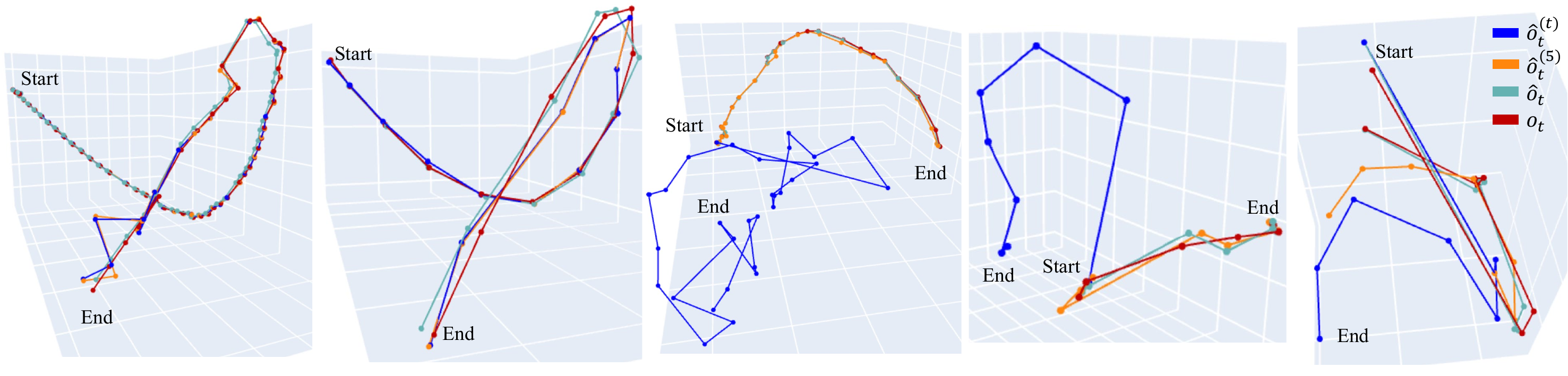}} \\
    \makebox[0.19\textwidth]{\footnotesize(a) Go} & 
    \makebox[0.19\textwidth]{\footnotesize(b) Gomoku} & 
    \makebox[0.19\textwidth]{\footnotesize(c) Breakout} & 
    \makebox[0.19\textwidth]{\footnotesize(d) Ms. Pacman} & 
    \makebox[0.19\textwidth]{\footnotesize(e) Pong}
    \end{tabular}}
    \caption{The PCA projections of true and different decoded observations in five games. Labels ``Start'' and ``End'' mark the initial and the terminal of the trajectories.}
    \label{fig:pc_projection_go_breakout_in_ID}
\end{figure*}

Also, following a similar approach to \cite{vries_visualizing_2021}, we perform principal component analysis (PCA) on the same games and visualize the PCA projections for $o_t$, $\hat{o}_t$, $\hat{o}_t^{(5)}$, and $\hat{o}_t^{(t)}$, as shown in Fig. \ref{fig:pc_projection_go_breakout_in_ID}.
For board games, although there are several mistakes after long unrolling, the trajectories of all decoded observations align well with $o_t$, indicating that the dynamics network effectively learns the game rules and can simulate future observations regardless of the unrolling steps.
In contrast, for Atari games, only $\hat{o}_t^{(5)}$ remains aligned, implying that the dynamics networks are significantly less effective.
This may be because MuZero only unrolls five steps during training; when unrolling exceeds five steps, the decoder performance significantly degrades, as shown by $\hat{o}_t^{(t)}$.
In summary, the error between decoded and true observations gradually accumulates as the unrolling step increases across all games, with errors in board games generally increasing less than in Atari games.

\begin{figure*}[tp]
    \centering
    \small

    \begin{tabular}{ccc}
        \hspace{3.5em}
        &
        \begin{minipage}[t]{0.59\textwidth}
            \centering
            \includegraphics[width=\textwidth]{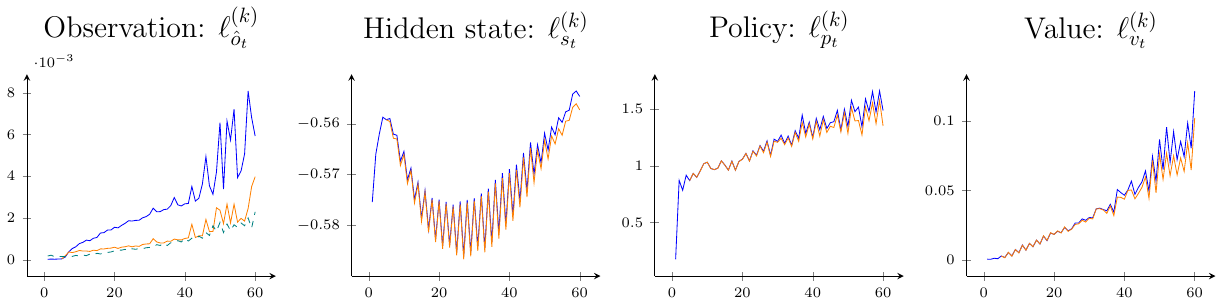}

            \footnotesize(a) Errors in \revision{recent-training} game trajectories
            \vspace{1em}

            \includegraphics[width=\textwidth]{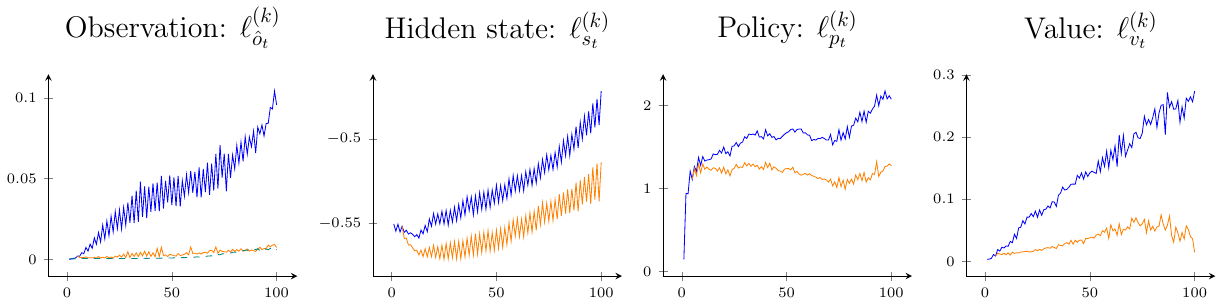}

            \footnotesize(b) Errors in \revision{early-training} game trajectories
            \vspace{1em}
        \end{minipage}
        &
        \begin{minipage}[t]{0.1\textwidth}
            \vspace{0pt} 
            \begin{tikzpicture}
                \begin{axis}[
                        hide axis,
                        legend style={
                            at={(0,1)},
                            anchor=north west,
                            draw=none,
                            fill=none,
                            legend cell align={left},
                            nodes={right},
                        },
                    ]
                    \addplot[blue] coordinates {(0,0)};
                    \addlegendentry{$k=t$}
                    \addplot[orange] coordinates {(0,0)};
                    \addlegendentry{$k=5$}
                    \addplot[teal, dashed] coordinates {(0,0)};
                    \addlegendentry{$k=0$}
                \end{axis}
            \end{tikzpicture}
        \end{minipage}
    \end{tabular}

    \caption{The average unrolling errors for observations, hidden states, policies, and values in Go. The x-axis represents the time step ($t$), and the y-axis represents the error.}
    \label{fig:unroll-errors-Go}
\end{figure*}

\begin{figure*}[tp]
    \centering
    \small

    \begin{tabular}{ccc}
        \hspace{3.5em}
        &
        \begin{minipage}[t]{0.59\textwidth}
            \centering
            \includegraphics[width=\textwidth]{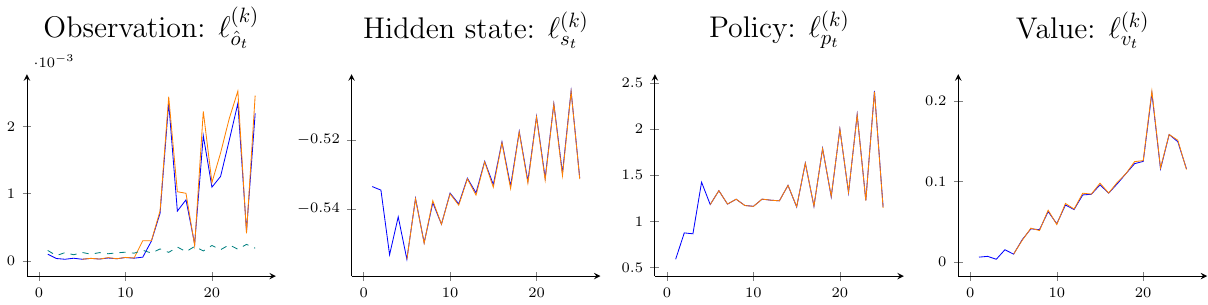}

            \footnotesize(a) Errors in \revision{recent-training} game trajectories
            \vspace{1em}

            \includegraphics[width=\textwidth]{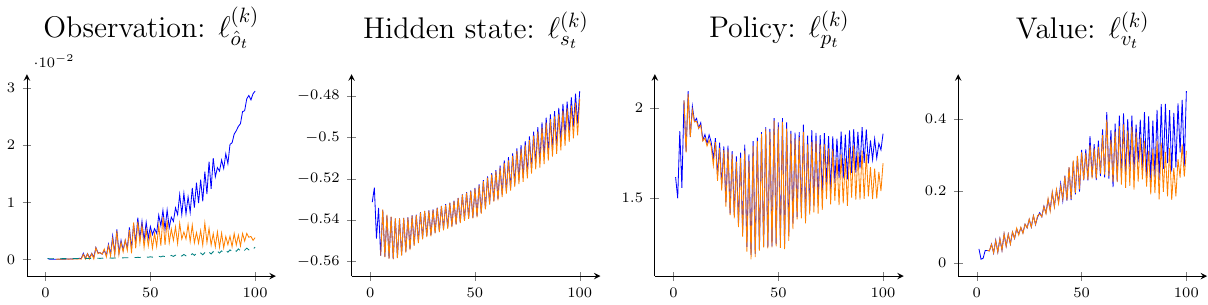}

            \footnotesize(b) Errors in \revision{early-training} game trajectories
            \vspace{1em}
        \end{minipage}
        &
        \begin{minipage}[t]{0.1\textwidth}
            \vspace{0pt} 
            \begin{tikzpicture}
                \begin{axis}[
                        hide axis,
                        legend style={
                            at={(0,1)},
                            anchor=north west,
                            draw=none,
                            fill=none,
                            legend cell align={left},
                            nodes={right},
                        },
                    ]
                    \addplot[blue] coordinates {(0,0)};
                    \addlegendentry{$k=t$}
                    \addplot[orange] coordinates {(0,0)};
                    \addlegendentry{$k=5$}
                    \addplot[teal, dashed] coordinates {(0,0)};
                    \addlegendentry{$k=0$}
                \end{axis}
            \end{tikzpicture}
        \end{minipage}
    \end{tabular}

    \caption{The average unrolling errors for observations, hidden states, policies, and values in Gomoku. The x-axis represents the time step ($t$), and the y-axis represents the error.}
    \label{fig:unroll-errors-Gomoku}
\end{figure*}

\begin{figure*}[tp]
    \centering
    \small

    \begin{tabular}{ccc}
        \hspace{2em}
        &
        \begin{minipage}[t]{0.72\textwidth}
            \centering
            \includegraphics[width=\textwidth]{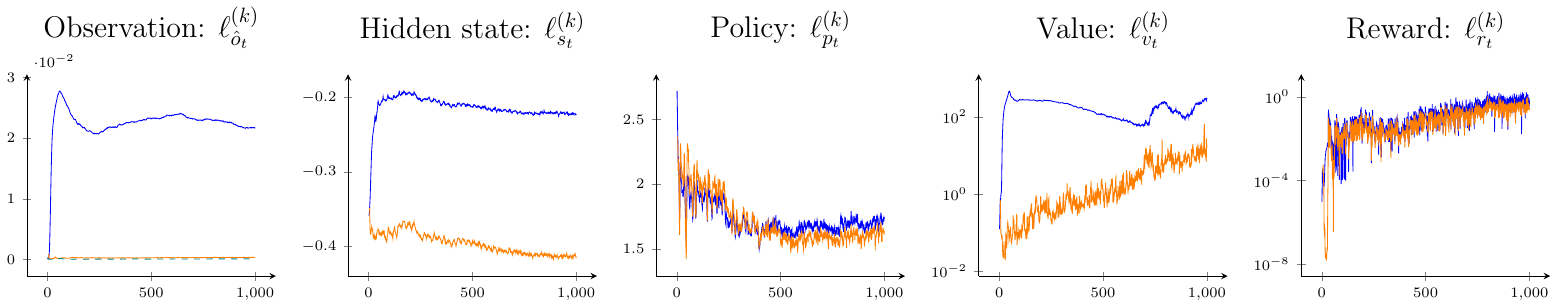}

            \footnotesize(a) Errors in \revision{recent-training} game trajectories
            \vspace{1em}

            \includegraphics[width=\textwidth]{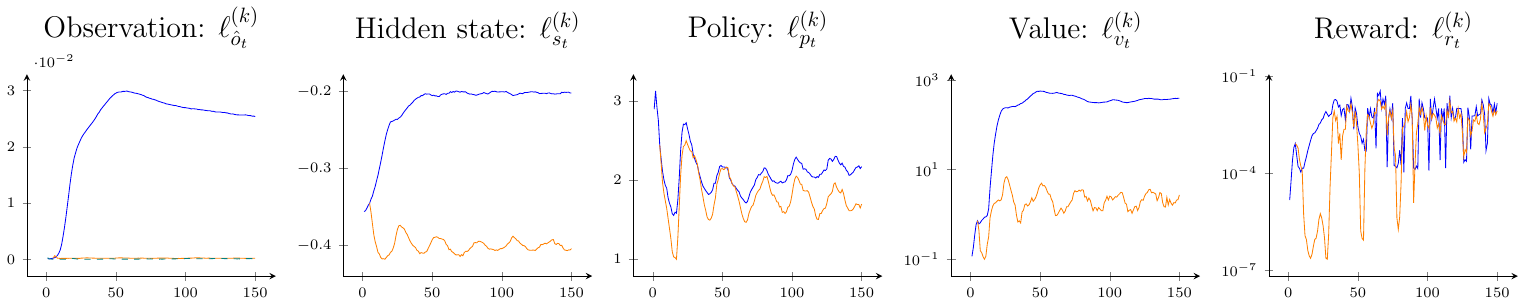}

            \footnotesize(b) Errors in \revision{early-training} game trajectories
            \vspace{1em}
        \end{minipage}
        &
        \begin{minipage}[t]{0.1\textwidth}
            \vspace{0pt} 
            \begin{tikzpicture}
                \begin{axis}[
                        hide axis,
                        legend style={
                            at={(0,1)},
                            anchor=north west,
                            draw=none,
                            fill=none,
                            legend cell align={left},
                            nodes={right},
                        },
                    ]
                    \addplot[blue] coordinates {(0,0)};
                    \addlegendentry{$k=t$}
                    \addplot[orange] coordinates {(0,0)};
                    \addlegendentry{$k=5$}
                    \addplot[teal, dashed] coordinates {(0,0)};
                    \addlegendentry{$k=0$}
                \end{axis}
            \end{tikzpicture}
        \end{minipage}
    \end{tabular}

    \caption{The average unrolling errors for observations, hidden states, policies, values, and rewards in Breakout. The x-axis represents the time step ($t$), and the y-axis represents the error.}
    \label{fig:unroll-errors-Breakout}
\end{figure*}

\begin{figure*}[tp]
    \centering
    \small

    \begin{tabular}{ccc}
        \hspace{2em}
        &
        \begin{minipage}[t]{0.72\textwidth}
            \centering
            \includegraphics[width=\textwidth]{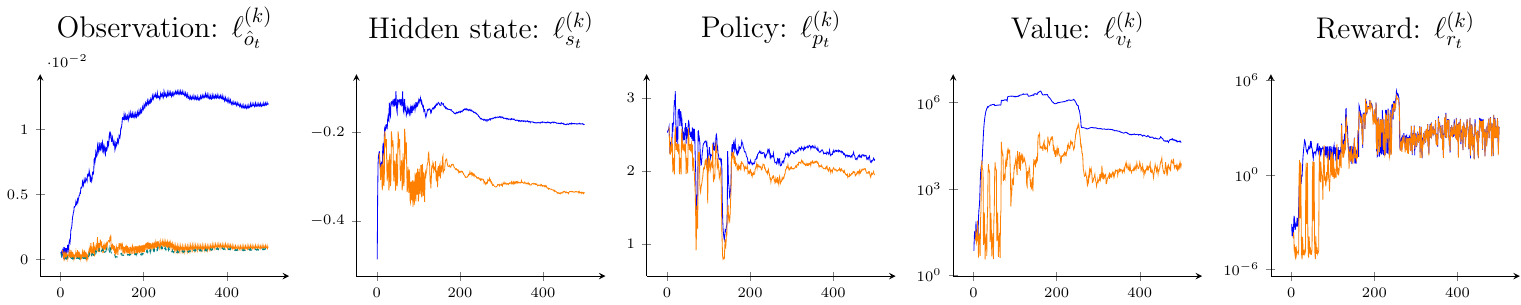}

            \footnotesize(a) Errors in \revision{recent-training} game trajectories
            \vspace{1em}

            \includegraphics[width=\textwidth]{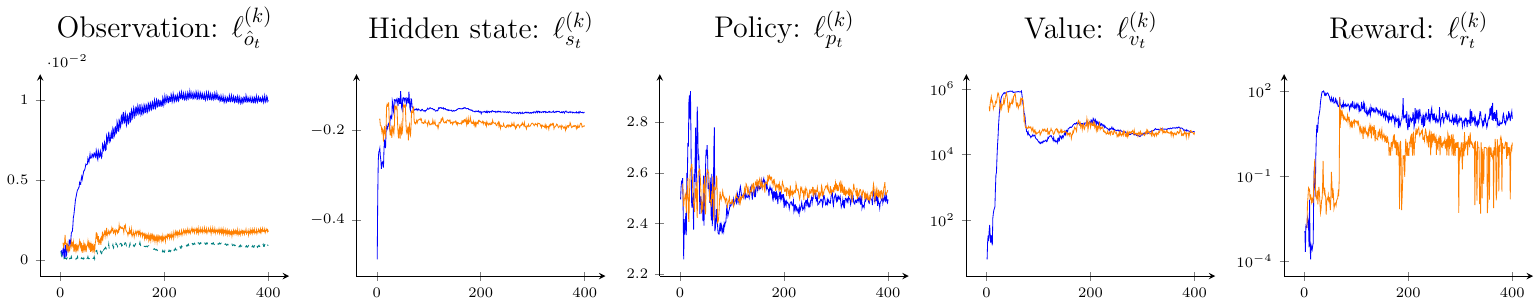}

            \footnotesize(b) Errors in \revision{early-training} game trajectories
            \vspace{1em}
        \end{minipage}
        &
        \begin{minipage}[t]{0.1\textwidth}
            \vspace{0pt} 
            \begin{tikzpicture}
                \begin{axis}[
                        hide axis,
                        legend style={
                            at={(0,1)},
                            anchor=north west,
                            draw=none,
                            fill=none,
                            legend cell align={left},
                            nodes={right},
                        },
                    ]
                    \addplot[blue] coordinates {(0,0)};
                    \addlegendentry{$k=t$}
                    \addplot[orange] coordinates {(0,0)};
                    \addlegendentry{$k=5$}
                    \addplot[teal, dashed] coordinates {(0,0)};
                    \addlegendentry{$k=0$}
                \end{axis}
            \end{tikzpicture}
        \end{minipage}
    \end{tabular}

    \caption{The average unrolling errors for observations, hidden states, policies, values, and rewards in Ms. Pacman. The x-axis represents the time step ($t$), and the y-axis represents the error.}
    \label{fig:unroll-errors-Ms_pacman}
\end{figure*}

\begin{figure*}[tp]
    \centering
    \small

    \begin{tabular}{ccc}
        \hspace{3.5em}
        &
        \begin{minipage}[t]{0.72\textwidth}
            \centering
            \includegraphics[width=\textwidth]{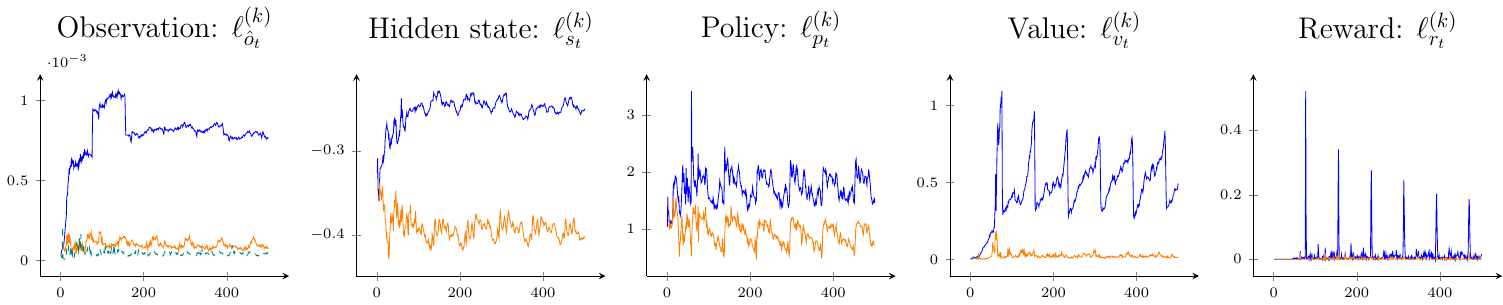}

            \footnotesize(a) Errors in \revision{recent-training} game trajectories
            \vspace{1em}

            \includegraphics[width=\textwidth]{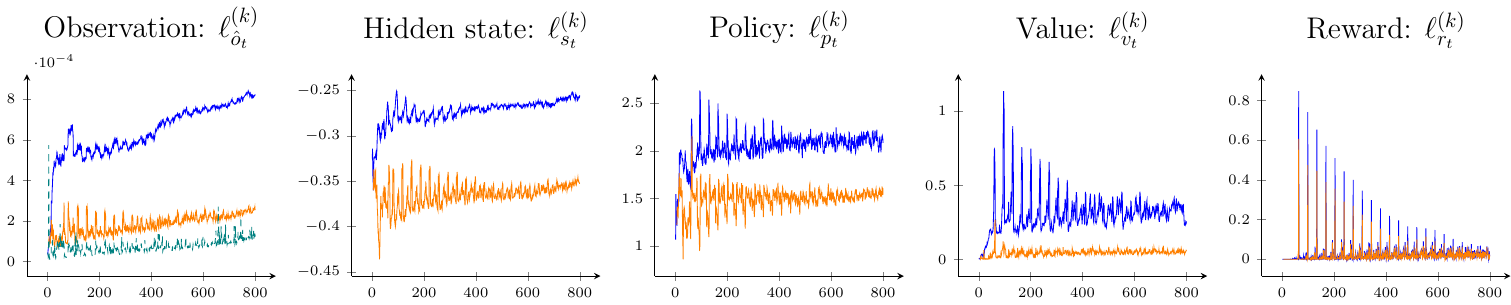}

            \footnotesize(b) Errors in \revision{early-training} game trajectories
            \vspace{1em}
        \end{minipage}
        &
        \begin{minipage}[t]{0.1\textwidth}
            \vspace{0pt} 
            \begin{tikzpicture}
                \begin{axis}[
                        hide axis,
                        legend style={
                            at={(0,1)},
                            anchor=north west,
                            draw=none,
                            fill=none,
                            legend cell align={left},
                            nodes={right},
                        },
                    ]
                    \addplot[blue] coordinates {(0,0)};
                    \addlegendentry{$k=t$}
                    \addplot[orange] coordinates {(0,0)};
                    \addlegendentry{$k=5$}
                    \addplot[teal, dashed] coordinates {(0,0)};
                    \addlegendentry{$k=0$}
                \end{axis}
            \end{tikzpicture}
        \end{minipage}
    \end{tabular}

    \caption{The average unrolling errors for observations, hidden states, policies, values, and rewards in Pong. The x-axis represents the time step ($t$), and the y-axis represents the error.}
    \label{fig:unroll-errors-Pong}
\end{figure*}

Furthermore, we examine the quality of the unrolled hidden states explicitly using \textit{unrolling errors}, providing in-depth analyses from more perspectives: observations, hidden states, policies, values, and rewards.
Specifically, we calculate the average unrolling errors for each time step $t$ in 500 \revision{recent-training} games and 500 \revision{early-training} games using the following metrics:
\begin{itemize}
    \item $\ell_{\hat{o}_{t}}^{(k)}$, the mean squared error between $\hat{o}_t^{(k)} = d(s_t^{(k)})$ and $o_t$.
    \item $\ell_{s_{t}}^{(k)}$, the cosine similarity between $s_t^{(k)} = g(s_{t-1}^{(k-1)}, a_{t-1})$ and $s_t = h(o_t)$.
    \item $\ell_{p_{t}}^{(k)}$, the cross-entropy loss between $p_t^{(k)} = f(s_t^{(k)})$ and $p_t = f(s_t)$.
    \item $\ell_{v_{t}}^{(k)}$, the mean squared error between $v_t^{(k)} = f(s_t^{(k)})$ and $v_t = f(s_t)$.
    \item $\ell_{r_{t}}^{(k)}$, the mean squared error between $r_t^{(k)} = g(s_{t-1}^{(k-1)}, a_{t-1})$ and $u_t$.
\end{itemize}
The calculated unrolling errors in five games are plotted in Fig. \ref{fig:unroll-errors-Go}, \ref{fig:unroll-errors-Gomoku}, \ref{fig:unroll-errors-Breakout}, \ref{fig:unroll-errors-Ms_pacman}, and \ref{fig:unroll-errors-Pong}, respectively, where $k=t,5,0$ correspond to the errors of unrolling from the initial state to the current time step $t$, unrolling from five steps earlier, and no unrolling (only for observations), respectively.
Our findings suggest \revision{that} the errors generally increase; furthermore, we discover more error trends that hint at what the models learned.
First, the errors in \revision{the early-training states} are generally larger than those in \revision{the recent-training} states, which aligns with the visualization results of the decoded observations.
Second, for \revision{the recent-training} trajectories, the errors may decrease during unrolling, e.g., $\ell_{s_{t}}^{(k)}$ in Go and $\ell_{p_{t}}^{(k)}$ in Breakout, implying that the model predicts less accurate in the beginning phase than in the later phases of the game.
Third, the errors may present zigzag trends, e.g., $\ell_{s_{t}}^{(k)}$, $\ell_{p_{t}}^{(k)}$, and $\ell_{v_{t}}^{(k)}$ in Go, showing that the model predicts the first player's moves better than the second player's.
Fourth, the errors may converge as $t$ increases, e.g., $t \gtrapprox 100$ in Breakout and Ms. Pacman, implying that the model may prevent the errors from increasing indefinitely.
Fifth, the errors may correlate with game-specific characteristics.
In Go and Gomoku, the errors $\ell_{s_{t}}^{(k)}$, $\ell_{p_{t}}^{(k)}$, and $\ell_{v_{t}}^{(k)}$ present zigzag trends, where the second player's moves have higher errors, showing that the model predicts the first player's moves better than the second player's.
In Pong, the errors $\ell_{v_{t}}^{(k)}$ and $\ell_{r_{t}}^{(k)}$ show periodic spikes, which are due to \revision{the} misalignment of the prediction for the periodically obtained rewards $u_t$.
In summary, the error between decoded and true observations gradually accumulates as the unrolling step increases across all games, with errors in board games generally increasing less than in Atari games, implying that the dynamics models learned for board games are more \revision{robust than} those in Atari games.

\section{Exploring the impact of hidden states during planning}\label{sec:exploring-planning}

The previous section demonstrates that hidden states are not always accurate.
However, MuZero still achieves superhuman performance despite these inaccuracies.
This section aims to interpret the search behavior in MuZero and explore the impact of hidden states during planning.

\subsection{Visualizing the search tree}\label{sec:exploring-planning:visualizing-tree}

\begin{figure*}[tp]
    \centering
    \includegraphics[width=0.8\linewidth, trim=1.3cm 1.3cm 1.3cm 1.3cm, clip]{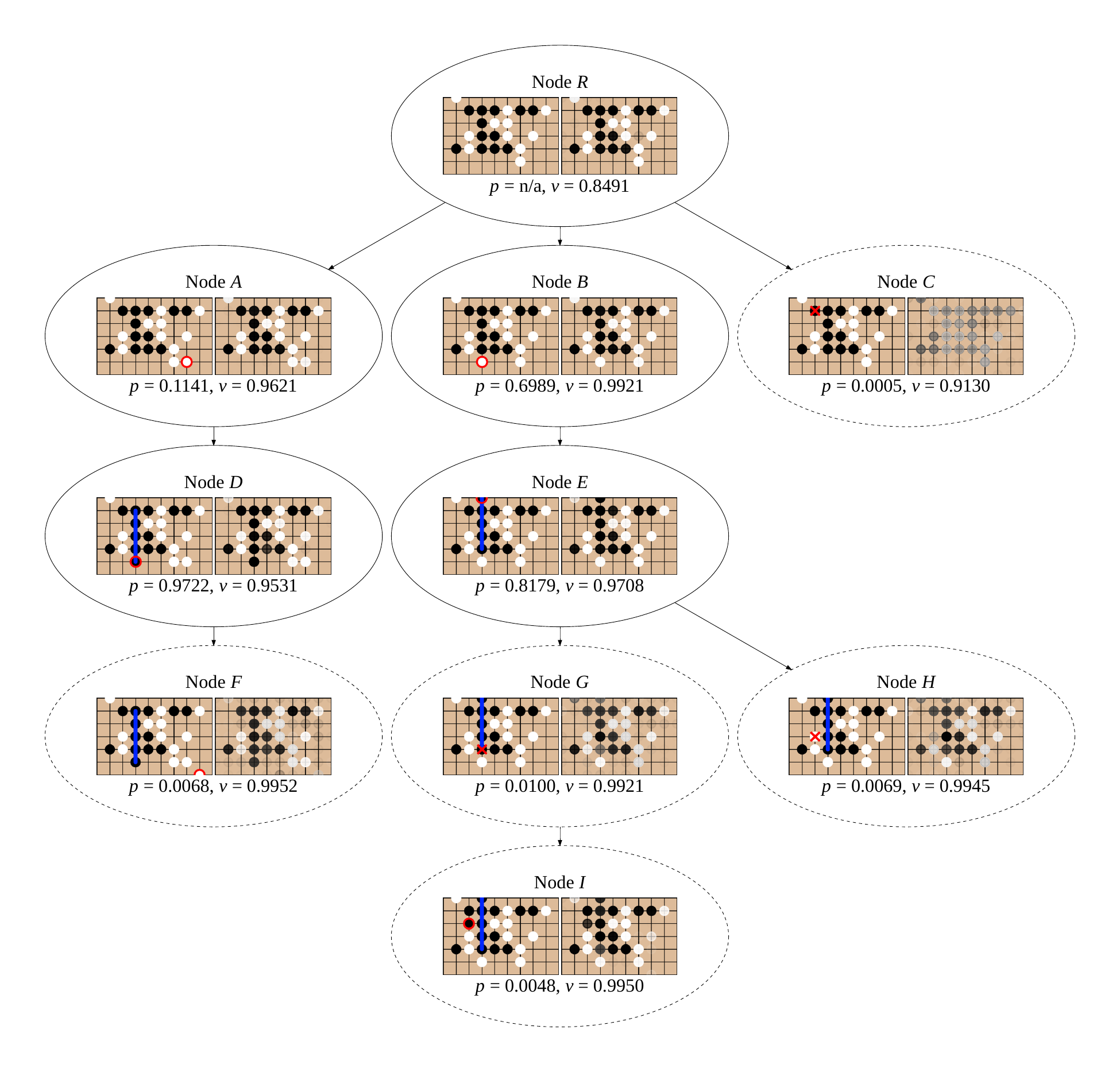}
    \caption{A visualized MuZero search tree in Gomoku. The left and right boards in each node represent true and decoded observations, respectively. Nodes enclosed by solid ellipses represent valid states during the game, while nodes enclosed by dashed ellipses represent invalid states.}
    \label{fig:muzero-search-tree}
\end{figure*}

We visualize an MCTS tree by decoding hidden states of internal nodes in Gomoku -- a game with simple and easily understandable rules, where a player wins by connecting five stones in a row -- as shown in Fig. \ref{fig:muzero-search-tree}.
The search tree is rooted at node $R$, and each node contains two boards (cropped for clarity), the left shows the Gomoku board with the action marked, and the right shows the decoded observation from the hidden state.
Each node also includes its corresponding policy $p$ and value $v$, where the values are from Black's perspective.
The search tree contains both valid and invalid states.
For valid states (nodes $R$, $A$, $B$, $D$, and $E$), the decoder produces accurate visualizations.
Invalid states (nodes $C$, $F$, $G$, $H$, and $I$) include actions such as placing a stone on an occupied position, as in node $C$, or actions taken after terminal states, as in node $F$.
The decoded observations for these invalid states become blurry, reflecting that MuZero is unfamiliar with them.

\begin{figure}[tp]
    \centering
    \footnotesize
\begin{tikzpicture}
    \begin{axis}[
        width=1\linewidth, height=6cm,
        xlabel={Steps to the terminal ($k$)},
        ylabel={Proportion in the tree},
        legend pos=north east,
        legend cell align={left},
        ymin=0, ymax=81,
        xmin=1, xmax=20,
        yticklabel={\pgfmathprintnumber{\tick}\%},
        legend style={
            fill=none,
            column sep=1ex,  
            row sep=0.5ex,   
            draw=none, 
            nodes={inner sep=1pt}, 
        }
    ]
    \pgfplotstableread{
        before_terminal	16	32	50	100	200	400	800	1600
        1	67.20	70.61	76.44	71.76	72.02	69.98	72.27	79.36
        2	51.24	57.33	62.34	55.57	59.89	61.42	60.40	72.88
        3	30.88	40.53	41.78	38.98	44.94	44.87	49.46	56.85
        4	22.34	30.69	31.96	31.05	38.08	40.24	45.49	53.99
        5	13.61	23.98	25.74	24.69	31.91	33.43	38.37	45.79
        6	10.77	17.85	21.68	21.56	28.67	28.22	33.67	41.03
        7	6.44	9.87	15.05	16.90	22.02	23.45	27.25	34.65
        8	4.52	7.36	11.23	13.05	17.41	18.86	23.18	35.20
        9	3.09	5.38	6.59	10.20	12.04	15.30	20.03	28.95
        10	1.79	3.53	4.40	8.27	10.95	12.73	17.92	25.75
        11	0.80	2.94	3.66	6.72	7.91	9.60	13.53	23.23
        12	0.93	1.55	2.61	5.62	6.62	10.75	12.44	20.18
        13	1.36	1.14	2.85	4.40	5.39	7.86	10.56	17.90
        14	0.43	0.65	1.68	3.76	4.30	6.16	9.26	16.41
        15	0.50	0.54	1.11	2.80	4.90	4.44	7.57	15.52
        16	0.25	0.91	1.34	2.87	3.06	4.89	7.59	13.87
        17	0.58	0.50	0.41	2.45	2.63	5.16	5.91	10.79
        18	0.84	0.41	0.70	2.59	2.70	4.55	4.82	10.66
        19	1.10	0.07	0.48	1.79	2.28	2.66	3.70	9.50
        20	0.47	0.04	0.69	1.56	2.59	2.54	3.58	10.25
    }\datatable

    \addplot table[x=before_terminal, y=16] from \datatable;
    \addplot table[x=before_terminal, y=50] from \datatable;
    \addplot table[x=before_terminal, y=100] from \datatable;
    \addplot table[x=before_terminal, y=400] from \datatable;
    \addplot table[x=before_terminal, y=800] from \datatable;
    \addplot table[x=before_terminal, y=1600] from \datatable;

    \legend{16, 50, 100, 400, 800, 1600}
    \end{axis}
\end{tikzpicture}
\caption{Percentages of hidden states unrolled beyond the terminal in the search tree at time step $T-k$ for different simulations in Gomoku. $T$ indicates the time step of the terminal state.}
\label{fig:nodes-after-terminal-when-approaching-terminal}
\end{figure}

Fig. \ref{fig:muzero-search-tree} also shows some interesting results.
First, several invalid states have been unrolled beyond the ground truth terminal states, hereafter referred to as \textit{beyond-terminal states}, such as node $F$.
To check how many beyond-terminal states are inside the search tree, we count their occurrences in the search trees with different sizes, illustrated in Fig. \ref{fig:nodes-after-terminal-when-approaching-terminal}.
The proportion of beyond-terminal states grows not only when approaching the environment terminal (e.g., within five steps to the terminal) but also when the search tree size is increased (e.g., 1600 simulations), which is reasonable as the search tree has no choice but to expand them when there are fewer and fewer valid moves.
Second, at node $F$, White achieves five-in-a-row after Black has already won; nevertheless, this move does not affect the value $v$.
The values of the beyond-terminal states maintain confidence about the result as the states go deeper.
For example, the value of node $I$ is higher than \revision{those of} nodes $G$ and $H$, and their values are all higher than \revision{that of} node $E$.
To further confirm this phenomenon, we calculate the value error of the beyond-terminal states in 500 \revision{recent-training} game trajectories.
For each game, we obtain the unrolled terminal state $s_{T}^{(T)}$, and further unroll 100 steps beyond the terminal to obtain $s_{T+k}^{(T+k)}$ for $1 \leq k \leq 100$, then calculate the mean square errors between $v_{T+k}^{(T+k)}$ and the game outcome $z$, plotted in Fig. \ref{fig:after-terminal-value-error-gomoku}.
Surprisingly, the values remain accurate within 10 unrolling steps beyond the terminal.
Our finding suggests that the values remain accurate when they do not go beyond the terminal state too much, implying that the information on the game outcome is correctly passed during unrolling.
This explains why MuZero maintains its performance even with inevitable beyond-terminal states during planning.

\begin{figure}[tp]
    \centering
    \footnotesize
\begin{tikzpicture}
    \begin{axis}[
            width=9cm, height=5cm,
            xlabel={Unrolling steps after terminal ($k$)},
            ylabel={MSE},
            legend pos=outer north east,
            legend cell align={left},
            xticklabel style={/pgf/number format/1000 sep=},
            ytick={0.1,1,2},
            yticklabel={\pgfmathprintnumber{\tick}},
        ]
        \pgfplotstableread[col sep=comma]{figures/after-terminal-gomoku/MuZero_Decoder-Gomoku-After-Terminal-Value-Error.csv}\datatable
        
        \addplot+[mark size=0pt] table[x=step, y=mse
] from \datatable;
        
    \end{axis}
\end{tikzpicture}
\vspace*{0.1em}%
    \caption{The average value errors between $v_{T+k}^{(T+k)}$ and game outcome $z$ in Gomoku.}
    \label{fig:after-terminal-value-error-gomoku}
\end{figure}

\subsection{Analyzing unrolled values in search tree}\label{sec:exploring-planning:analyzing-tree}
In MCTS, the search relies on \textit{mean values} averaged from subtree nodes rather than a single value from the current hidden state.
In this subsection, we investigate how the mean values impact the search tree \revision{in} Go, Gomoku, and Pong, whose dynamics networks are trained comparably robust.
Specifically, we compare the N-step mean value error for every time step $t$ between the averaged mean values from the representation network $\frac{1}{N}\sum_{n=1}^{N}{v_{t+n-1}}$\revision{,} and \revision{those from} the dynamics networks $\frac{1}{N}\sum_{n=1}^{N}{v_{t+n-1}^{(n)}}$ to approximate the error accumulated during unrolling. 
A larger $N$ indicates that the mean values are averaged with deeper unrolled states, simulating scenarios of planning in a large search tree.
Fig. \ref{fig:N-mean-error} illustrates the N-step mean value error in Go, Gomoku, and Pong for $N=1,15,30$, corresponding to \revision{the} average tree depths when using 1, 400, \revision{and} over 10000 simulations in Go and Gomoku, and using 1, 50, \revision{and} over 400 simulations in Pong.
Interestingly, the N-step mean value errors show different trends in the three examined games.
For Go, when $N=1$, the errors increase with more unrolling steps, reflecting that the unrolled values become inaccurate with longer unrolling.
Surprisingly, the errors become significantly lower when $N>1$ compared to when $N=1$.
This is likely because the unrolled values fluctuate, both overestimating and underestimating, resulting in a more stable mean value that mitigates the overall error.
For Gomoku, the trend of errors is similar to that \revision{of} Go.
However, the errors converge around 0.2, suggesting that, despite hidden states becoming more inaccurate as the number of unrolling steps increases, the errors in the dynamics network do not grow indefinitely.
Note that the errors of $N=15$ and $N=30$ are nearly identical since Gomoku games typically last around 30 to 40 steps, leaving most mean values calculated without sufficient $N$ unrolling steps.
For Pong, the errors of $N=1$ also converge, even in longer episodes with hundreds of unrolling steps.
Furthermore, changing $N$ has no significant impact on the errors, indicating that the unrolled values are bounded and stably overestimate (or underestimate) the target values within a confined error range.
Overall, the results demonstrate that the errors may be mitigated or bounded.
Therefore, MuZero can still achieve superhuman performance even if \revision{the} unrolled values are inaccurate.

\begin{figure*}[tp]
    \centering
    \footnotesize

\hspace{2.5em}
\begin{tikzpicture}
    \begin{axis}[
            width=5.2cm, height=5cm,
            ylabel={MSE},
            legend pos=outer north east,
            legend cell align={left},
            xticklabel style={/pgf/number format/1000 sep=},
            yticklabel={\pgfmathprintnumber{\tick}},
            tick label style={/pgf/number format/fixed}
        ]
        \pgfplotstableread[col sep=comma]{figures/unroll-N-mean-error/MuZero_Decoder-Go-Tree-Search-Error.csv}\datatable
        
        \addplot[teal] table[x=step, y={N=1}] from \datatable;
        \addplot[orange] table[x=step, y={N=15}] from \datatable;
        \addplot[blue, dashed] table[x=step, y={N=30}] from \datatable;
    \end{axis}
    \node[anchor=north, yshift=-0.5cm] at (current axis.south) {(a) Go};
\end{tikzpicture}
\begin{tikzpicture}
    \begin{axis}[
            width=5.2cm, height=5cm,
            legend pos=outer north east,
            legend cell align={left},
            xticklabel style={/pgf/number format/1000 sep=},
            yticklabel={\pgfmathprintnumber{\tick}},
            tick label style={/pgf/number format/fixed}
        ]
        \pgfplotstableread[col sep=comma]{figures/unroll-N-mean-error/MuZero_Decoder-Gomoku_N-mean-final.csv}\datatable
        
        \addplot[teal] table[x=step, y={N=1}] from \datatable;
        \addplot[orange] table[x=step, y={N=15}] from \datatable;
        \addplot[blue, dashed] table[x=step, y={N=30}] from \datatable;
    \end{axis}
    \node[anchor=north, yshift=-0.5cm] at (current axis.south) {(b) Gomoku};
\end{tikzpicture}
\begin{tikzpicture}
    \begin{axis}[
            width=5.2cm, height=5cm,
            legend pos=outer north east,
            legend cell align={left},
            log basis y={10},
            xticklabel style={/pgf/number format/1000 sep=},
            legend style={
                fill=none,
                column sep=1ex,  
                row sep=0.5ex,   
                draw=none, 
                nodes={inner sep=1pt}, 
            }
        ]
        \pgfplotstableread[col sep=comma]{figures/unroll-N-mean-error/MuZero_Decoder-Pong_N-mean-final.csv}\datatable
        
        \addplot[teal] table[x=step, y={N=1}] from \datatable;
        \addplot[orange] table[x=step, y={N=15}] from \datatable;
        \addplot[blue, dashed] table[x=step, y={N=30}] from \datatable;
        
    \end{axis}
    \node[anchor=north, yshift=-0.5cm] at (current axis.south) {(c) Pong};
\end{tikzpicture}
            \begin{tikzpicture}
                \begin{axis}[
                        hide axis,
                        legend style={
                            at={(0,1)},
                            anchor=north west,
                            draw=none,
                            fill=none,
                            legend cell align={left},
                            nodes={right},
                        },
                    ]
                    \addplot[teal] coordinates {(0,0)};
                    \addlegendentry{$N=1$}
                    \addplot[orange] coordinates {(0,0)};
                    \addlegendentry{$N=15$}
                    \addplot[blue, dashed] coordinates {(0,0)};
                    \addlegendentry{$N=30$}
                \end{axis}
            \end{tikzpicture}
    \caption{The average $N$-step mean value errors in Go, Gomoku, and Pong. The x-axis represents the time step ($t$), and the y-axis represents the error.}
    \label{fig:N-mean-error}
\end{figure*}

\revision{
\subsection{Analyzing planning with an inaccurate dynamics network}\label{sec:exploring-planning:analyzing}

The inaccuracy of hidden states in the dynamics network with increasing unroll steps is complex, as MuZero training involves many factors.
However, in Section \ref{sec:exploring-planning:analyzing-tree}, using the $N$-step mean value helps alleviate this issue.
We hypothesize that although the dynamics network introduces errors, they remain within bounds such that the overall value predictions still guide the planning effectively.
In this subsection, we provide a theoretical perspective based on the UCT (Upper Confidence bounds applied to Trees) foundation.

We begin by recalling the UCT foundation that the search converges to the optimal action as the number of simulations grows \cite{kocsis_bandit_2006, auer_finitetime_2002}.
The search process for a state with $K$ actions is formulated as a $K$-armed bandit problem defined by random variables $X_{i,t}$.
For each action $i$ with its selected times $t$, the playout results $X_{i,1}, X_{i,2}, \dots$ satisfy $\mathbb{E}[X_{i,t}] = \mu_i$, where $\mu_i$ is the unknown expectation of $i$.
At each time step $T$, UCT selects an action
\begin{equation}
I_T = \operatorname*{arg\,max}_{i\in\{1,\dots,K\}}\Bigl(\overline{X}_i + \sqrt{\frac{2 \ln n}{n_i}}\Bigr),
\end{equation}
where $\overline{X}_i = \frac{1}{n_i}\sum_{t=1}^{n_i} X_{i,t}$ is the empirical mean, $n_i$ the count of plays of $i$, and $n$ is the total count of plays at current time step $T$.
As the time step $T$ grows, the failure probability converges to zero \cite[Theorem~5]{kocsis_bandit_2006}, namely UCT asymptotically converges to the best action $I^*$:
\begin{equation}\label{eq:mcts-convergence}
\lim_{T \to \infty} \Pr\bigl[I_T \neq I^*\bigr] = 0.
\end{equation}

Next, we explain why the $N$-step mean value mitigates the inaccuracies in MuZero under two assumptions drawn from our empirical findings.
\begin{assumption}[Error increases with more unrolling, derived from Fig. \ref{fig:unroll-errors-Go} and Fig. \ref{fig:unroll-errors-Gomoku}]\label{as:error-increases-with-more-unrolling}
For an arbitrary hidden state $s^{(k)}$ that has been unrolled for $k$ steps, its value estimate $v^{(k)}$ contains an underlying value prediction error $\delta^{(k)}$ that tends to increase with the unroll depth $k$. 
\end{assumption}
\begin{assumption}[Mean error decreases with more value predictions, derived from Fig. \ref{fig:N-mean-error}]\label{as:mean-error-decreases-with-more-values}
For an arbitrary sequence of hidden states $s^{(0)}, s^{(1)}, \dots, s^{(N)}$ that recursively unrolls child actions from $s$, the mean of $\delta^{(0)}, \delta^{(1)}, \dots, \delta^{(N)}$ tends to decrease as more steps are accumulated. 
\end{assumption}

For MuZero, consider a bandit problem that uses a value prediction by neural network for each playout, namely $X_{i,t} = v_{i,t} = \mu_i + \delta_{i,t}$, where $v_{i,t}$ contains a true expected value $\mu_i$ and an error $\delta_{i,t}$.
Here, the empirical mean 
\begin{equation}\label{eq:empirical-mean-with-error}
\overline{X}_i = \frac{1}{n_i}\sum_{t=1}^{n_i} {(\mu_i+\delta_{i,t})} = \mu_i + \frac{1}{n_i}\sum_{t=1}^{n_i} {\delta_{i,t}} = \mu_i + \Delta_i,
\end{equation}
where $\Delta_i$ is the mean error at time step $T$.
Following the assumptions\footnote{\revision{Assumption \ref{as:error-increases-with-more-unrolling} holds for $\delta_{i,t}$ since the selected times $t$ positively correlate to the unrolling steps $k$; Assumption \ref{as:mean-error-decreases-with-more-values} holds for $\Delta_i$ since the search tree can be broken down into multiple non-overlapping sequences.}},
$|\delta_{i,t}|$ tends to increase while $|\Delta_i|$ tends to decrease, reflecting that although the value prediction $v_{i,t}=\mu_i+\delta_{i,t}$ becomes less accurate, the overall error $\Delta_i$ might remain unchanged in expectation.
Ideally, the convergence in \eqref{eq:mcts-convergence} holds if $\Delta_i \to 0$ as time step $T \to \infty$.
In summary, any arbitrary zero‐mean noise does not alter the asymptotic guarantee, and such noise can exist in MuZero within the assumptions.
}

\subsection{Evaluating playing performance with different simulations}\label{sec:exploring-planning:evaluating-simulations}

The previous investigation suggests that MuZero maintains the performance using the mitigated or bounded value errors of unrolled hidden states.
Following this finding, we are interested in whether \revision{the} playing performance can be maintained as the search tree grows deeper.
We conduct experiments to evaluate the playing performance of five games using different numbers of simulations in MCTS, as shown in Fig. \ref{fig:mcts-with-more-simulations}.
We anchor the performance of 400 simulations to 100\% for each game.
For Go and Gomoku, higher numbers of simulations lead to better performance, indicating that the dynamics network is robust when planning with a large tree, where inaccurate values are efficiently mitigated.
For Pong, since the game is easy to master and the agent can consistently achieve an optimal score, performance remains the same regardless of the number of simulations.
In this case, although the dynamics network is not learned as well as in board games, it is robust enough to maintain a bounded value error.
In contrast, for Breakout and Ms. Pacman, the performance decreases after reaching thousands of simulations, possibly due to the accumulated unrolling value errors eventually \revision{harming} the search and \revision{leading} to worse performance.
This experiment demonstrates that performance is generally maintained with a small number of simulations across all games, but an excessively large search tree can decrease performance, especially when the dynamics network is not sufficiently robust.

\begin{figure}[t]
\centering
\footnotesize
\begin{tikzpicture}
    \begin{axis}[
        width=1\linewidth, height=6cm,
        xlabel={\# Simulation},
        ylabel={Playing performance},
        xlabel style={font=\footnotesize},
        ylabel style={font=\footnotesize},
        legend pos=north west,
        legend cell align={left},
        xmode=log,
        log basis x={10},
        xmin=16, xmax=20000,
        ymin=60, ymax=140,
        xtick={10, 100, 1000, 10000},
        xticklabel style={/pgf/number format/1000 sep=},
        ytick={60,80,100,120,140},
        xticklabel style={font=\scriptsize},
        yticklabel style={font=\scriptsize},
        yticklabel={\pgfmathprintnumber{\tick}\%},
        legend style={
            fill=none,
            legend columns=2, 
            column sep=1ex,  
            row sep=0.5ex,   
            font=\scriptsize,  
            draw=none, 
            nodes={inner sep=1pt}, 
        }
    ]
    \pgfplotstableread{
        sim     go      gomoku  breakout pacman pong
        1       48.6    63.1    0.0     79.0    28.3
        2       38.9    54.9    42.3    65.6    62.5
        4       50.0    53.7    82.4    88.3    79.4
        8       53.4    56.0    92.2    97.8    84.2
        16      63.2    63.9    nan     nan     nan
        18      nan     nan     102.9   104.2   97.2
        50      76.8    72.9    105.2   95.1    98.9
        100     86.7    80.5    99.9    92.6    100.0
        200     92.2    95.7    107.5   92.4    99.2
        400     100.0   100.0   100.0   100.0   100.0
        800     106.3   104.1   106.8   92.9    99.3
        1600    114.0   110.6   103.4   88.2    99.3
        3200    122.2   111.8   99.5    87.3    99.1
        5000    124.0   116.8   100.8   87.0    99.0
        10000   119.8   117.8   89.9    81.8    99.1
        20000   130.3   122.2   72.4    73.5    102.9
    }\datatable
    
    \addplot table[x=sim, y=go] from \datatable;
    \addplot table[x=sim, y=gomoku] from \datatable;
    \addplot table[x=sim, y=breakout] from \datatable;
    \addplot table[x=sim, y=pacman] from \datatable;
    \addplot table[x=sim, y=pong] from \datatable;

    \legend{Go, Gomoku, Breakout, Ms. Pacman, Pong}
    \end{axis}
\end{tikzpicture}
\caption{Playing performance with different numbers of simulations in five games.}
\label{fig:mcts-with-more-simulations}
\end{figure}

\revision{
\section{Ablation Study}\label{sec:ablation_study}

To assess how the observation reconstruction and the state consistency affect the MuZero performance, we conduct an ablation study in Go and Breakout.
The hyperparameters for training the ablation models follow those in Table \ref{tab:hyperparameters}, except that $\lambda_d\in\{0.1,0.5,1.0,2.0\}$ and $\lambda_c\in\{0,1\}$ for Go; $\lambda_d\in\{1,15,25,35\}$ and $\lambda_c\in\{0,1\}$ for Breakout.
From Table \ref{tab:ablation-compare-ld-lc-go} and Table \ref{tab:ablation-compare-ld-lc-breakout}, adjusting $\lambda_d$ and $\lambda_c$ shows no clear trend in playing performance.
The scores stay within a range without consistently going up or down.
Since our goal is to interpret the planning processes rather than maximize win rates, these moderate scores are sufficient.

\begin{table}[t]
\centering
\revision{
    \caption{The comparison of the playing performance between MuZero with different $\lambda_d$ and $\lambda_c$ in Go.}
    \begin{tabular}{lrrrr}
        \toprule
        \diagbox[height=1.5\line, innerwidth=0.3cm, innerleftsep=0.1cm, innerrightsep=0.1cm]{$\lambda_c$}{$\lambda_d$} & $0.1$ & $0.5$ & $1.0$ & $2.0$ \\
        \midrule
        $0$ & \textbf{1130.94} & 918.63 & 1088.74 & 947.49 \\
        $1$ & 965.14 & 1034.86 & 977.39 & 1040.13 \\
        \bottomrule
    \end{tabular}
    \label{tab:ablation-compare-ld-lc-go}
\vspace*{0.8em}
    \caption{The comparison of the playing performance between MuZero with different $\lambda_d$ and $\lambda_c$ in Breakout.}
    \begin{tabular}{lrrrr}
        \toprule
        \diagbox[height=1.5\line, innerwidth=0.3cm, innerleftsep=0.1cm, innerrightsep=0.1cm]{$\lambda_c$}{$\lambda_d$} & $1$ & $15$ & $25$ & $35$ \\
        \midrule
        $0$ & 350.90 & 351.47 & 330.61 & 367.12 \\
        $1$ & \textbf{384.12} & 382.47 & 358.90 & 359.39 \\
        \bottomrule
    \end{tabular}
    \label{tab:ablation-compare-ld-lc-breakout}
}
\end{table}

\begin{figure}[t]
    \revision{
    \centering
    \footnotesize
    {\footnotesize\setlength{\tabcolsep}{0.1em}
    \begin{tabular}[]{cccccc}
        & \diagbox[font=\footnotesize,height=1.6\line,innerwidth=0.6cm]{$\lambda_c$}{$\lambda_d$} & $0.1$ & $0.5$ & $1.0$ & $2.0$ \\
        \multirow{2}{*}{\includegraphics[height=0.085\textwidth, valign=c]{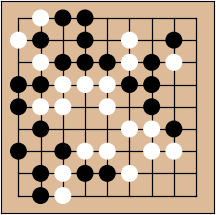}} &
        $0$ &
        \includegraphics[height=0.085\textwidth, valign=c]{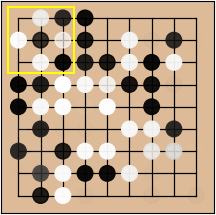} &
        \includegraphics[height=0.085\textwidth, valign=c]{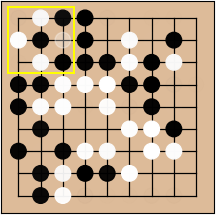} &
        \includegraphics[height=0.085\textwidth, valign=c]{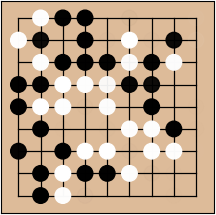} &
        \includegraphics[height=0.085\textwidth, valign=c]{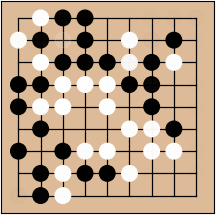} \vspace{0.3em} \\
        & $1$ &
        \includegraphics[height=0.085\textwidth, valign=c]{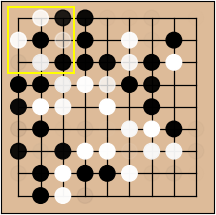} &
        \includegraphics[height=0.085\textwidth, valign=c]{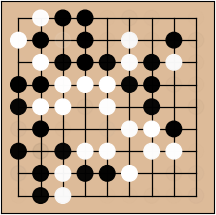} &
        \includegraphics[height=0.085\textwidth, valign=c]{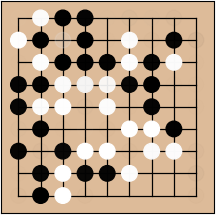} &
        \includegraphics[height=0.085\textwidth, valign=c]{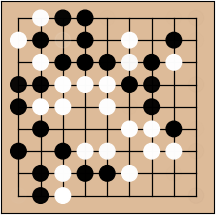} \vspace{0.5em} \\
        \multicolumn{6}{c}{(a) $o_t$ (leftmost) and $\hat o_t$ in Go} \vspace{0.5em} \\
        
        & \diagbox[font=\footnotesize,height=1.6\line,innerwidth=0.6cm]{$\lambda_c$}{$\lambda_d$} & $1$ & $15$ & $25$ & $35$ \\
        \multirow{2}{*}{\includegraphics[height=0.087\textwidth, valign=c]{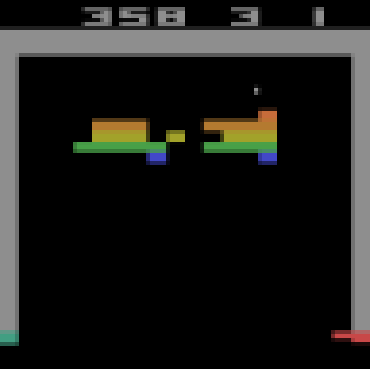}} &
        $0$ &
        \includegraphics[height=0.087\textwidth, valign=c]{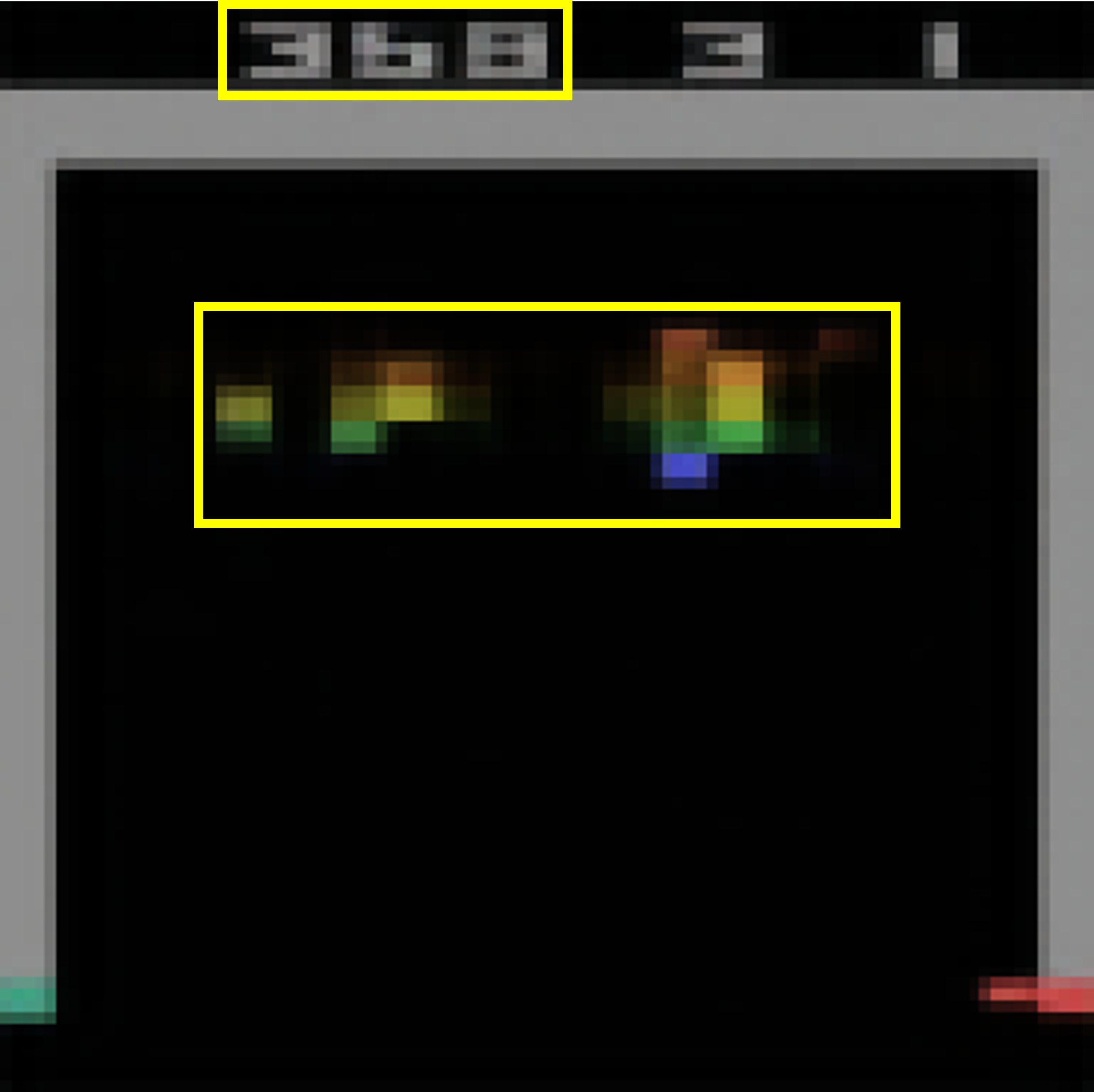} &
        \includegraphics[height=0.087\textwidth, valign=c]{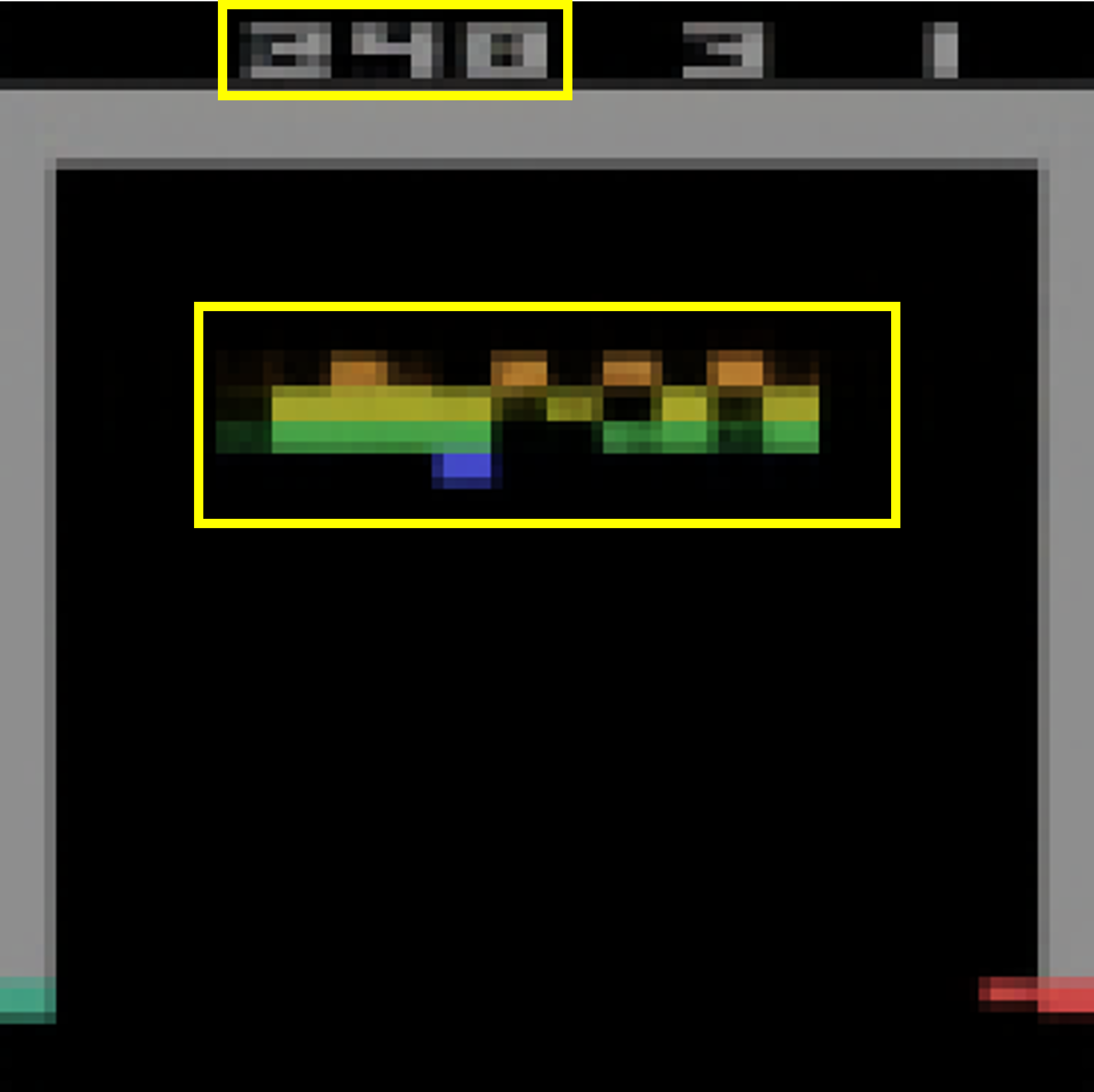} &
        \includegraphics[height=0.087\textwidth, valign=c]{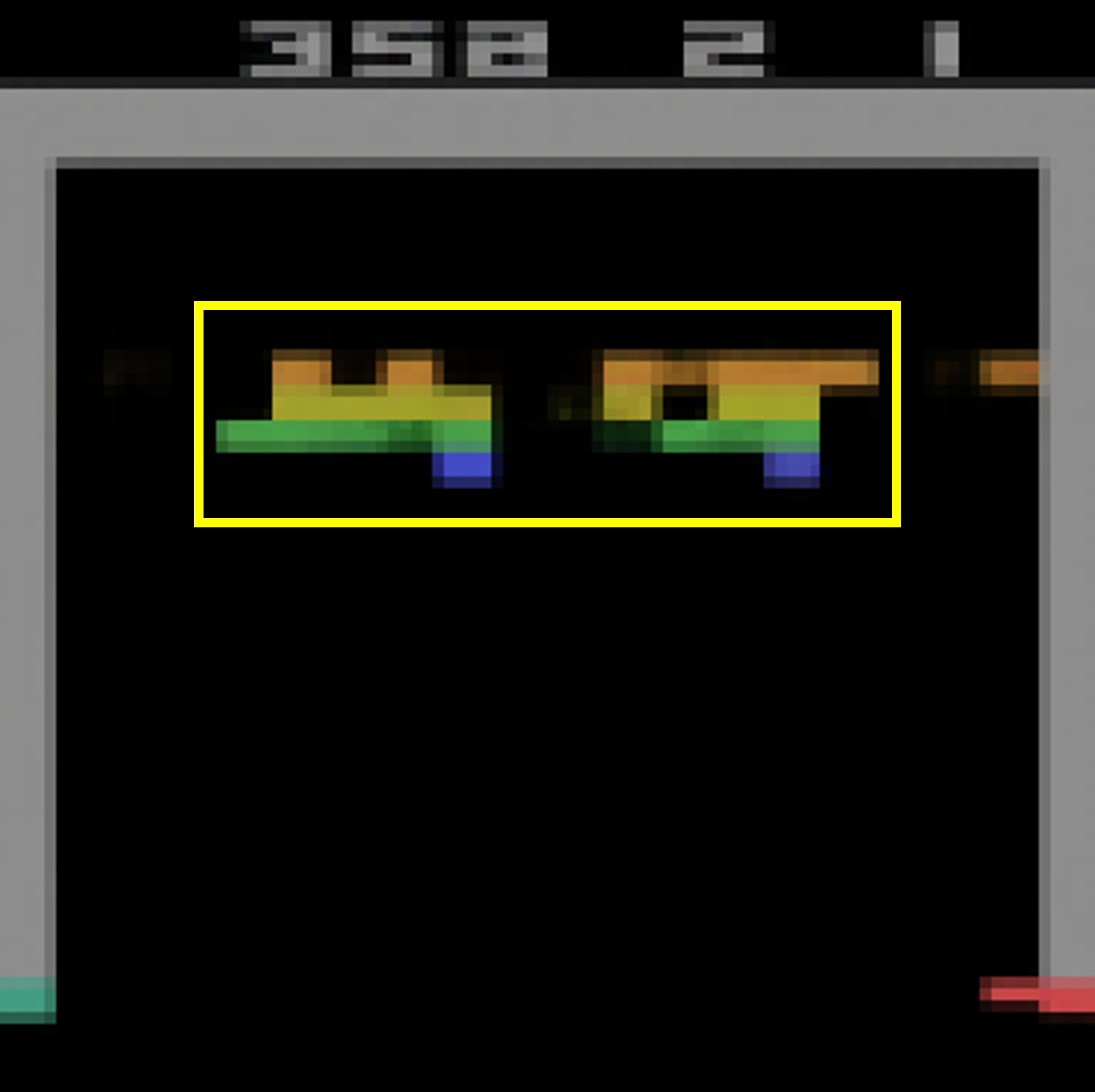} &
        \includegraphics[height=0.087\textwidth, valign=c]{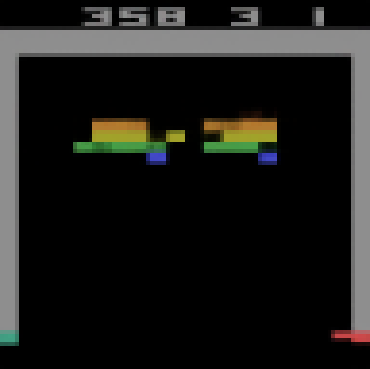} \vspace{0.2em} \\
        & $1$ &
        \includegraphics[height=0.087\textwidth, valign=c]{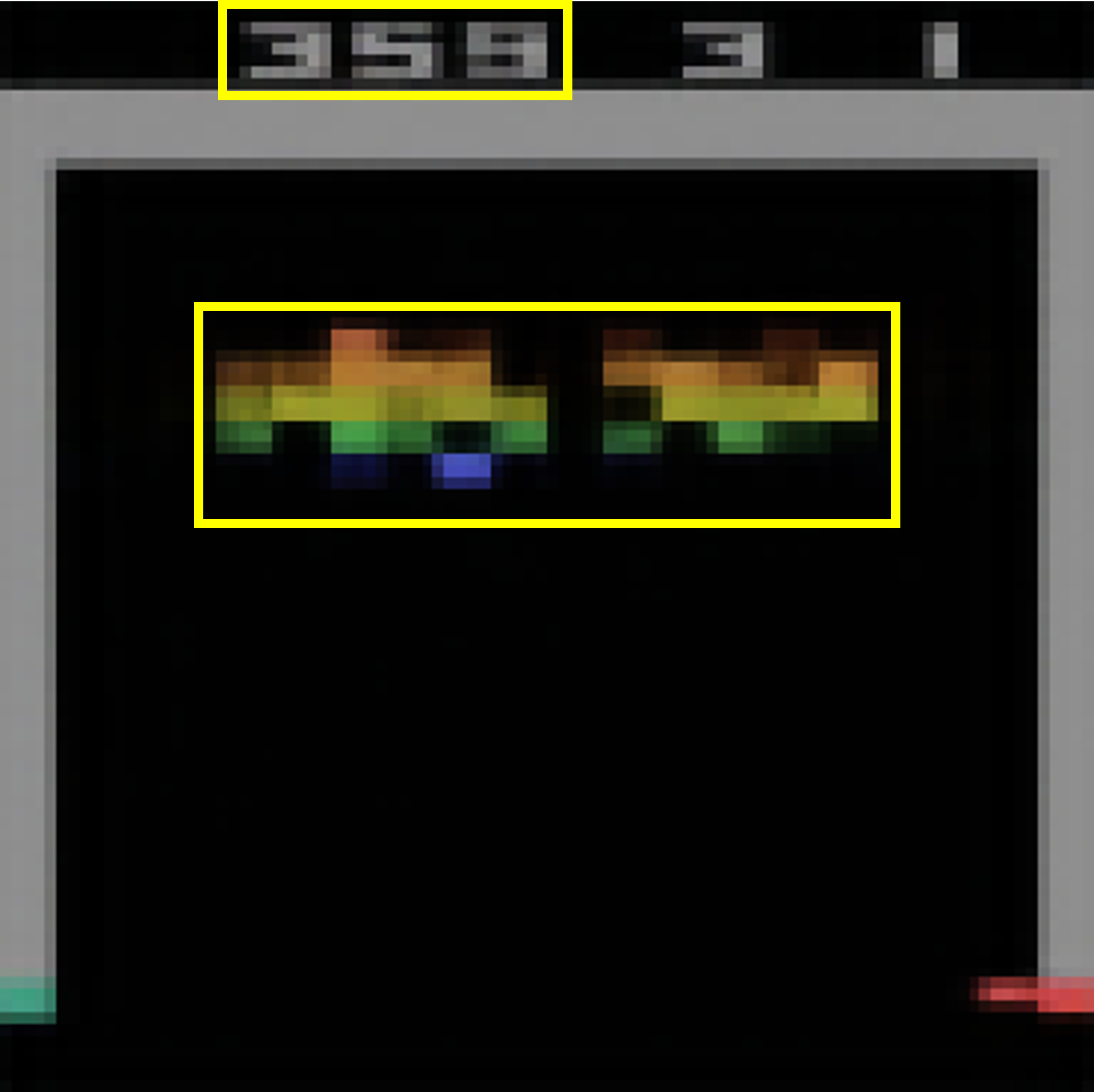} &
        \includegraphics[height=0.087\textwidth, valign=c]{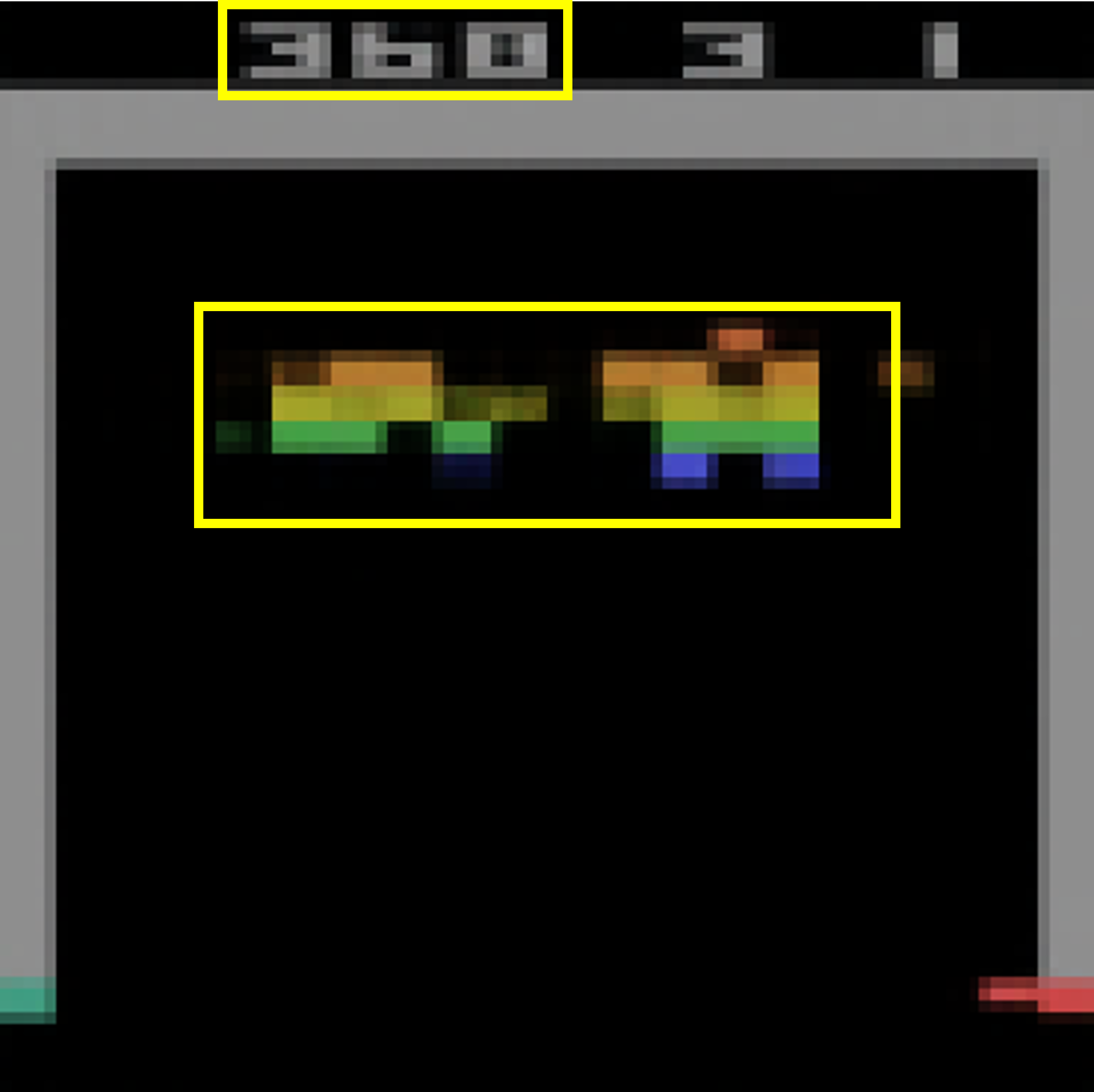} &
        \includegraphics[height=0.087\textwidth, valign=c]{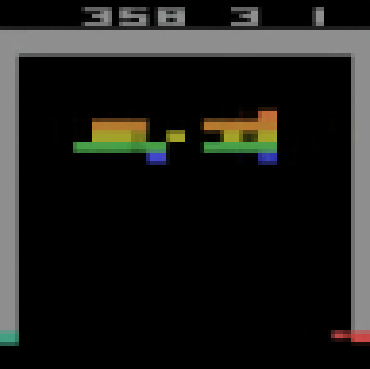} &
        \includegraphics[height=0.087\textwidth, valign=c]{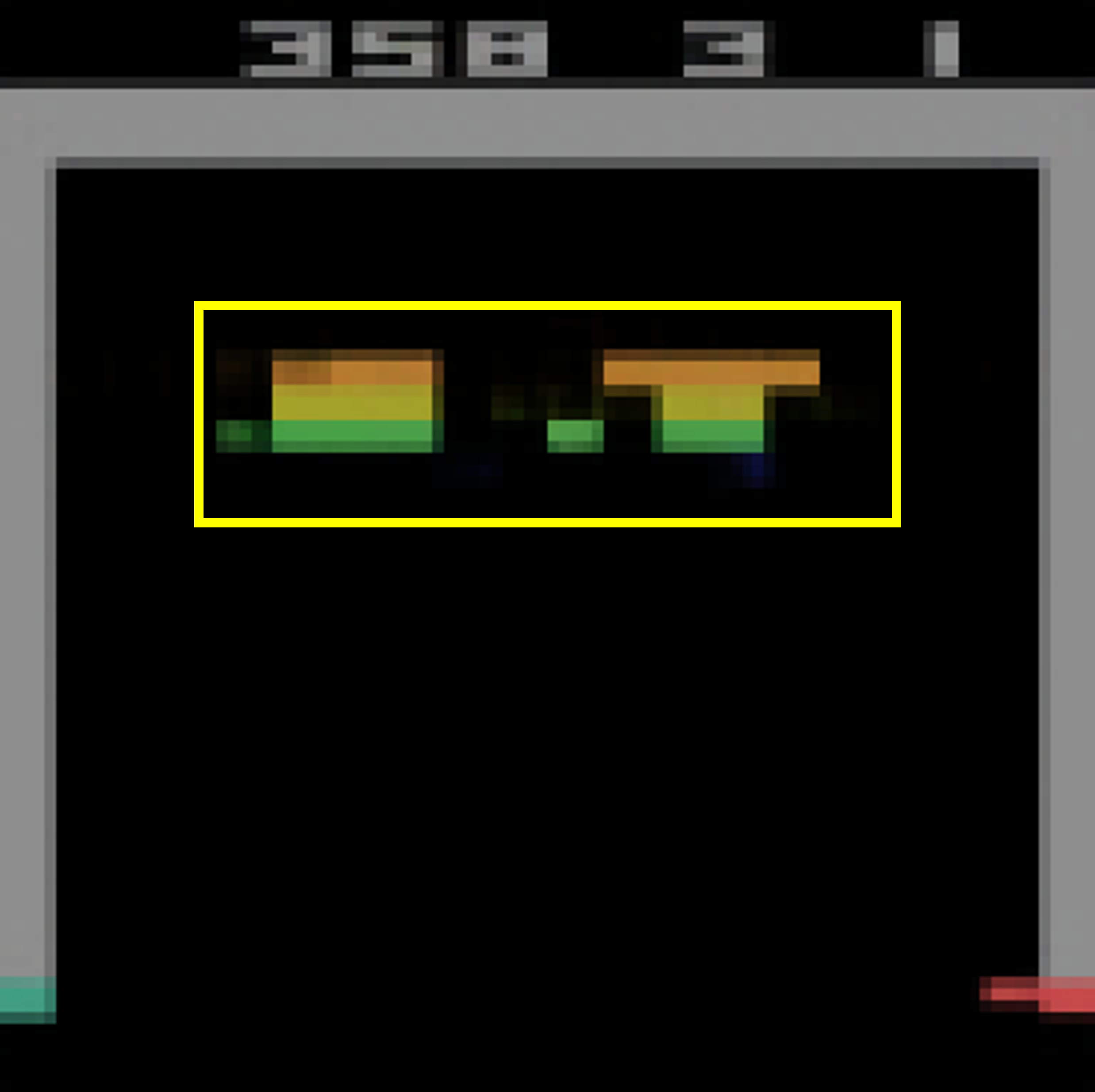} \vspace{0.5em} \\
        \multicolumn{6}{c}{(b) $o_t$ (leftmost) and $\hat o_t$ in Breakout} \vspace{0.5em} \\

    \end{tabular}}
    \caption{The comparison between true and different decoded observations for MuZero with different $\lambda_d$ and $\lambda_c$ in Go and Breakout.
    The significant errors are highlighted in the yellow boxes.}
    \label{fig:ablation-ld-lc-decoder-go-breakout}
    }
\end{figure}

Next, Fig. \ref{fig:ablation-ld-lc-decoder-go-breakout} compares the decoded observations $\hat o_t$ under different $\lambda_d$ and $\lambda_c$ settings.
In Go, all settings yield clear $\hat o_t$; only when $\lambda_d<1.0$ do misplaced stones errors appear, specifically in a \textit{ko}\footnote{\revision{In Go, \textit{ko} refers to a situation where two alternating single-stone captures would repeat the same board position, which is prohibited by the rules.}} situation in the top-left area of the board.
In addition, the choice of $\lambda_c$ has negligible effect.
As $\lambda_d=1.0$ already functions well, and to reduce the computational overhead, $\lambda_d=1.0$ with $\lambda_c=0$ are selected for our analyses in board games.
On the other hand, in Breakout, $\lambda_c$ strongly affects the reconstruction results; with $\lambda_c=1$, $\hat o_t$ becomes markedly clearer.
Across $\lambda_d$, higher values generally produce more accurate $\hat o_t$: $\lambda_d \leq 15$ is mostly incorrect, while $\lambda_d \geq 25$ is substantially better.
Nonetheless, excessively large $\lambda_d$ may destabilize training.
In practice, we found that when using large $\lambda_d$, the occasional training spikes of the decoder loss $L_d$ seriously affect both the playing and decoding performance.
To balance stability and reconstruction accuracy, we eventually choose $\lambda_d=25$ with $\lambda_c=1$ and employ gradient clipping (as mentioned in \ref{sec:interpreting-hidden-states:training}) for Atari games.
}

\section{Discussion}\label{sec:discussion}

This paper presents an in-depth analysis to demystify the learned latent states in MuZero planning across two board games and three Atari games.
Our empirical experiments demonstrate \revision{that} even if the dynamics network becomes inaccurate over longer unrolling, MuZero still performs effectively by correcting errors during planning.
Our findings also offer several future research directions.
For example, using observation reconstruction, researchers can further investigate the types of hidden states MuZero is unfamiliar with and design adversarial attack methods \cite{lan_are_2022,wang_adversarial_2023} to identify weaknesses.
Developing methods to improve state alignment can also be a promising direction, such as leveraging other world model architectures \cite{ha_world_2018, hafner_mastering_2023}.
In conclusion, we hope our contributions can assist future research in further improving the interpretability and performance of MuZero when applied to other domains.

\section*{Acknowledgment}

The authors acknowledge the use of OpenAI's ChatGPT \cite{openai_chatgpt_2024} for proofreading and language refinement of this manuscript.
The AI-assisted revisions were limited to grammar, style, and clarity improvements, without altering the technical content of the paper.

\bibliography{TAI_2024.bib}

\end{document}